\documentclass{article}

\usepackage[main,final]{neurips_2026}

\usepackage[utf8]{inputenc} 
\usepackage[T1]{fontenc}    
\usepackage{url}            
\usepackage{booktabs}       
\usepackage{amsfonts}       
\usepackage{nicefrac}       
\usepackage{microtype}      
\usepackage{xcolor}         
\usepackage{graphicx}
\usepackage{subcaption}
\definecolor{linkblue}{rgb}{0.1, 0.45, 0.8}
\usepackage[colorlinks=true, citecolor=linkblue, linkcolor=linkblue, urlcolor=linkblue]{hyperref}       
\usepackage{doi}   
\usepackage{mwe}

\usepackage{algorithm}
\usepackage{algorithmic}
\usepackage{amsmath}
\usepackage{amssymb}
\usepackage{mathtools}
\usepackage{amsthm}
\usepackage{siunitx}
\usepackage[capitalize,noabbrev]{cleveref}
\DeclareMathOperator*{\argmax}{argmax}
\DeclareMathOperator*{\argmin}{argmin}

\newcommand{\dataset}{\mathbb{D}}
\newcommand{\resterm}{\mathcal{F}}
\newcommand{\skipterm}{\mathcal{S}}

\newcommand{\skiplayerset}{\mathcal{K}_{\skipterm}}
\newcommand{\inblock}{\mathcal{V}}
\newcommand{\outblock}{\mathcal{O}}
\newcommand{\outset}{\mathcal{K}_{\mathcal{O}}}
\newcommand{\outsetfinal}{\mathcal{K}_{\mathcal{O}} \cup \left\{ N\right\}}
\newcommand{\outsetfinaltilde}{\tilde{\mathcal{K}}_{\mathcal{O}} \cup \left\{ N\right\}}
\newcommand{\lossfunc}{\mathcal{L}}
\newcommand{\entropy}{\mathcal{H}}
\newcommand{\sample}{x^{(i)}}
\newcommand{\outsample}{\hat{y}^{(i)}}
\newcommand{\labelsample}{y^{(i)}}
\newcommand{\seq}{\operatorname{seq}}

\theoremstyle{plain}
\newtheorem{theorem}{Theorem}[section]

\newtheorem{lemma}[theorem]{Lemma}

\theoremstyle{definition}
\newtheorem{definition}[theorem]{Definition}
\newtheorem{assumption}[theorem]{Assumption}
\theoremstyle{remark}

\newenvironment{proofsketch}{\begin{proof}[Proof sketch]}{\end{proof}}
\usepackage{thmtools}

\usepackage{url}
\usepackage{wrapfig}
\usepackage{tikz}
\usepackage{pgfplots}
\usepackage{varwidth}
\usepackage{enumitem}
\usetikzlibrary{arrows.meta,bending,chains,decorations.markings,decorations.pathreplacing,positioning, shapes.arrows, calc}

\title{Towards an Optimal Control Foundation of Early Exits -- Turnpikes and ResNets}

\author{%
  Jens P{\"u}ttschneider\thanks{Equal contribution.} \\
  Institute of Control Systems\\
  Hamburg University of Technology\\
  Hamburg, Germany \\
  \texttt{jens.puettschneider@tuhh.de} \\
  \And
  Simon Heilig\footnotemark[1] \\
  Faculty of Computer Science\\
  Ruhr University Bochum\\
  Bochum, Germany \\
  \texttt{simon.heilig@rub.de} \\
  \AND
  Asja Fischer \\
  Faculty of Computer Science\\
  Ruhr University Bochum\\
  Bochum, Germany \\
  \And
  Timm Faulwasser \\
  Institute of Control Systems\\
  Hamburg University of Technology\\
  Hamburg, Germany \\
}

\begin{document}

\maketitle

\begin{abstract}
The feed-forward propagation through neural networks is crucial for  
training and inference. To trade off efficiency and accuracy, it has been proposed
to consider intermediate outputs of hidden states as data-dependent regularization
during training and, ultimately, as early-exits during inference. 
In this work, we analyze the joint optimization of loss functions computed for the output layer and intermediate layers using optimal control theory. 
We prove that ResNets trained in this joint fashion exhibit the so-called turnpike phenomenon, this implies that the number of layers required to reach a prescribed loss level is bounded from above independently of the overall network depth.
This induces a similarity property across the internal representations of early-exiting networks of varying depths.
We connect the turnpike to the overthinking phenomenon, and provide theoretical insights into the success of early-exit criteria derived from softmax probabilities, specifically entropy and max-probability. 
In a range of experiments, we demonstrate the existence and effect of turnpikes and the resulting similarity in the early-exit performance for ResNets of varying depths.

\end{abstract}

\section{Introduction}
\begin{figure}[t]
	\centering
	\begin{subfigure}[b]{0.49\linewidth}
		\centering
		\resizebox{\linewidth}{!}{%
\newcommand{\layerdist}{1.35cm}
\newcommand{\layerwidth}{2.5cm}
\newcommand{\layerheight}{0.75cm}
\newcommand{\eedist}{1.75cm}   
\newcommand{\downdist}{1.0cm} 

\begin{tikzpicture}[
	rb/.style={
		draw, rectangle,
 		fill=red!15,
		rounded corners=2pt,   
		rotate=90, transform shape,
		minimum width=\layerwidth,
		minimum height=\layerheight,
		inner sep=2pt
	},
	midbox/.style={
		draw, rectangle,
 		fill=green!15,
		rounded corners=2pt,   
		minimum width=9mm,
		minimum height=6mm,
		inner sep=1pt
	},
	eebox/.style={
		draw, rectangle,
 		fill=blue!15,
		rounded corners=2pt,   
		minimum width=9mm,
		minimum height=6mm,
		inner sep=1pt
	},
	sum/.style={
		draw, rectangle,
		rounded corners=2pt,   
		rotate=90, transform shape,
		minimum width=\layerheight,
		minimum height=\layerheight,
		inner sep=2pt
	}
	]
	
	\node[rb] (input) at (-\layerdist, 0) {Input Layer};
	
	\node[rb, anchor=center] (rb0) at (0,0) {Residual Block};
	
	\node[rb, anchor=center] (rb1)   at (\layerdist, 0) {Residual Block};
	
	\node[anchor=center] (dots)  at (2*\layerdist, 0) {$\cdots$};
	
	\node[rb, anchor=center] (rbNm1) at (3*\layerdist, 0) {Residual Block};
	\node[rb, anchor=center] (rbN)   at (4*\layerdist, 0) {Residual Block};
	\node[rb, anchor=center] (outblock)   at (5*\layerdist, 0) {Output Layer};
	
	\coordinate (out-arrow-end) at (5.0*\layerdist+1cm, 0);
	\coordinate (in-arrow-end) at (-1.0*\layerdist-1cm, 0);
	\draw[->] (in-arrow-end) -- (input);
		
	\draw[->] (input) -- (rb0);
	\draw[->] (rb0) -- (rb1);
	\draw[->] (rb1) -- (dots);
	\draw[->] (dots) -- (rbNm1);
	\draw[->] (rbNm1) -- (rbN);
	\draw[->] (rbN) -- (outblock);
	\draw[->] (outblock) -- (out-arrow-end);
			
	\coordinate (p01)   at ($(rb0)!0.5!(rb1)$);
	\coordinate (p1d)   at ($(rb1)!0.5!(dots)$);
	\coordinate (pdNm1) at ($(dots)!0.5!(rbNm1)$);
	\coordinate (pNm1N) at ($(rbNm1)!0.5!(rbN)$);
	\coordinate (outputmidpoint) at ($(outblock)!0.75!(out-arrow-end)$);
	
	\fill (p01)   circle (1.2pt);
	\fill (p1d)   circle (1.2pt);
	\fill (pdNm1) circle (1.2pt);
	\fill (pNm1N) circle (1.2pt);
	\fill (outputmidpoint) circle (1.2pt);
	
	\node[eebox] (ee01)   at ($(p01)   + (0,-\eedist)$)   {EE};
	\node[eebox] (ee1d)   at ($(p1d)   + (0,-\eedist)$)   {EE};
	\node[eebox] (eedNm1) at ($(pdNm1) + (0,-\eedist)$)   {EE};
	\node[eebox] (eeNm1N) at ($(pNm1N) + (0,-\eedist)$)   {EE};
	
	\node[midbox] (loss1)   at ($(ee01)   + (0,-\downdist)$) {Loss};
	\node[midbox] (loss1d)   at ($(ee1d)   + (0,-\downdist)$) {Loss};
	\node[midbox] (lossNm1) at ($(eedNm1) + (0,-\downdist)$) {Loss};
	\node[midbox] (lossN) at ($(eeNm1N) + (0,-\downdist)$) {Loss};
	\node[midbox] (outloss) at ($(outputmidpoint) + (0, -\downdist-\eedist)$) {Loss};

	\draw[->] (p01)   -- (ee01);
	\draw[->] (ee01)  -- (loss1);
	
	\draw[->] (p1d)   -- (ee1d);
	\draw[->] (ee1d)  -- (loss1d);
	
	\draw[->] (pdNm1) -- (eedNm1);
	\draw[->] (eedNm1) -- (lossNm1);
	
	\draw[->] (pNm1N) -- (eeNm1N);
	\draw[->] (eeNm1N) -- (lossN);
	
	\draw[->] (outputmidpoint) -- (outloss);

	\node[midbox, anchor=center] (sum)   at ($(outputmidpoint) + (0, -\downdist-1.0*\eedist-1cm)$) {$\Sigma$}; 


	\coordinate (jtsumout) at ($(sum) + (1.0cm,0)$);
	\draw[->] (sum) -- 	(jtsumout);
	
	\draw[->] (outloss) -- (sum);
	
	\def\losssumarrowbase{1.0cm}
	\def\losssumarrowoffset{1.5mm}

	\coordinate (loss1-mid) at
	($ (loss1.south) + (0,-\losssumarrowbase-3*\losssumarrowoffset+3.5*\losssumarrowoffset) $);
	
	\draw[->]
	(loss1.south) |- (loss1-mid)
	-- (sum.west |- loss1-mid);
	
	\coordinate (loss1d-mid) at
	($ (loss1d.south) + (0,-\losssumarrowbase-2*\losssumarrowoffset+3.5*\losssumarrowoffset) $);
	
	\draw[->]
	(loss1d.south) |- (loss1d-mid)
	-- (sum.west |- loss1d-mid);
	
	\coordinate (lossNm1-mid) at
	($ (lossNm1.south) + (0,-\losssumarrowbase-1*\losssumarrowoffset+3.5*\losssumarrowoffset) $);
	
	\draw[->]
	(lossNm1.south) |- (lossNm1-mid)
	-- (sum.west |- lossNm1-mid);
	
	\coordinate (lossN-mid) at
	($ (lossN.south) + (0,-\losssumarrowbase-0*\losssumarrowoffset+3.5*\losssumarrowoffset) $);
	
	\draw[->]
	(lossN.south) |- (lossN-mid)
	-- (sum.west |- lossN-mid);

\end{tikzpicture}
		}
		\caption{ResNet with early-exits and joint training.}\label{fig:intro_ee_resnet}
	\end{subfigure}
	\hfill
	\begin{subfigure}[b]{0.49\linewidth}
		\centering
		\includegraphics[width=\linewidth, trim={10pt 0 0 0}, clip]{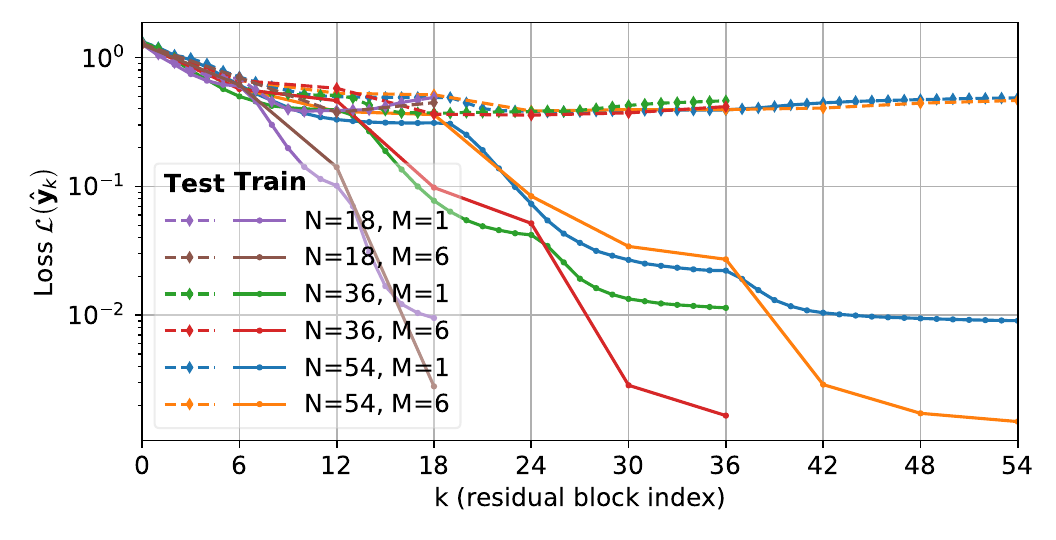}
		\caption{Loss trajectories.}\label{fig:intro_loss_trajectories}
	\end{subfigure}
	\caption{Illustration of a ResNet with early-exits (EE), the joint training formulation induces the turnpike phenomenon from optimal control, which implies that early-exiting ResNets of different depths have similar trajectories of the early-exit losses along the network depth.}
	\label{fig:loss_turnpike_loss}
\end{figure}
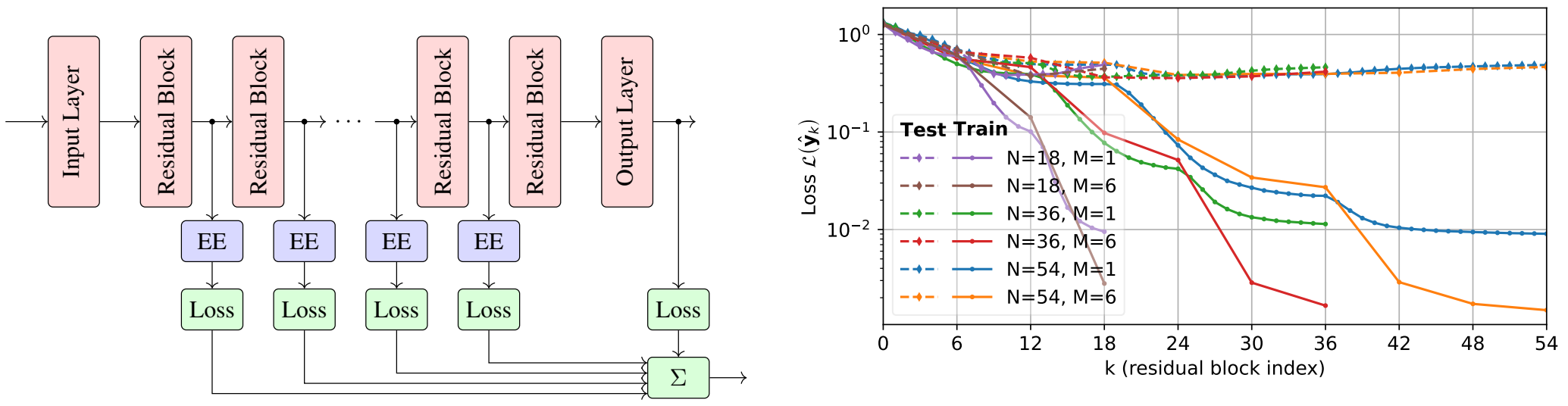

Training \textit{deep} neural networks is arguably the tipping point of today's success of deep learning models.
Key techniques are residual connections and intermediate output heads informing the network's hidden layers about their current performance \citep{szegedy2015going, he2016deep}. What was initially used only as training scheme \citep{pmlr-v38-lee15a}, quickly evolved as inference method to save computation on instances handled equally well by shallower parts of the network \citep{branchyNet} and has since been applied to early-exiting language models and vision transformers 
\citep{scardapane2020should,zhou2020bert,schuster2022confident,chen2023cf}. 
Without early-exiting, a key failure mode of deep neural networks is \textit{overthinking} \citep{kaya2019shallow}: samples correctly classified by earlier layers are propagated further, incurring unnecessary computation and potentially degrading predictions at later stages.
At the core of training these early-exit networks lies an optimization problem considering a combined loss of all auxiliary and final output heads, as illustrated 
in~\cref{fig:intro_ee_resnet}. 
As we show in this work, it is precisely this joint training formulation that gives rise to the turnpike phenomenon from optimal control, offering a principled explanation for why early-exit classifiers perform well and why overthinking can be avoided.
With the turnpike phenomenon at hand, we are able to show that the early-exit depth required to reach a prescribed loss level is independent of the overall backbone depth. Consequently, the early-exit losses of networks of different depths are similar, as illustrated in~\cref{fig:intro_loss_trajectories}.

Prior theoretical work has analyzed early-exit training either under restrictive 
assumptions such as strong convexity \citep{pmlr-v38-lee15a} or focused on 
layer-wise supervision regimes that lack the upstream gradient signal of joint 
optimization \citep{sakamoto2024endtoend}. Consequently, a fundamental understanding 
of how joint early-exit training behaves and why one can expect intermediate 
classifiers to perform well has remained elusive; a gap this work addresses through 
the lens of optimal control theory.

Motivated by residual connection-based neural networks (ResNets) and their connection to neural ODEs 
\citep{he2016deep,10.1007/978-3-319-46493-0_38,NEURIPS2018_69386f6b}, research started 
using tools from optimal control to investigate these deep learning models, where trainable 
parameters act as control inputs that steer hidden states towards a target representation determined by the loss function \citep{esteve2021largetime,ESTEVEYAGUE2023105452,ruiz2023neural}. 
A central technique is to use a \textit{stage cost} over the full forward trajectory accumulating (loss-) penalties along the trajectory the data takes through the network. This can be understood as 
shaping the transient loss behavior, which enables stronger convergence guarantees and 
explicit depth bounds \citep{esteve2021largetime,FAULWASSER2024100290}.
A commonality of these works is the turnpike phenomenon: hidden representations tend to 
cluster around the loss minimizer for the majority of layers, independent of network depth. 
However, while existing analyses focus on continuous space-time networks 
\citep{esteve2021largetime}, the application of this framework to standard ResNets 
architectures remains an open problem.
In addition, our novel optimal control perspective on early-exiting enables new avenues towards formal analysis of training formulations. 

In this work, we exploit the close relation between stage cost terms in discrete-time optimal control problems and the joint optimization formulation of early-exit neural networks. 
This allows us to show under which conditions this approach leads to the turnpike phenomenon. 
In turn this helps to uncover the fundamental mechanisms of joint training of early-exit neural networks. 
In short, our paper makes the following contributions:
\begin{itemize}[parsep=2pt, partopsep=0pt] 	
    \item A turnpike-based foundation of early-exit-based ResNet training and inference.
    \item Novel asymptotic upper bounds on the cross-entropy loss during forward propagation revealing the loss turnpike phenomenon.
    \item Theoretical bounds on guaranteeing correct classification and bounded early-exiting depths, derived for entropy and maximum softmax probability criteria.
    \item Empirical support of the turnpike analysis and a principled approach to  obtain the intermediate outputs in standard ResNets.
\end{itemize}

\section{Preliminaries}
We begin with a review of the concepts of early-exiting neural networks and optimal control theory.

\subsection{Early-Exiting ResNets}
Early-exiting, as proposed by \citet{branchyNet}, augments a neural network with intermediate classifiers that provide predictions before the final layer. The inference process is interrupted as soon as an intermediate prediction is deemed sufficiently confident, thereby reducing computational cost and preventing the model from overthinking \citep{kaya2019shallow}.

In this work, we consider standard ResNets \citep{10.1007/978-3-319-46493-0_38, dosovitskiy2021an} with $N$ residual blocks (RBs) which for a given input sample $\sample$ produce the output $\outsample$ by
\begin{equation}
\label{eq:resnet}
    \sample =\inblock \left(\sample,\mathbf{w}_{\inblock}\right), \quad
   	\sample_{k+1} =  \skipterm_k \left(\sample_k, \mathbf{w}_{\skipterm, k} \right) + \resterm_k \left(\sample_k, \mathbf{w}_{\resterm, k}\right), \quad 	\outsample = \outblock(\sample_N, \mathbf{w}_\outblock).
\end{equation}
This architecture includes the input block $\inblock$ with parameters $\mathbf{w}_\inblock$, $N$ residual blocks with skip connection $\skipterm_k$ and residual term $\resterm_k$ with parameters $\mathbf{w}_{\skipterm, k}$ and $\mathbf{w}_{\resterm, k}$ (summarized as $\mathbf{w}_k = \left[\mathbf{w}_{\skipterm, k},  
\mathbf{w}_{\resterm, k}\right]$), as well as the output head $\outblock$ with parameters $\mathbf{w}_\outblock$. 
We collect all parameters of the ResNet into $\mathbf{w} = \left[\mathbf{w}_\inblock, \mathbf{w}_0, \ldots, \mathbf{w}_{N-1}, \mathbf{w}_\outblock \right]$.
Our subsequent analysis applies to both networks with varying state dimensions, such as convolutional ResNets~\citep{10.1007/978-3-319-46493-0_38}, or ResNets with a constant state dimension.
For details on the ResNet implementation and our control-inspired notation, we defer to Appendix~\ref{sec:resnet_basics}.

In the classical case, the output head produces the \textit{final prediction} $\outsample $, while early-exit neural networks additionally have intermediate output heads $\outblock_k$  with parameters $\mathbf{w}_{\outblock, k}$ attached to the $k$-th hidden layer (where $k \in \mathcal{K}_\outblock\subseteq \mathbb{N}_{[0,N-1]}$) to obtain \textit{intermediate predictions}: $\outsample_k = \outblock_k \left(\sample_k, \mathbf{w}_{\outblock, k}\right)$.
During inference, common decision rules are based on the indication of high confidence such as by low entropy $\entropy(\outsample_k)$ or high maximum softmax probability $\hat{p}_\mathrm{max}(\outsample_k)$ \citep{Huang2017MultiScaleDN,passalis2020efficient,liu2022anytime}.

To train the backbone network and the early-exits, different strategies have been 
proposed~\citep{10.1145/3698767,krzepkowski2024joint,regol2024jointlylearned}. 
In this work, we set out to study the joint training formulation through its intrinsic 
connection to optimal control theory, considering a supervised classification problem 
with labeled training data $\dataset = \left\{\left(x^{(1)}, y^{(1)}\right), ..., 
\left(x^{(D)}, y^{(D)}\right)\right\}$, where $\mathbf{x} = \left[x^{(1)}, x^{(2)}, 
\ldots, x^{(D)}\right]$ denotes the stacked features (boldface symbols used for their 
propagation through the network), $\hat{\mathbf{y}}$ the network output over the entire 
dataset, and $\lossfunc(\hat{\mathbf{y}})$ the corresponding average cross-entropy 
loss\footnote{For simplicity of exposition, we omit the ground-truth label as an 
additional argument of the loss function.}. The main training approaches are either 
sequential---first optimizing the backbone and then keeping it frozen while training 
the early-exits---or joint, where backbone and early-exits are trained simultaneously.
This is achieved by minimizing the \emph{joint training objective}, a weighted sum of intermediate and final losses,  
\begin{equation*}
	\sum_{k \in \mathcal{K}_\outblock} \gamma\lossfunc (\hat{\mathbf{y}}_k) + 
\lossfunc (\hat{\mathbf{y}}),
\end{equation*}
where the scalar hyperparameter $\gamma \in (0,1]$ controls the trade-off between the intermediate and final losses.
Larger values of $\gamma$ emphasize the intermediate losses, while smaller values prioritize the final loss.

\subsection{The Turnpike Phenomenon in Optimal Control}
Control theory aims to determine input signals or feedback policies for dynamic systems to induce a desired trajectory behavior.
Optimal control, in particular, determines these control inputs by solving an optimization problem. 

\begin{wrapfigure}{r}{0.5\columnwidth}
	\vspace{-0.5cm}
	\centering
	\includegraphics[width=\linewidth]{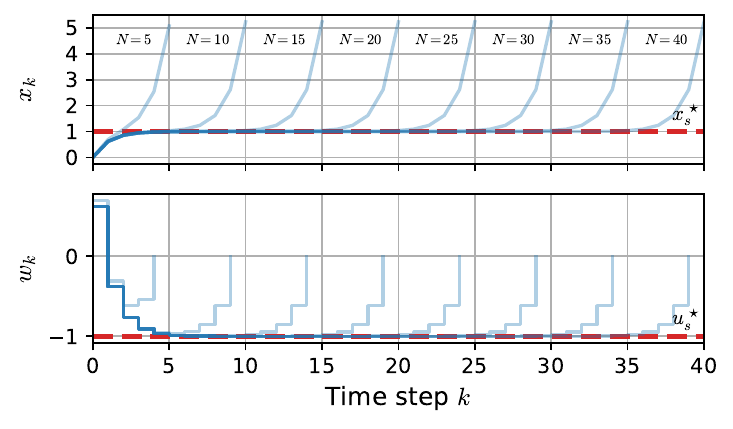}
	{\caption{Example of optimal trajectories of the OCP~\eqref{eq:basic_ocp} for varying horizons $N=5,10,\dots,40$. The red dashed lines indicate the turnpike $(x_s^\star, w_s^\star) = (1,-1)$.}
	\label{fig:example_ocp}}
	\vspace{-0.4cm}
\end{wrapfigure}

Consider the dynamic system $x_{k+1} = f(x_k, w_k)$, with the initial condition $x_0 = x^0$.
The evolution of the state $x_k \in \mathbb{R}^n$ over the discrete time index $k$ is steered by the control inputs $w_k \in \mathbb{R}^m$.
The goal of control is to determine the inputs $w_k$ as to achieve a desired behavior of the state trajectories $x_k$.
This is achieved by optimizing over the possible trajectories of this dynamic system.
A simple optimal control problem (OCP) with finite horizon $N \in \mathbb N$ reads
\begin{equation}
	\label{eq:basic_ocp}
		\min_{w_0,\dots, w_{N-1}} \sum_{k = 0}^{N-1} \ell(x_k, w_k)  + J_\mathrm{f}(x_N) \quad
		\text{s.t.} \;  x_{k+1} = f(x_k, w_k), \; x_0 = x^0, \; \forall k = 0,{\scriptsize\ldots}, N{\scriptsize-}1.
\end{equation}
The objective entails the stage cost $\ell:\mathbb{R}^n\times \mathbb{R}^m\to \mathbb{R}$, which penalizes the states and control inputs along the trajectory  and the terminal cost $J_\mathrm{f}: \mathbb{R}^n \to \mathbb{R}$ which considers the terminal state $x_N$.
The structure of the OCP depends on the system dynamics $f$ and the choice of the objective. It can range from linear systems with quadratic costs to nonlinear systems with non-convex costs (or it may include inequality constraints which we neglect here).
Such problems can be solved using various methods, ranging from direct solution using numerical optimization to applying dynamic programming/reinforcement learning to compute/approximate optimal value function~\citep{Bertsekas2005}.
\cref{fig:example_ocp} shows solutions of the simple OCP~\eqref{eq:basic_ocp} with $x_{k+1} = 2 x_k + w_k$, $x^0 = 0$, $\ell(x_k, w_k) = (x_k-2)^2 + w_k^2$, and $J_\mathrm{f}(x_N) = 0$ for varying horizons  $N=5,10,\dots, 40$.
In the middle of the horizon, all trajectories cluster around $x_s^\star = 1$ and $w_s^\star = -1$, while they depart towards the end of the horizon. 
One may observe that $x_s^\star = 1$ and $w_s^\star = -1$ is the optimal steady state pair of the OCP~\eqref{eq:basic_ocp}, i.e., the steady state pair that minimizes the stage cost $(x^\star_s, w^\star_s) = \argmin_{(x_s, w_s)} \ell(x_s, w_s), \; \text{s.t.} \; x_s = f(x_s,w_s)$. 
However, near the end of the finite horizon, the objective is minimized by departing from the optimal equilibrium, a phenomenon known as a \textit{leaving arc}.

This characteristic clustering of optimal trajectories close to the optimal steady state is known as the \textit{turnpike property}, early observations of which can be traced back to \citet{vonNeumann38}, predating the fundamental works of \citet{Bellman54a} and \citet{PMP1956}. 
The term \textit{turnpike} has been coined in the book by \citet{Dorfman58} and it has received widespread attention in control, cf.~\citet{McKenzie1976,Trelat15a}.
Indeed, the turnpike property is a similarity property of the OCP solutions, i.e., the input and state trajectories remain close to the optimal steady state for most of the time independent of the initial conditions $x_0$ and horizon lengths $N$.
A formal definition of the phenomenon is as follows.
\begin{definition}[Turnpike phenomenon]
	The OCP~\eqref{eq:basic_ocp} is said to exhibit the turnpike phenomenon with respect to $(x_s^\star, w_s^\star) = const$ if for every threshold $\varepsilon>0$, there exists a constant $\kappa < \infty$ such that the number of timesteps for which the optimal states and input lie outside of an $\varepsilon$-neighborhood of $(x_s^\star, w_s^\star)$ 
    satisfies the uniform bound $\# \left\{ k = 0, ..., N \; \middle| \; \lVert (x_k , w_k) - \left(x_s^\star, w_s^\star \right) \rVert > \varepsilon \right\} \leq \kappa$
	independent of the horizon $N$ and the initial condition $x^0$.
\end{definition}
Indeed, there exist different approaches on how turnpikes can be analyzed, we refer to \citet{Trelat15a} and \citet{tudo:faulwasser22a} for recent overviews.

\section{Optimal Control Theory for Early-Exiting}
We now formalize the joint training formulations for early-exiting ResNets from an optimal control perspective.
We begin by considering ResNets as dynamic systems and view early exit training as optimal control problem in \cref{sec:optimzation}. 
Our main contributions are twofold: in \cref{sec:mainresult}, we introduce a novel turnpike notion, and in \cref{sec:exitingturnpike}, we extend this analysis to early-exit criteria.

\subsection{Training Early-Exit ResNets as Optimal Control}\label{sec:optimzation}

ResNets with constant state dimension can be regarded as a \textit{discrete-time dynamic system} 
$
\mathbf{x}_{k+1} = f_k(\mathbf{x}_k, \mathbf{w}_k)= \mathbf{x}_k + \resterm_k(\mathbf{x}_k, \mathbf{w}_k), \quad \mathbf{x}_{0} = \mathbf{x},
$
on the finite horizon $k = 0,..., N-1$ corresponding to the number of residual blocks.
The hidden state $\mathbf{x}_k$ corresponds to the \textit{state of the system} while the trainable parameters $\mathbf{w}_k$ act as \textit{control inputs}, see also~\cite{Li17a} and~\cite{ruiz2023neural}.
An important property of ResNets is that zero parameters reduce a residual block to an identity mapping via the skip connection $\mathbf{x}_{k+1} = f_k(\mathbf{x}_k, 0) = \mathbf{x}_k + \resterm_k(\mathbf{x}_k, 0) = \mathbf{x}_k$, implying that under convergence of the training, the performance does not degrade with increasing depth, since additional residual blocks can always be bypassed~\cite{10.1007/978-3-319-46493-0_38}.
From the optimal control perspective, this implies that each state can be rendered a steady state by $\mathbf{w}_k= 0 = const$.

Next, we phrase the early-exit joint training formulation for ResNets as an OCP.
We begin with the joint training formulation for an early-exiting neural network based on ResNet~\citep{10.1007/978-3-319-46493-0_38}. Our insights are summarized in \cref{fig:illustration_equivalence_ocp_joint_training}.
We use a constant weight $0<\gamma\leq 1$ to trade off the loss function of the intermediate outputs and the final output.

In contrast to classical OCPs, where the state dimension of the dynamical system is fixed, in standard ResNets  the hidden-state dimension varies with depth~\citep{10.1007/978-3-319-46493-0_38}.
In principle, these ResNets can still strictly be considered as a dynamic system with constant state dimension by embedding the entire network dynamic into a fixed state space corresponding to the maximum dimension in the network.
Here, however, we allow the state dimension to vary and instead focus our analysis on the outputs, which have a fixed dimension and are used to obtain predictions and evaluate the loss function. 
In classification tasks, this dimension typically equals the number of classes over which softmax probabilities are computed.

\begin{figure*}[t]
	\resizebox{\textwidth}{!}{%
		\tikzset{MyArrow/.style={single arrow, draw=black, minimum width=10mm, minimum height=30mm, inner sep=2mm, single arrow head extend=1mm, align=center}}
\def\boxwidth{7.1cm}
\def\boxheight{5cm}
\def\boxgap{0.2cm} 
\def\eqnumsep{\enspace} 
\def\figeqnum{\refstepcounter{equation}\text{(\theequation)}} 

\begin{tikzpicture}[
	box/.style={
		draw,
		rectangle,
		inner sep=0pt,
		align=center
	},
	arrowbox/.style={
		draw,
		rectangle,
		minimum width=\boxwidth,
		minimum height=\boxheight,
		inner sep=0pt,
		font=\footnotesize,
		align=center
	},
	]

	\def\arrowheight{3.4cm} 

	\def\a{1.9cm} 
	\def\b{8mm} 
	\def\c{1.65cm} 

	\def\arrowxends{3.4cm}

	\node[box, draw=none, anchor=north east] (block1) at ({-\arrowxends-\boxgap},\a) {%
	\begin{subequations}%
		\label{eq:fig_ocp}%
	\begin{tabular}{@{}c@{}}
		\textbf{Optimal Control Problem}\\[1mm]
		$\begin{array}{@{}r@{}l@{\eqnumsep}r@{}}
		\hspace{-0.3cm}\displaystyle\min_{\mathbf{w}} \; J_N(\mathbf{w})
		&{}= \displaystyle\sum_{k=0}^{N-1} \ell_k(\mathbf{x}_k, \mathbf{w}_k) + J_{\mathrm{f}}(\mathbf{x}_N) & \figeqnum \\[2mm]
		\text{s.t.}\quad
		k&{}=0,\dots,N-1 & \\[1mm]
		\mathbf{x}_{k+1} &{}= f_k(\mathbf{x}_k,\mathbf{w}_k), & \figeqnum \\[1mm]
		\mathbf{x}_0 &{}= \mathbf{x}^0. & \figeqnum
		\end{array}$
	\end{tabular}%
	\end{subequations}%
	};

	\node[box, draw=none, anchor=north west] (block2) at ({\arrowxends+\boxgap}, \a) {%
	\begin{subequations}%
		\label{eq:fig_ee_training}%
	\begin{tabular}{@{}c@{}}
		\textbf{Early-Exit Joint Training}\\[1mm]
		$\begin{array}{@{}r@{}l@{\eqnumsep}r@{}}
				\displaystyle\inf_{\mathbf{w}} \; J_N(\mathbf{w})
				&{}= \displaystyle\sum_{k \in \mathcal{K}_\outblock} \gamma \lossfunc(\hat{\mathbf{y}}_k)
				+\lossfunc(\hat{\mathbf{y}}) & \figeqnum \\[2mm]
				\text{s.t.} \quad k &{}= 0,\dots,N-1, & \\[1mm]
				\mathbf{x}_{k+1} &{}=  \mathbf{x}_k + \resterm_k(\mathbf{x}_k, \mathbf{w}_k), & \figeqnum \\[1mm]
				\mathbf{x}_0 &{}= \inblock(\mathbf{x}, \mathbf{w}_\inblock), & \figeqnum \\[1mm]
				\hat{\mathbf{y}}_{k} &{}= \outblock_k(\mathbf{x}_k, \mathbf{w}_{\outblock, k}),
				\; k \in \mathcal{K}_\outblock & \figeqnum \\[1mm]
				\hat{\mathbf{y}} &{}= \outblock(\mathbf{x}_N, \mathbf{w}_{\outblock}). & \figeqnum
		\end{array}$
	\end{tabular}%
	\end{subequations}%
	};

	\node[right=10pt of block1] (arrow1) {};
	\node[left=10pt of block2]  (arrow2)   {};
	
	\node[arrowbox, draw=none, minimum height=\arrowheight, inner sep=0pt] (arrowbox) at (0,0) {
		\textbf{ResNet as dynamic system}\\
		$\displaystyle \mathbf{x}_{k+1} = f_k\left(\mathbf{x}_{k},\mathbf{w}_k \right) = \mathbf{x}_k + \resterm_k(\mathbf{x}_k,\mathbf{w}_k) $\\[1mm]
		\textbf{Initial state from input layer} \\
		$\displaystyle \mathbf{x}^0 = \inblock(\mathbf{x}, \mathbf{w}_\inblock)$ \\[1mm]
		\textbf{Early-exit loss}\\
		$\displaystyle \ell_k(\mathbf{x}_k, \mathbf{w}_k) = \gamma \lossfunc(\outblock_k(\mathbf{x}_k, \mathbf{w}_{\outblock, k}))$, $k \in \mathcal{K}_\outblock$ \\[1mm]
		\textbf{Final loss} \\
		$J_{\mathrm{f}}(\mathbf{x}_N) = \lossfunc(\outblock(\mathbf{x}_N, \mathbf{w}_{\outblock}))$
	};

	\def\nodeinnersep{0pt} 

	\newif\ifdebugpoints
	\debugpointsfalse 

	\tikzset{
		debugpoint/.style={
			circle,
			inner sep=1.5pt,
			fill=yellow
		},
		debugpointoff/.style={
			circle,
			inner sep=0pt,
			minimum size=0pt,
			draw=none,
			fill=none
		}
	}

	\draw
	(\arrowxends, 0)
	node[\ifdebugpoints debugpoint\else debugpointoff\fi] {}
	-- ({\arrowxends-\b}, \a)
	node[\ifdebugpoints debugpoint\else debugpointoff\fi] {}
	-- ({\arrowxends-\b}, \c)
	node[\ifdebugpoints debugpoint\else debugpointoff\fi] {}
	-- ({-\arrowxends+\b}, \c)
	node[\ifdebugpoints debugpoint\else debugpointoff\fi] {}
	-- ({-\arrowxends+\b}, \a)
	node[\ifdebugpoints debugpoint\else debugpointoff\fi] {}
	-- ({-\arrowxends}, 0)
	node[\ifdebugpoints debugpoint\else debugpointoff\fi] {}
	-- ({-\arrowxends+\b}, -\a)
	node[\ifdebugpoints debugpoint\else debugpointoff\fi] {}
	-- ({-\arrowxends+\b}, -\c)
	node[\ifdebugpoints debugpoint\else debugpointoff\fi] {}
	-- ({\arrowxends-\b}, -\c)
	node[\ifdebugpoints debugpoint\else debugpointoff\fi] {}
	-- ({\arrowxends-\b}, -\a)
	node[\ifdebugpoints debugpoint\else debugpointoff\fi] {}
	-- (\arrowxends, 0);

\end{tikzpicture}%
	}
	\caption{The fundamental connection between early-exiting joint training formulation and training as an optimal control problem.}\label{fig:illustration_equivalence_ocp_joint_training}
\end{figure*}
It remains to cast the early-exit joint training problem~\eqref{eq:fig_ee_training} as an OCP in the form of~\eqref{eq:fig_ocp}, cf. \cref{fig:illustration_equivalence_ocp_joint_training}.
To this end, we define the stage cost as the loss function for the intermediate outputs, i.e., $\ell_k(\mathbf{x}_k, \mathbf{w}_k) =\gamma \lossfunc(\outblock_k(\mathbf{x}_k, \mathbf{w}_{\outblock,k}))$ for the RBs with early-exits ($k \in \mathcal{K}_\outblock$), where with slight abuse of notation, $\mathbf{w}_k$ contains both the residual block parameters and the intermediate output parameters, and $\ell_k(\mathbf{x}_k, \mathbf{w}_k) =0$ for the RBs without ($k \notin \mathcal{K}_\outblock$).
The loss function of the final output is used as terminal penalty, i.e., $J_{\mathrm{f}}(\mathbf{x}_N) = \lossfunc(\outblock(\mathbf{x}_N, \mathbf{w}_{\outblock}))$.
In contrast to the fixed initial state in the OCP, the initial state for the first residual block comes from the trainable input layer, $\mathbf{x}_0 = \inblock(\mathbf{x}, \mathbf{w}_\inblock)$.
With this, we obtain a direct equivalence between the joint training formulation for early-exiting ResNets~\eqref{eq:fig_ee_training} and the OCP~\eqref{eq:fig_ocp} as is illustrated in \cref{fig:illustration_equivalence_ocp_joint_training}.

\subsection{Turnpikes in Early-Exiting ResNets}\label{sec:mainresult}
With the optimal control formulation of the joint training for early-exiting ResNets at hand, we analyze the implications of this perspective.
Since the joint training~\eqref{eq:fig_ee_training} with the cross-entropy loss might not attain its minimum, we analyze $\delta$ suboptimal solution with  $J_N(\mathbf{w}) \leq \inf_{\mathbf{w}'} J_N(\mathbf{w}') + \delta$, for a fixed $0\leq \delta < \infty$.
Crucially, our analytic findings derived this way are supported by the numerical experiments presented in \cref{sec:experiments}.
We begin by making an assumption on the loss decay of  intermediate outputs as the network depth tends to infinity.
\begin{restatable}[Exponential loss decay]{assumption}{assexpreachability}
	\label{ass:exp_reachability}
	For the given training dataset $\dataset$, there exist constants $\rho \in [0,1)$ and $\beta > 0$, and an infinite-depth ResNet with early-exits attached to each residual block, with parameters $\tilde{\mathbf{w}}_\inblock$ for the input block, $\tilde{\mathbf{w}}_k$ for the infinite sequence of residual blocks $k \in \mathbb{N}_{[0,\infty)}$, and $\tilde{\mathbf{w}}_{\outblock,k}$ for the corresponding intermediate outputs, such that the training loss of the intermediate outputs $\tilde{\mathbf{y}}_k$ satisfies
	$
		\lossfunc(\tilde{\mathbf{y}}_k) \leq \beta\rho^k,
	$
	for $k \in \mathbb{N}_{[0,\infty)}$.
\end{restatable}
Intuitively, this assumption reflects the idea that, with increasing depth the achievable loss improves.
In addition, the constants $\beta$ and $\rho$ capture the representational capacity of the network relative to the complexity of the training data: for a more challenging learning problem or a network with lower representational capacity per residual block, they become larger, reflecting that a deeper network is required to reach a given loss level.
For loss functions with finite minimizers, such as the $\ell_2$ loss, this decay assumption can be verified using well-known memorization results~\citep{hardt2017identity, pmlr-v80-nguyen18a,kim2023provable}.
However, for the cross-entropy loss this decay requires the logits to diverge which we show in Theorem~\ref{thm:exponential_decrease_cross_entropy} for fully connected ResNets with ReLU activations by extending memorization results of~\citet{hardt2017identity}. 
Hence this assumption is not overly restrictive. 
Consequently, in the following loss turnpike results, in contrast to classical turnpike results in optimal control, the outputs do not become stationary; rather, they diverge to further reduce the cross-entropy loss.
Under this assumption we obtain the following result. 
\begin{restatable}[Loss turnpike in ResNet training]{theorem}{thmturnpikeresnet}
	\label{thm:turnpike_resnet}
	Suppose that Assumption~\ref{ass:exp_reachability} holds.
	Let a ResNet~\eqref{eq:resnet} with depth $N$ be trained with stage cost loss weighted by $0 < \gamma \leq 1$ using the optimal control training formulation \eqref{eq:fig_ee_training} with the early-exit placement $\mathcal{K}_\outblock$, and let its parameters $\mathbf{w}$ be $\delta$-suboptimal for some fixed $0 \leq \delta < \infty$, i.e., $J_N(\mathbf{w}) \leq \inf_{\mathbf{w}'} J_N(\mathbf{w}') + \delta$, then the training objective is bounded independent of the depth $N$ and the early-exit placement $\mathcal{K}_\outblock$ by $$J_N(\mathbf{w}) \leq \bar{J} = \frac{\gamma\beta}{1-\rho} + \beta + \delta,$$ and the ResNet exhibits a loss turnpike on the training data.
	More precisely, for a loss level $\varepsilon>0$, consider the set of intermediate outputs with training losses exceeding $\varepsilon$,
	$
		\hat{\mathcal{Q}}_\varepsilon \doteq
		\left\{ k \in \outsetfinal \,\middle|\, 
		\lossfunc(\hat{\mathbf{y}}_k) > \varepsilon
		\right\}.
	$
	Then, for every loss level $\varepsilon>0$ there exists a constant $\kappa(\varepsilon) < \infty$, independent of the network depth $N$ and the early-exit placement $\mathcal{K}_\outblock$, such that the cardinality of $\hat{\mathcal{Q}}_\varepsilon$ is bounded
	$
		\# \hat{\mathcal{Q}}_\varepsilon \leq \kappa(\varepsilon)= \frac{\bar{J}}{\gamma\varepsilon}.
	$
\end{restatable}
\begin{proofsketch}
	By Assumption~\ref{ass:exp_reachability}, there exists an infinite-depth network with early-exits at every residual block whose training losses are bounded by the convergent sequence $\beta\rho^k$.
	For any early-exit placement $\outset \subseteq \mathbb{N}_{[0,N-1]}$, taking the first $N$ blocks and the early-exits in $\outset$ gives a feasible solution to the training OCP, whose objective, by $\outset \subseteq \mathbb{N}_{[0,\infty)}$ and non-negativity of the loss, is bounded by the corresponding geometric series.
	Hence, the limit of corresponding geometric series yields a constant upper bound $\bar{J}$ on the joint training objective, independent of the network depth $J_N(\mathbf{w}) \leq \bar{J}$. 
	By construction, each intermediate output $k \in \hat{\mathcal{Q}}_\varepsilon$ contributes at least $\gamma\varepsilon$ to the joint training objective. 
	Since the total objective is bounded by $\bar{J} = \frac{\gamma\beta}{1-\rho} + \beta {} + \delta$, it follows that the number of such intermediate outputs is finite, i.e., $\#\hat{\mathcal{Q}}_\varepsilon \leq \kappa(\varepsilon) \doteq \frac{\bar{J}}{\gamma\varepsilon}$. For the full proof we refer to  Appendix~\ref{sec:turnpike_analysis}.
\end{proofsketch}
This loss turnpike ensures that, for any prescribed loss level, only a limited forward propagation steps are needed in a jointly trained early-exit ResNet.
Consequently, each sample is correctly classified after a finite depth implying that the turnpike not only occurs in the loss but also in the accuracy, which we formalize in Theorem~\ref{thm:accuracy_turnpike_resnet}.
This result provides a theoretical explanation for the overthinking phenomenon in deep networks by showing that the depth at which a training sample enters the stable classification regime is bounded independently of the total network depth~\citep{kaya2019shallow}.

\subsection{Early-Exiting the Turnpike}\label{sec:exitingturnpike}
We now investigate the implications for early-exiting inference, where inference is terminated once an uncertainty measure falls below a threshold  \citep{Huang2017MultiScaleDN,passalis2020efficient,liu2022anytime}. 
At a block $k \in \outset$, an early-exit is taken if the uncertainty measure satisfies $\phi(\outsample_k) \leq T$.
Commonly used uncertainty measures include the entropy $\phi(\outsample_k) = \entropy(\outsample_k)$ or the maximum softmax probability complement $\phi(\outsample_k) = 1 - \hat{p}_\mathrm{max}(\outsample_k)$.
Since the loss turnpike bounds the depth required to reach any loss level $\varepsilon>0$, the softmax probabilities lie in an $\varepsilon$-neighborhood of the one-hot label, and hence any derived uncertainty measure is similarly bounded.
Based on this, we show that a jointly trained ResNet exhibiting a loss turnpike also exhibits turnpike behavior in the entropy (Theorem~\ref{thm:entropy_turnpike}) and maximum softmax probability complement (Theorem~\ref{thm:maximum_softmax_probability_turnpike}) on the training data.
Consequently, using these quantities as exit criteria yields that for any exiting threshold $T>0$, the corresponding exit depth on the training data is bounded independent of the network depth.
In addition, we show that there exists an uncertainty threshold $T>0$ that guarantees correct classification of early-exited samples.
Together, these results guarantee that both uncertainty criteria can be used to correctly exit the turnpike, avoiding overthinking on the training data without sacrificing accuracy as formulated in the following result.
\begin{theorem}[Bounded and correct early-exiting with turnpikes]
	\label{thm:turnpikes_in_early_exiting}
	Let Assumption~\ref{ass:exp_reachability} hold and consider a ResNet of depth $N$ and early-exits attached residual blocks $\outset$ trained by \eqref{eq:fig_ee_training} with stage cost weight $0 < \gamma \leq 1$, and let its parameters $\mathbf{w}$ be $\delta$-suboptimal for some fixed $0 \leq \delta < \infty$, i.e., $J_N(\mathbf{w}) \leq \inf_{\mathbf{w}'} J_N(\mathbf{w}') + \delta$.
	Then, there exists an early-exiting threshold $T>0$ and an exit depth $\kappa$ for the early-exit criteria entropy and max softmax probability complement, such that every training sample $i = 1,..., D$ is exited at latest at the $(\kappa+1)$-th exit
	$
		\# \left\{k \in \outsetfinal | \phi(\outsample_k) > T \right\} \leq \kappa,
	$
	independent of the depth $N$ and the early-exit placement.
	Consequently, the early-exit depth in terms of the number of residual blocks is bounded by
	\begin{equation*}
			k^{(i)} = \min \left( \left\{k \in \outset | \phi(\outsample_k) \leq T \right\} \cup \left\{ N \right\}\right) \leq \seq \left(\outsetfinal \right)_{\kappa+1}.
	\end{equation*}
	Additionally, if a sample exits before the final output, i.e., $ \phi(\outsample_k) \leq T$ for $k \in \outset$, it is classified correctly
	$\hat{c}(\outsample_{k}) = y^{(i)}$.
\end{theorem}
We prove this result separately for each criterion in Theorem~\ref{thm:turnpikes_in_early_exiting_entropy} (entropy) and Theorem~\ref{thm:turnpikes_in_early_exiting_max_softmax} (max softmax probability complement), respectively.
Note that if these early-exits are placed with fixed spacing $M \geq 1$, i.e., $\outset = \{0, M, 2M, \dots, N-M\}$, then the early-exiting depth of each sample $i = 1, \dots, D$ is bounded by
$
k^{(i)} = \min \left( \left\{k \in \outset | \phi(\outsample_k) \leq T \right\} \cup \left\{ N \right\}\right) \leq \kappa \cdot M,
$
independent of the depth $N$. For more discussion on the asymptotic exit placement, see Appendix~\ref{sec:asymptotic_early_exiting}. 

\section{Experiments}
\label{sec:experiments}
To underscore our turnpike analysis for early-exit networks, we conduct experiments\footnote{Code to reproduce our results is available here \href{https://github.com/OptCon/early-exit-turnpike}{https://github.com/OptCon/early-exit-turnpike}.} on standard ResNet architectures of varying depths, following the deeply supervised net of \citet{pmlr-v38-lee15a} supervising every hidden state and the partially supervised approach of \citet{kaya2019shallow}. We detail the early-exit architectures used in Section \ref{sec:architecture}.
Then, in \cref{sec:observeTurnpike} we examine whether the turnpike similarity predicted by Theorem~\ref{thm:turnpike_resnet} and Theorem~\ref{thm:accuracy_turnpike_resnet} is observable in practice for ResNets of different depth trained via gradient-based optimization.
In \cref{sec:exp_translation_inference} we analyze how the turnpike behavior empirically relates to overthinking \citep{kaya2019shallow} and how it translates to the inference time and demonstrate that the turnpike similarity from  Theorem~\ref{thm:turnpikes_in_early_exiting} in the early-exit inference.
Finally, \cref{sec:evaluation_stage_cost_weight} evaluates the influence of the stage cost loss weight and its influence of the turnpike bound from Theorem~\ref{thm:turnpike_resnet}.

\subsection{Early-Exit Architecture}
\label{sec:architecture}
Instead of having independent exit paths, we design the intermediate outputs $\outblock_k$ for the hidden state $\sample_k$ by following the computational graph through the following skip connections and finally the output layer of the neural networks
\begin{equation*}
    	\outsample_k 
    	= \outblock_k(\sample_k, \mathbf{w}_{\outblock,k}) 
    	=\outblock(\skipterm_{N-1}({\scriptsize\cdots} \skipterm_{k}(\sample_k, \mathbf{w}_{\skipterm,k}), {\scriptsize\cdots} ,  \mathbf{w}_{\skipterm,N-1}), \mathbf{w}_{\outblock}).
\end{equation*}
This architecture of exits ensures that if from some layer $k < N$ onward the parameters of all residual blocks are set to zero, the intermediate output at layer $k$ and all subsequent intermediate outputs coincide with the final network output.
Leveraging this property, we prove in Theorem~\ref{thm:turnpikes_without_leaving_arc} that the training loss of the final output exceeds the loss of any intermediate output by at most the suboptimality $\delta$.
Therefore, the leaving arc of the turnpike is bounded by the suboptimality of the training and vanishes for optimal training.

Since most skip connections are identity mappings and only at stage transitions a $1\times1$ convolution is employed \citep{10.1007/978-3-319-46493-0_38}, this design introduces only marginal overhead.
Additionally, this design naturally enables parameter sharing between the backbone and the early-exits, which we examine experimentally in Appendix~\ref{sec:early_exits_shared_parameters}.

\begin{figure*}[t]
  \centering
  \begin{subfigure}{0.49\linewidth}
    \centering
    \includegraphics[width=\columnwidth]{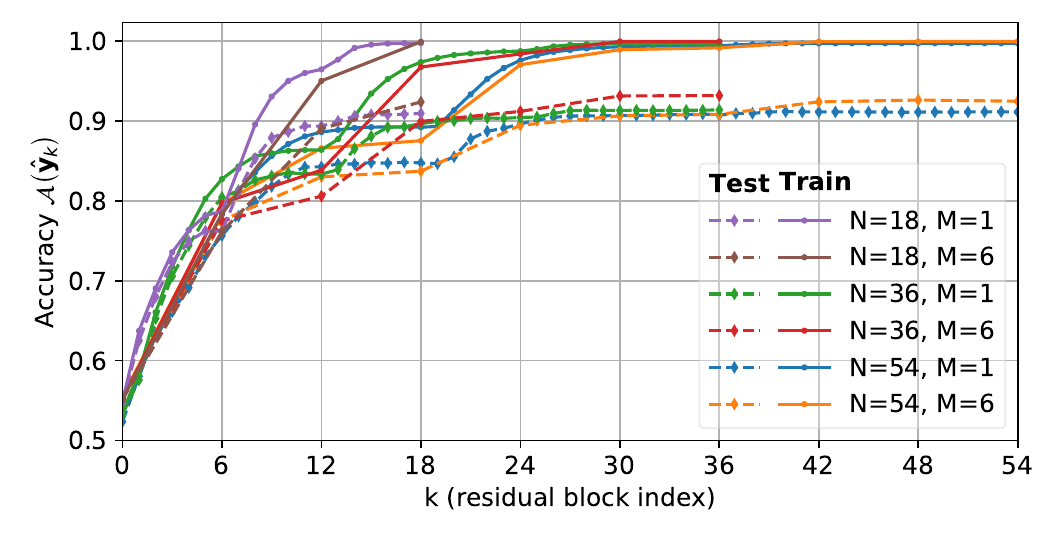}
    \caption{Accuracy trajectories.}\label{fig:loss_turnpike_acc}
  \end{subfigure}
  \hfill
  \begin{subfigure}{0.49\linewidth}
    \centering
    \includegraphics[width=\linewidth]{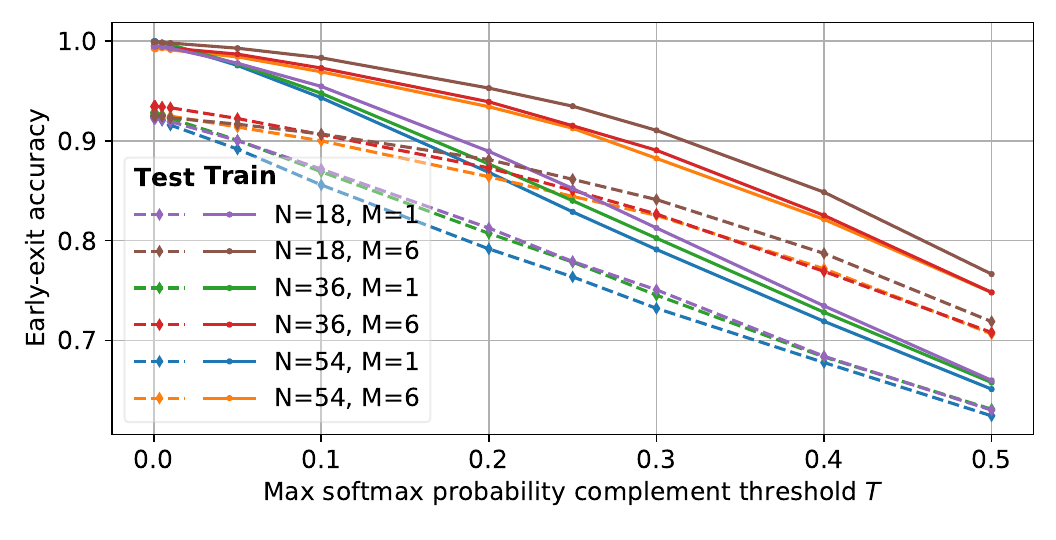}
    \caption{Early-exit accuracy over exiting thresholds.}
	\label{fig:accuracy_by_threshold}
  \end{subfigure}
  \caption{Accuracies on CIFAR-10 for ResNets with residual blocks $N \in \{18,36,54\}$ with two levels of early-exit supervision: every residual block ($M=1$) and every sixth ($M=6$).}
  \label{fig:loss_turnpike}
\end{figure*}
\subsection{Can we Observe the Turnpike?}\label{sec:observeTurnpike}
To examine the turnpike phenomenon for varying depths, we conduct experiments on CIFAR-10, CIFAR-100, and TinyImagenet (Appendix~\ref{sec:further_experiments}) \citep{krizhevsky2009learning, Deng2009} using ResNet-110 (N=54) and two shallower variants with $N=18$ and $N=36$ residual blocks as backbone architectures.
In addition, we vary the early-exit placement by attaching exits either to every residual block ($M=1$) or to every
$M=6$ residual blocks, following \citet{pmlr-v38-lee15a} and \citet{kaya2019shallow}.
The experimental setup is further outlined in Appendix~\ref{sec:experimental_setup} and we plan to release experimental code publicly for the camera-ready version.

In \cref{fig:loss_turnpike_loss}, we show the loss trajectories obtained at the respective early-exits of the trained networks and averaged over all samples.
The training losses of all networks decrease very sharply with depth and converge to near-zero values, indicating that the networks enter the interpolation regime and thereby verifying the exponential loss decrease from Assumption~\ref{ass:exp_reachability}.
Moreover, the training loss improves with the backbone depth and over the depth of the early-exits for all networks. 
The qualitatively similar training-loss trajectories observed for ResNets of different depths and early-exit placements are consistent with the loss turnpike behavior predicted by Theorem~\ref{thm:turnpike_resnet}.
The monotonic decrease of the loss implies that the corresponding trajectories do not exhibit a leaving arc, which is expected for this choice of intermediate outputs and a near optimal training by Theorem~\ref{thm:turnpikes_without_leaving_arc}.
In addition, due to the different stages, the rate at which the loss decays increase for the deeper stages, which possess more trainable parameters per residual block, leading to kinks in the plot. 
For a detailed discussion of the influence of the architecture, we refer to Appendix~\ref{sec:further_experiments}.

\cref{fig:loss_turnpike_acc} shows the corresponding accuracy trajectories. 
Within numerical precision, all models reach almost \SI{100}{\percent} accuracy at their terminal output, confirming a perfect fit to the CIFAR-10 training data. Across all backbone depths and early-exit placements, this is preceded by a steep initial rise in the first early-exits, with the accuracy turnpike reached within at most 30 residual blocks, consistent with the accuracy turnpike of Theorem~\ref{thm:accuracy_turnpike_resnet}.
Reaching this turnpike enables correct early-exit predictions; however, remaining on the turnpike without exiting constitutes wasteful overthinking.

\subsection{Turnpikes during Test-Time and Early Exiting?}
\label{sec:exp_translation_inference}

\begin{wrapfigure}{r}{0.55\columnwidth}
	\centering
	\vspace{-0.3cm}
	\includegraphics[width=\linewidth]{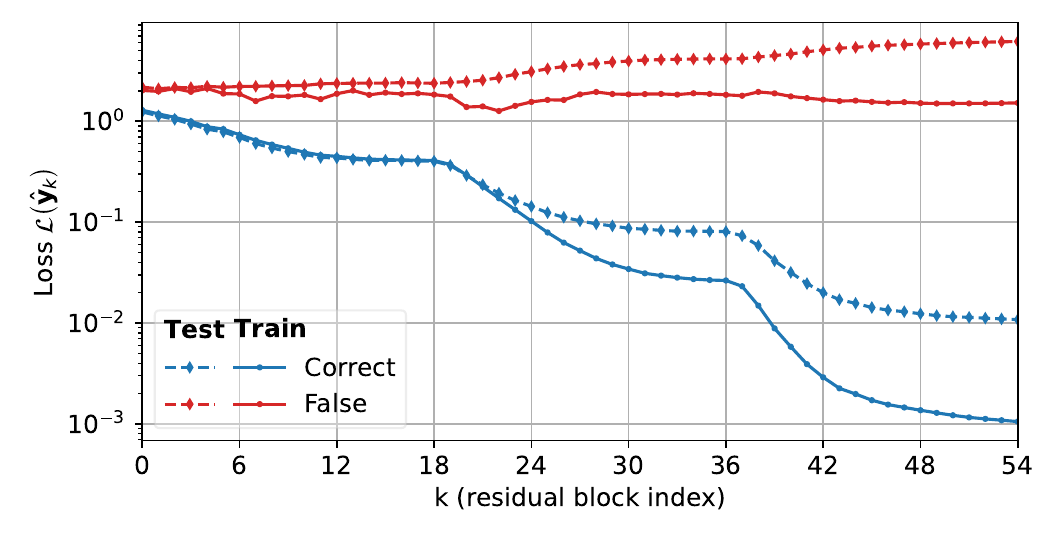}
	{\caption{Training and test loss trajectories across network depth, conditioned on the final prediction outcome on CIFAR-10 with ResNet $N=54$ and early exits placed at every block ($M = 1$).}
	\label{fig:train_test_loss}}
	\vspace{-0.2cm}
\end{wrapfigure}
To answer this questions, we take the test splits of CIFAR-10, CIFAR-100, and TinyImageNet (in Appendix~\ref{sec:further_experiments}) as in-distribution samples. 
Since our theoretical guarantees only hold for the training data, we investigate empirically how they transfer to the test data.
The test accuracies in \cref{fig:loss_turnpike_acc} show a similar convergence behavior as the training accuracies, albeit with the well-reported generalization gap of the ResNets of about $\SI{7}{\percent}$ points on CIFAR-10~\citep{10.1007/978-3-319-46493-0_38}. 
In line with the empirical findings of \citet{Huang2017MultiScaleDN}, providing the model with fewer (every $M=6$) early-exits leads to slight improvement of the test accuracy.

While the test accuracy trajectories behave similarly to their training counterparts, the test loss shows a different pattern: it initially decreases but then increases again, and none of the networks attain their minimum test loss at the final layer. As shown in \cref{fig:train_test_loss}, falsely classified test samples become increasingly overconfident in their incorrect predictions, incurring much larger loss than correctly classified samples, which saturate near zero. Hence, test samples exhibit similar turnpike behavior as their training counterparts, but for misclassified samples the turnpike drives predictions toward the wrong class with growing confidence. This overconfidence of misclassified samples may help explain why destructive overthinking occurs primarily at the first exits~\citep{kaya2019shallow}.

Building on the turnpike analysis, we now investigate its implications for early-exit inference, using the max softmax probability complement as the uncertainty-based exiting criterion.
\cref{fig:accuracy_by_threshold} shows the early-exiting accuracy over the thresholds for varying ResNet depths and early-exit placements.
As the threshold increases, and exiting is allowed at lower confidences, the accuracies decrease for all networks, strikingly, the accuracies are nearly identical across the network depths $N$, reflecting the turnpike similarity property in the exiting behavior from Theorem~\ref{thm:turnpikes_in_early_exiting}.
In addition, the decrease is more pronounced for exits at every block $M=1$ than for $M=6$, since the greater number of exit opportunities for $M=1$ favors shallower exits.
We further analyze the turnpike behavior on test data and during early-exit inference, considering the exit depth and entropy as the exit criterion, in Appendix~\ref{sec:exp_translation_inference_continued}.

\subsection{Effect of Stage-Cost Weight on the Turnpike Behavior?}
\label{sec:evaluation_stage_cost_weight}
\begin{wrapfigure}{R}{0.55\columnwidth}
	\centering
	\vspace{-0.4cm}
	\includegraphics[width=\linewidth]{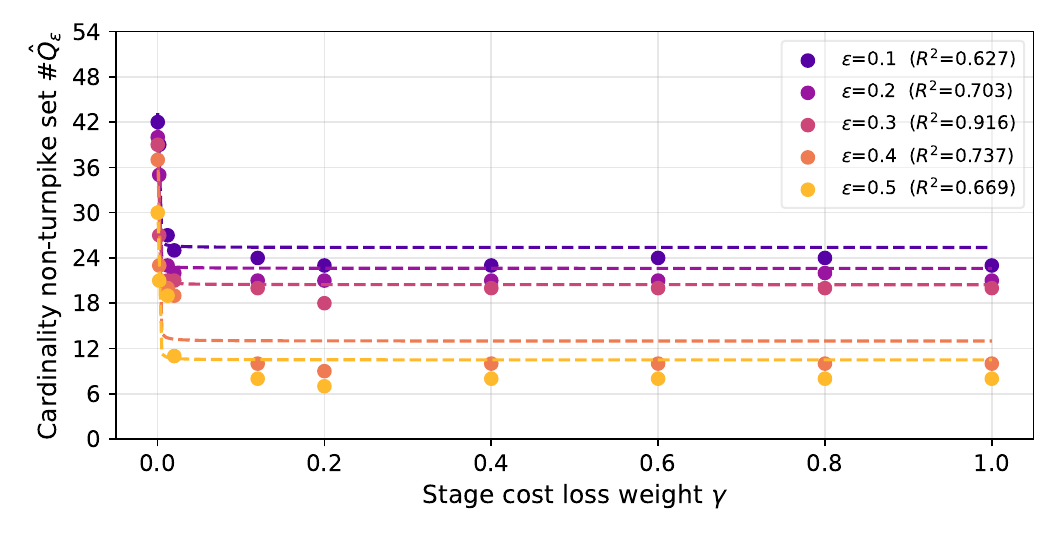}
	{\caption{Cardinality of the turnpike set $\#\hat{\mathcal{Q}}_\varepsilon$ and fitted curves $A/\gamma + B$ in dashed over the stage cost loss weight $\gamma$ for fixed loss thresholds $\varepsilon$. Results are for CIFAR-10, ResNet with $N = 54$ residual blocks and exits at every block ($M = 1$).}
	\label{fig:gamma_ablation}}
	\vspace{-0.4cm}
\end{wrapfigure}
The stage-cost loss weight $\gamma > 0$ is an additional hyperparameter in the joint training objective relative to standard training. 
It controls the trade-off between the final loss and the early-exit losses: larger values of $\gamma$ place greater emphasis on intermediate outputs, whereas smaller values focus training on the final output, yielding better final-output performance.
This relationship is also reflected in the turnpike bound from Theorem~\ref{thm:turnpike_resnet}.
In particular, the number of non-turnpike residual blocks for which the cross-entropy loss exceeds a loss threshold $\varepsilon > 0$ is bounded by
$
\#\hat{\mathcal{Q}}_\varepsilon \leq \frac{1}{\gamma\varepsilon} \left[\frac{\gamma\beta}{1-\rho} + \beta {} + \delta\right] 
= \frac{1}{\varepsilon} \left[\frac{A}{\gamma} + B\right],
$
where $A$ and $B$ absorb the model and training-dependent constants $\beta$, $\rho$ and $\delta$.
Hence, when the stage cost loss weight becomes small ($\gamma \to 0$) the number of intermediate outputs with higher loss values increases.

To investigate the effect of $\gamma$ and empirically evaluate the number of non-turnpike layers bounded by the turnpike, \cref{fig:gamma_ablation} shows $\#\hat{\mathcal{Q}}_\varepsilon$ for the varying $\gamma$ values.
Consistent with the bound, the cardinality is largely stable for larger weights ($\gamma \geq 0.2$) and grows 
sharply as $\gamma \to 0$, indicating that the constant term $B$ dominates for moderate to large weights while the $A/\gamma$ term becomes relevant only for very small $\gamma$. 
Fitting $f(\gamma) = A/\gamma + B$ at fixed $\varepsilon$ yields $R^2 \in [0.627,\, 0.916]$ confirming the $1/\gamma$ dependence, with the tightest fit at $\varepsilon = 0.3$.
We continue this discussion in Appendix~\ref{sec:gamma_ablation}.

\section{Related Work}
\paragraph{Early-Exit Neural Networks}
The field of early-exit neural networks has seen an evolution from initial deeply supervised nets \citep{pmlr-v38-lee15a} and sparse supervision \citep{branchyNet} to more advanced architecture like MSDNet \citep{Huang2017MultiScaleDN} connecting latent features across depth and scale, and SDN \citep{kaya2019shallow} leveraging intermediate classifiers to avoid overthinking, i.e., unnecessary and even destructive feature processing, and anticipate misclassifications early on. \citet{han2023dynamic} argues that poor performance of early classifiers affects the final output and hence introduces a knowledge distillation path of the final classifier to the intermediate ones. All of these works optimize the network and additional classifiers via a joint training regime and consider entropy based confidence thresholds as exit policy. Recently, \citet{regol2024jointlylearned} introduced jointly trained gating networks that decide when to exit based on a fixed early-exit backbone network. \citet{bajpai2025beyond} moved beyond static thresholds relying on multi-arm bandit policies to adapt to test time distributions. \citet{kubaty2025failure} showed that even calibrated models can lead to an insufficient compute-accuracy trade-off and suggested to focus instead on failure prediction metrics. 
These gating-based training regimes are not considered in this work as they can not be cast into our theoretical framework \citep{scardapane2020should,krzepkowski2024joint}.

\paragraph{Optimal Control for Deep Learning}
The intrinsic connections between time-continuous Neural ODEs and time-discrete ResNets lead to numerous advances both theoretically and practically. \citet{deeplearningOCJournal} studied numerical methods for deep learning models and constrains the time discretization to lie on a simplex to form sparse time steps. Understanding the forward pass as integration time horizon also lead to works aiming for minimal time neural ODEs \citep{miao2024how}. Studying asymptotic behavior of neural networks in terms of depth is crucial to guide design choices even for finite time implementations. 
Typically, the goal is to drive a dynamic system to a desired set of states, which in asymptotics, will be spent most of the time. 
These turnpikes have been investigated by \citet{ESTEVEYAGUE2023105452} for $l_1$ regularized neural ODEs and by \citet{FAULWASSER2024100290} for ResNets.
\citet{Puettschneider2026} use the label smoothing to ensure finite minimizers of the cross-entropy loss and show turnpike behavior with respect to these minimizers.
\citet{puttschneider2025towards} study the implications of the steady-state property of ResNets in settings without reachability properties.
For the cross-entropy loss and $\ell_1$ weight decay as integral regularizer, \citet{ESTEVEYAGUE2023105452} show that the hidden states converge to a stationary point representing a trade-off between the minimization of the loss and the weight decay. 
However, previous work did not explore the close relations of early-exiting and the optimal control perspective on deep learning. 
In particular, our turnpike results for entropy and early-exiting criteria appear to be the first of their kind.

\section{Conclusion}
In this paper, we analyze the joint training of early-exit ResNets taking an optimal control perspective.
To this end, we formulate their training as an optimal control problem and we link early exiting to the occurrences of the turnpike phenomenon.
Under a mild asymptotic reachability assumption, we derive new turnpike notions for the cross-entropy loss, the accuracy, and the early-exiting criteria entropy and maximum softmax probability.
These turnpike guarantees yield explicit bounds on the exit depth sufficient for correct classification of the training data, independent of the total network depth.
This provides a theoretical explanation for the success of jointly trained early-exiting networks and justify their ability to avoid overthinking through depth-bounded, confidence-driven inference.
We support these findings with experiments on CIFAR-10/100 and TinyImageNet, showing that the turnpike similarity manifests in both the hidden-state trajectories and the resulting early-exit inference behavior.

Future work should broaden the reachability assumptions to architectures beyond fully connected ResNets with ReLU activation. 
Moreover, the formal verification of these assumptions is of major interest. 
We remark that while our theoretical analysis extends to other residual-based architectures, such as vision transformers~\citep{dosovitskiy2021an}, their reachability and turnpike behavior should be investigated experimentally.
Furthermore, given the relevance of early-exiting in transformer-based large language models \citep{sui2025stop}, applying the proposed framework to such architectures is a promising direction.

Similarly, the analysis of turnpike properties for architectures without skip connections, such as VGG networks, is an interesting direction for future work, for a first step, we refer to \cite{11625228}.
In addition, while our theoretical results are centered around training performance, future research will aim to close the gap to statistical learning theory and to provide generalization guarantees. An interesting avenue in this regard is the analysis via out-of-distribution methods like \citep{imezadelajara2025mysteries}.
 
\section*{Acknowledgments}
This work has been partially funded by the Deutsche Forschungsgemeinschaft (DFG, German Research Foundation) via the TRR 391 Spatio-temporal Statistics for the Transition of Energy and Transport (project 520388526) and the research unit Active Learning for Systems and Control (ALeSCo) (project 535860958).
 
\bibliography{sources}
\bibliographystyle{abbrvnat}

\newpage
\appendix
\onecolumn
\section*{Overview of Appendix}
\begin{itemize}[noitemsep, topsep=0pt, parsep=4pt, partopsep=0pt, leftmargin=1.2em] 
    \item Appendix~\ref{sec:experimental_supplements}: Experimental setup and additional results on CIFAR-100 and TinyImageNet
    \item Appendix~\ref{sec:resnet_basics}, \ref{sec:intermediate_outputs}, \ref{sec:asymptotic_early_exiting}: Background on Early-Exit Networks and specific output design 
    \item Appendix~\ref{sec:turnpike_analysis}: Detailed proofs of main results
    \item Appendix~\ref{sec:scope_and_limitations}: Scope and Limitations of our findings
\end{itemize}

\section{Experimental Supplements}
\label{sec:experimental_supplements}

This appendix supplements the experiments of \cref{sec:experiments} with additional results and ablations. 
\cref{sec:experimental_setup} details the experimental setup. 
\cref{sec:further_experiments} extends the turnpike observations to homogeneous ResNets and additional datasets (CIFAR-100, TinyImageNet). 
\cref{sec:exp_translation_inference_continued} complements the inference-time analysis with entropy and SDN confusion score trajectories, showing how intermediate predictions progressively converge toward the final output across network depth, and extends the early-exiting results by examining exit depths for varying backbone depths, demonstrating that these also exhibit the characteristic turnpike similarity across network depths and supervision frequencies.
\cref{sec:early_exits_shared_parameters} examines parameter sharing in early-exits enabled by the proposed exit architecture. 
Finally, \cref{sec:gamma_ablation} ablates the stage-cost weight $\gamma$, which controls the relative importance of intermediate output losses.

\subsection{Setup}
\label{sec:experimental_setup}
The experimental setup is based on \citet{10.1007/978-3-319-46493-0_38}.
The networks used in our experiments use a pre-activation ResNet with $N=54$ residual blocks, with two convolutional layers per residual block, which corresponds to a ResNet-110 in \citet{10.1007/978-3-319-46493-0_38}.
The standard ResNet architecture consists of an input embedding with of a 3x3 convolution with 16 filters followed by a batch normalization and ReLU.
This is followed by $N=54$ residual blocks organized in three stages, each with 18 residual blocks.
Within these stages, The convolutional layers use 16, 32, and 64 filters, respectively.
The shallower networks with $N=36$ and $N=18$ residual blocks are adapted from this architecture by reducing the number of residual blocks in each stage to 12 and 6 residual blocks respectively.
The output block consists of a batch normalization layer, a ReLU activation, global average pooling, and a final linear layer that predicts the logits.
Similar to \citet{10.1007/978-3-319-46493-0_38}, we train for 165 epochs with a batch size of 128 using SGD with an initial learning rate of $\alpha=0.1$  and a momentum of $\mu = 0.9$. 
After 82 and 123 epochs, the learning rate is reduced by a factor of 10.

In the joint training formulations, we use a constant weight of $\gamma = 0.02$ for networks in which early-exits are attached to every residual block. 
This choice assigns approximately equal total weight to the 54 early-exits and to the final output of the ResNet-110, which is weighted by one. 
When early-exits are attached only to every sixth residual block, we accordingly increase the stage-cost weight to $\gamma = 0.12$.
The learning rate for the additional parameters associated with the intermediate outputs is set to $\alpha / \gamma$ to ensure that their effective learning rate matches that of the final output layer.
For each of these states use a separate batch norm statistics, while sharing the batch norm weights and bias for the ResNets sharing weights for early exit.

For homogeneous ResNets with constant state dimension, the number of convolutional filters is kept constant at 64 for CIFAR-10/CIFAR-100 across all residual blocks as well as the input embedding.
Apart from this modification, their architecture is identical to the standard ResNet.

All results were obtained using PyTorch (version 2.4.1) on an NVIDIA RTX A6000 (48 GB) and NVIDIA A40 (48 GB).

\subsection{Can We Observe the Turnpike for Further Architectures and Datasets? (cont.\ from \cref{sec:observeTurnpike})}
\label{sec:further_experiments}

In addition to the results on CIFAR-10 presented in the main paper, we also conduct experiments on CIFAR-100 \citep{krizhevsky2009learning} and TinyImageNet \citep{Deng2009} and consider a homogeneous ResNet architecture with a fixed number of  filters per residual block.

\begin{figure*}[t]
  \centering
  \begin{subfigure}{0.48\linewidth}
    \centering
    \includegraphics[width=\columnwidth]{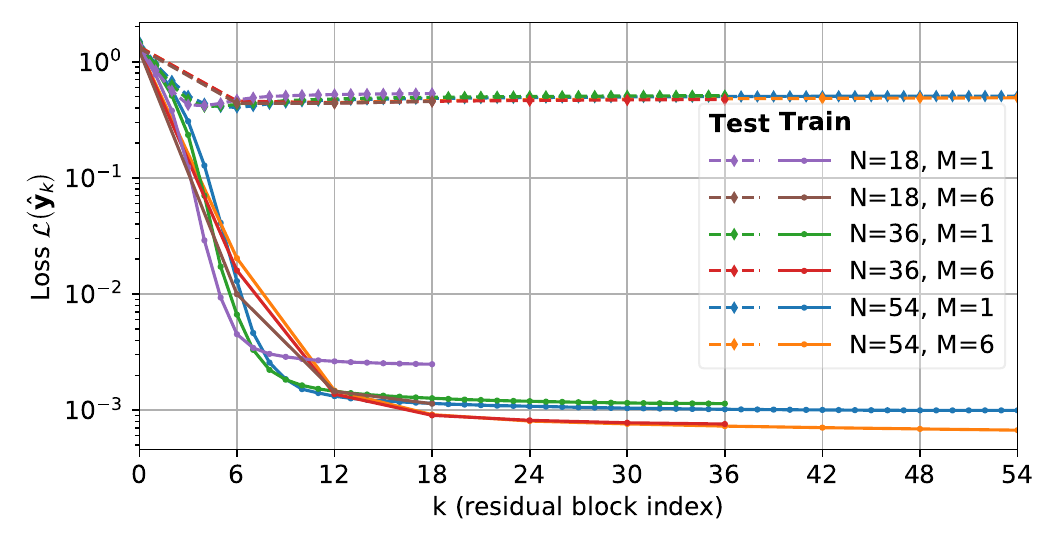}
    \caption{}\label{fig:loss_turnpike_loss_h}
  \end{subfigure}
  \hfill
  \begin{subfigure}{0.48\linewidth}
    \centering
    \includegraphics[width=\columnwidth]{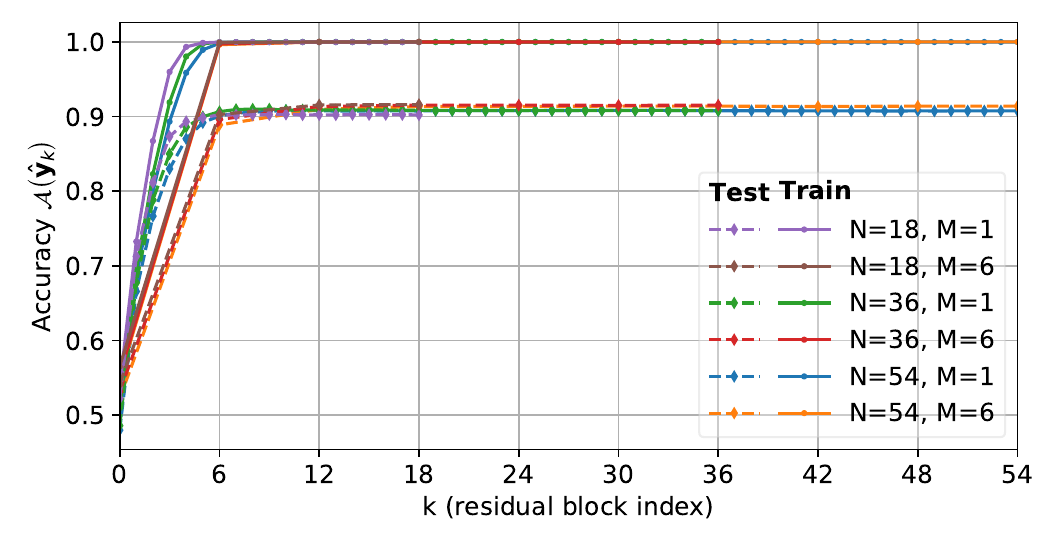}
    \caption{}\label{fig:loss_turnpike_acc_h}
  \end{subfigure}
  \caption{(a) Loss and (b) Accuracy trajectories for CIFAR-10 for the homogeneous ResNet architecture over the depth in terms of residual blocks obtained for ResNets with residual blocks $N \in \{18,36,54\}$ and with two levels of early-exit supervision: every residual block ($M=1$) and every sixth ($M=6$).}
  \label{fig:loss_turnpike_h}
\end{figure*}

\begin{figure*}[t]
  \centering
  \begin{subfigure}{0.48\linewidth}
    \centering
    \includegraphics[width=\columnwidth]{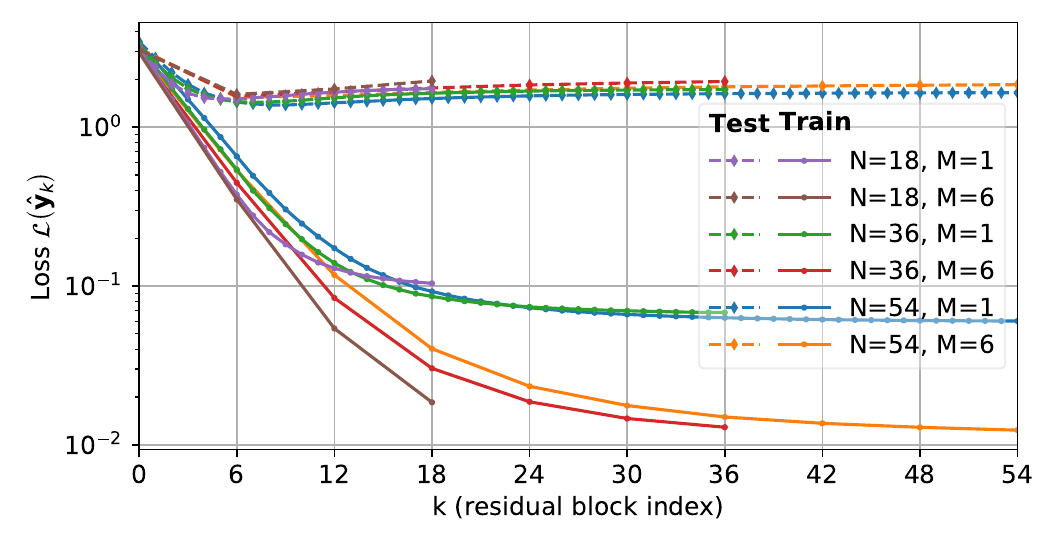}
    \caption{}\label{fig:loss_turnpike_cifar100_loss_h}
  \end{subfigure}
  \hfill
  \begin{subfigure}{0.48\linewidth}
    \centering
    \includegraphics[width=\columnwidth]{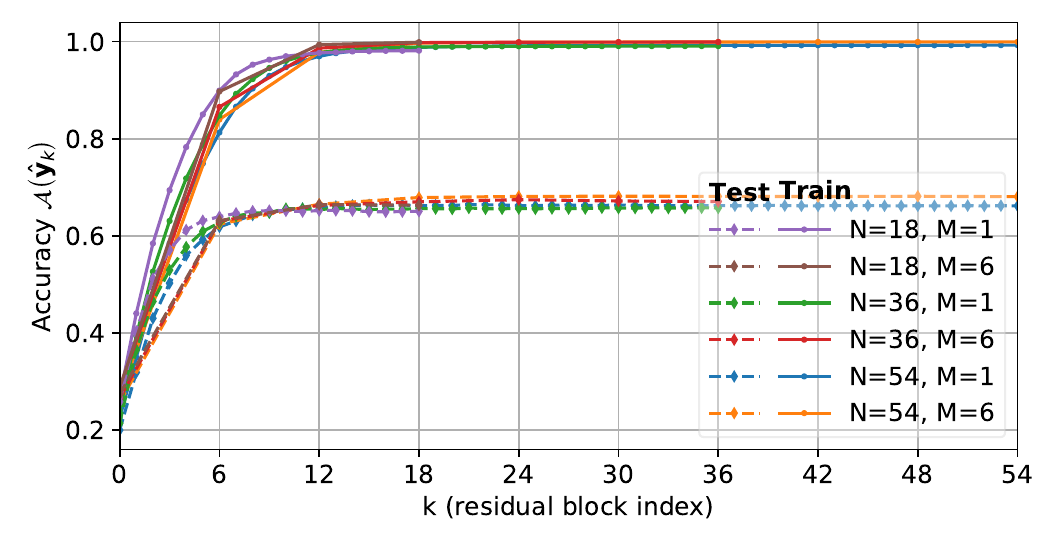}
    \caption{}\label{fig:loss_turnpike_cifar100_acc_h}
  \end{subfigure}
  \caption{(a) Loss and (b) Accuracy trajectories for CIFAR-100 for the homogeneous ResNet architecture over the depth in terms of residual blocks obtained for ResNets with residual blocks $N \in \{18,36,54\}$ and with two levels of early-exit supervision: every residual block ($M=1$) and every sixth ($M=6$).}
  \label{fig:loss_turnpike_cifar100_h}
\end{figure*}

The kinks observed in the loss and accuracy curves of the standard ResNets in \cref{fig:loss_turnpike} coincide with stage transitions, at which the feature dimension (number of channels) changes and spatial downsampling is applied. At these transitions, the number of trainable parameters per residual block increases by a factor of four, while the state dimension is reduced by a factor of two. Such abrupt changes in representational capacity likely underlie the observed non-smoothness in the training trajectories.
To isolate the effect of these stage-wise architectural changes, we also consider homogeneous ResNets with a constant number of convolutional filters, 64, across all residual blocks.
For CIFAR-10, the resulting loss and accuracy trajectories are shown in \cref{fig:loss_turnpike_h} and for CIFAR-100 in \cref{fig:loss_turnpike_cifar100_h}. 
In contrast to the standard architecture, the homogeneous networks yield markedly smoother curves without visible kinks, supporting the interpretation that the artifacts arise from stage-wise architectural variation.

Due to their increased capacity, the homogeneous ResNets reach substantially lower training losses than their standard counterparts. Interestingly, the similarity between the depth-36 and depth-54 trajectories becomes even more pronounced in the homogeneous setting. 
However, despite having more parameters compared to the standard ResNets, the homogeneous models achieve lower test accuracy, indicating poorer generalization.

\begin{figure*}[t]
  \centering
  \begin{subfigure}{0.48\linewidth}
    \centering
    \includegraphics[width=\columnwidth]{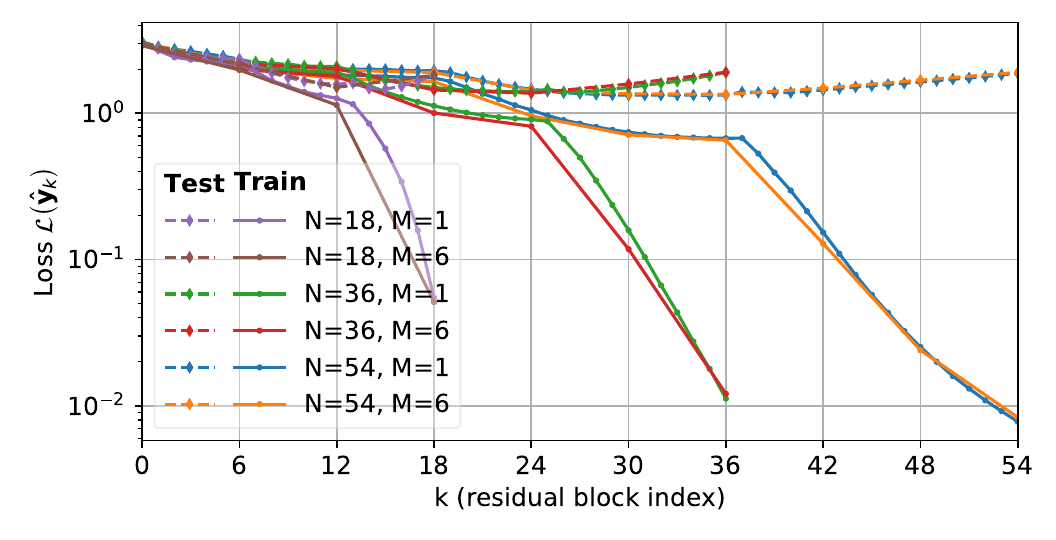}
    \caption{}\label{fig:loss_turnpike_cifar100_loss}
  \end{subfigure}
  \hfill
  \begin{subfigure}{0.48\linewidth}
    \centering
    \includegraphics[width=\columnwidth]{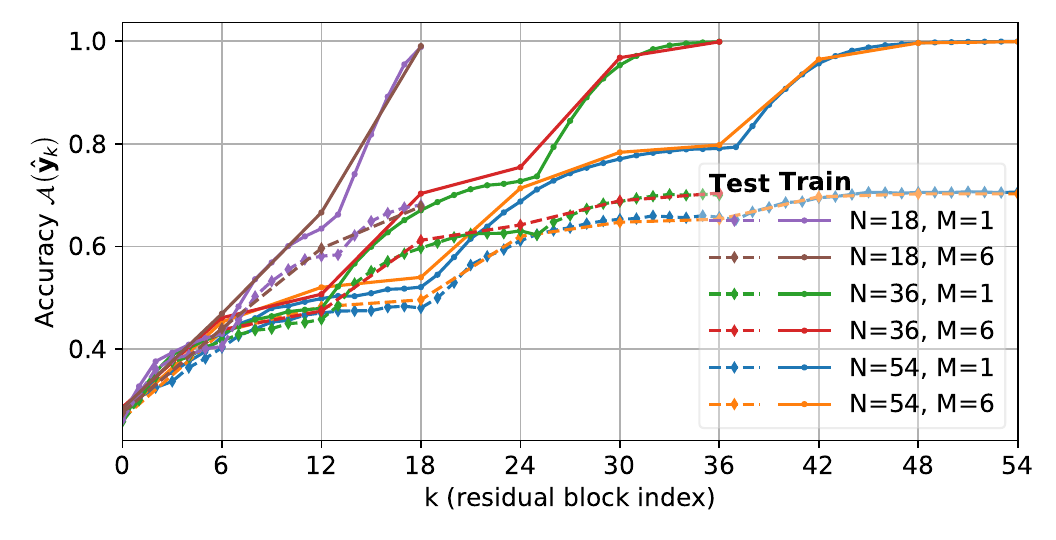}
    \caption{}\label{fig:loss_turnpike_cifar100_acc}
  \end{subfigure}
  \caption{(a) Loss and (b) Accuracy trajectories for CIFAR-100 for the standard ResNet architecture over the depth in terms of residual blocks obtained for ResNets with residual blocks $N \in \{18,36,54\}$ and with two levels of early-exit supervision: every residual block ($M=1$) and every sixth ($M=6$).}
  \label{fig:loss_turnpike_cifar100}
\end{figure*}

Figures~\ref{fig:loss_turnpike_cifar100} and \ref{fig:loss_turnpike_cifar100_h} present the loss and accuracy curves on CIFAR-100 for the standard and homogeneous ResNet architectures, respectively. 
The observed behavior closely resembles that of CIFAR-10 in both loss and accuracy. 
Owing to the larger number of classes and the increased task difficulty, CIFAR-100 yields higher training and test losses and lower accuracies than CIFAR-10, and the corresponding loss and accuracy curves converge more slowly. In particular, losses decrease and accuracies increase at a reduced rate compared to CIFAR-10. 
Nevertheless, consistent convergence patterns across network depth and early-exit configurations persist at this reduced performance level. These observations indicate that both CIFAR-10 and CIFAR-100 ResNets exhibit turnpike behavior consistent with our theoretical analysis.

\begin{figure*}[t]
	\centering
	\begin{subfigure}{0.48\linewidth}
		\centering
		\includegraphics[width=\columnwidth]{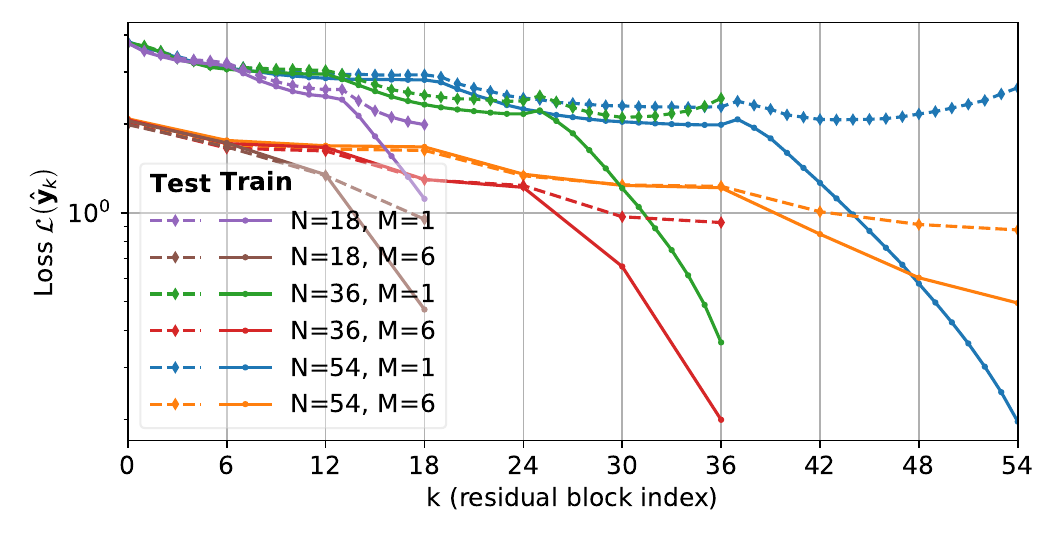}
		\caption{}\label{fig:loss_turnpike_tiny0_loss}
	\end{subfigure}
	\hfill
	\begin{subfigure}{0.48\linewidth}
		\centering
		\includegraphics[width=\columnwidth]{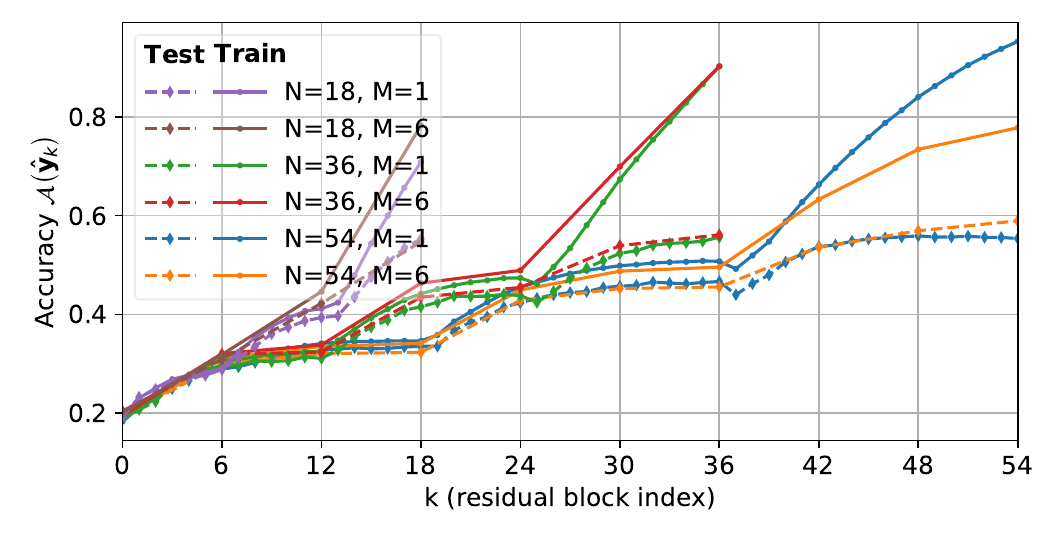}
		\caption{}\label{fig:loss_turnpike_tiny_acc}
	\end{subfigure}
	\caption{(a) Loss and (b) Accuracy trajectories for TinyImageNet for ResNet over the depth in terms of residual blocks obtained for ResNets with residual blocks $N \in \{18,36,54\}$ and with two levels of early-exit supervision: every residual block ($M=1$) and every sixth ($M=6$).}
	\label{fig:loss_turnpike_tiny}
\end{figure*}

\cref{fig:loss_turnpike_tiny} presents the loss and accuracy trajectories for the TinyImageNet dataset \citep{Deng2009}.
Unlike in the CIFAR experiments, the increased task difficulty prevents any network from reaching perfect training accuracy within the available depth, with losses remaining above $10^{-1}$ and accuracies below $\SI{95.2}{\percent}$.
This indicates that all networks are limited by their capacity and depth. Moreover, loss as well as accuracy improve contentiously over the depths of the networks. 
The loss decay rate satisfies the exponential bound required by Assumption~\ref{ass:exp_reachability} throughout all layers, and even exceeds 
it toward the final layers.
From the optimal control perspective, this implies that these depths does not suffice to fully reach the turnpike, suggesting that increasing network depth would lead to continued performance gains.
Nevertheless, the characteristic signature of the turnpike phenomenon persists, the trajectories of the training metrics from networks of different depths and supervision densities follow a markedly similar path through the intermediate layer, before diverging near the final blocks.
This divergence near the end reflects a more fundamental issue: due to the limited capacity of these networks, the joint training formulation in early-exiting imposes a much stronger trade-off between intermediate outputs and the final prediction than in the CIFAR setting, ultimately constraining the performance of the final classifier.

Despite the larger gaps in training performance, the test accuracies are remarkably similar across networks of varying depth and supervision frequencies and exhibit a much stronger turnpike behavior than the corresponding training trajectories.
In addition, due to the limited capacity of the network, the trade-off between intermediate output and final classifier of the joint training introduces a notable gap between the final test accuracies. 
The ResNets-110 with $N=54$ residual blocks and early-exits at every block ($M=1$) reaches \SI{58.3}{\percent} test accuracy and the network with supervision at every sixth block ($M=6$) reaches $\SI{55.9}{\percent}$, reflecting a stronger deterioration compared to the \SI{62.4}{\percent} accuracy reported for the ResNet-110 without early-exits in~\citet[Table 3]{Liu2022}.

\subsection{Turnpikes during Test-Time and Early Exiting? (cont.\ from \cref{sec:exp_translation_inference})}
\label{sec:exp_translation_inference_continued}

In this section, we further analyze the turnpike behavior on test data and during early-exit inference, considering the exit depth and entropy as the exit criterion.

\begin{figure*}[t]
	\centering
	\begin{subfigure}{0.48\linewidth}
		\centering
		\includegraphics[width=\columnwidth]{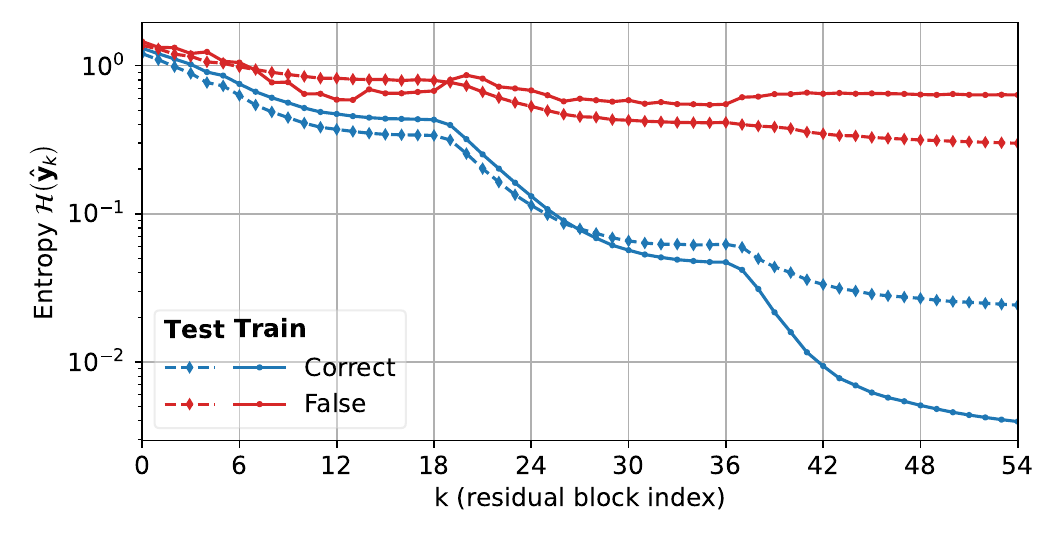}
		\caption{}\label{fig:train_test_entropy}
	\end{subfigure}
	\hfill
	\begin{subfigure}{0.48\linewidth}
		\centering
		\includegraphics[width=\columnwidth]{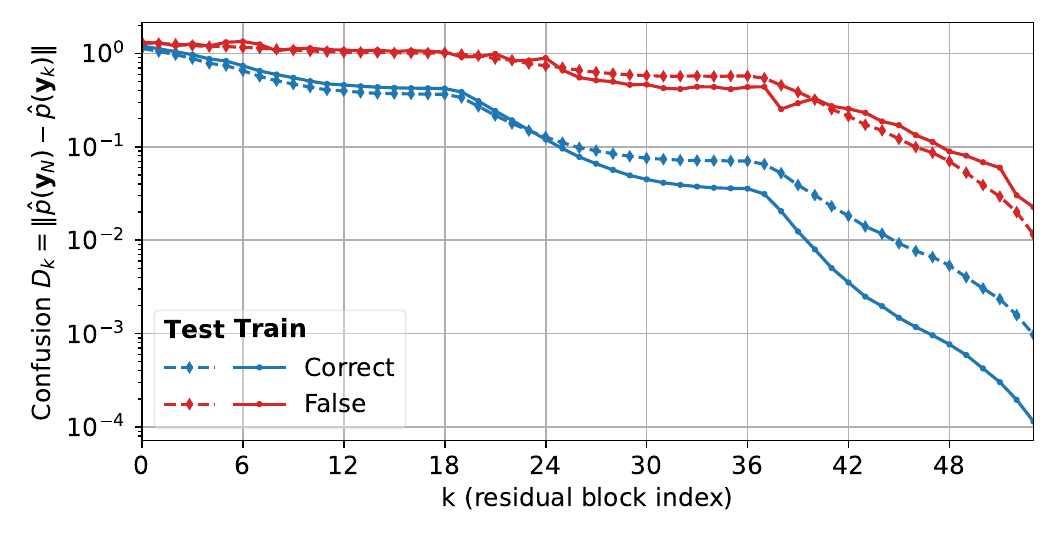}
		\caption{}\label{fig:train_test_sdn_confusion}
	\end{subfigure}
	\caption{(a) Entropy and (b) SDN confusion for training and test data across network depth, conditioned on the final prediction outcome on CIFAR-10 with ResNet $N=54,M=1$.}
	\label{fig:train_test_data_by_classification}
\end{figure*}

We begin by examining the convergence of entropy for test samples.
In \cref{fig:train_test_entropy}, and similar to \cref{fig:train_test_loss}, we plot the average entropy of samples conditioned on their final prediction outcome for the CIFAR-10 dataset split by training and test samples.
The plot shows that the entropy consistently decreases for both the training and test samples, which implies that all samples that are finally classified correctly become increasingly confident in their final correct classification.
Thus, these samples are amenable to early-exit by an entropy threshold, thereby reducing the computational demand for inference.
The incorrectly classified samples exhibit on average a higher entropy values, implying that they do not reach the confidence threshold required for early-exiting.
Thus, a well tuned early-exiting threshold can early-exit the correctly classified samples and avoid overthinking of samples that can be classified correctly before the final layer~\citep{kaya2019shallow}.

To quantify this overthinking effect, \citet{kaya2019shallow} introduced a confusion score measuring the difference in predicted softmax probabilities between intermediate exits and the final layer. \cref{fig:train_test_sdn_confusion} plots the corresponding trajectories, which shows the convergence of the intermediate predictions toward the final prediction, consistent with the turnpike phenomenon. 
This provides an explanation for why overthinking primarily manifests between the earlier intermediate outputs and the deeper layers, samples are already committed to their final prediction well before the terminal layer is reached.

\begin{figure*}[h]
	\centering
	\begin{subfigure}{0.48\linewidth}
		\centering
		\leavevmode\vbox{\offinterlineskip
			\hbox{\includegraphics[width=\linewidth, trim=0 31.2 0 0, clip]{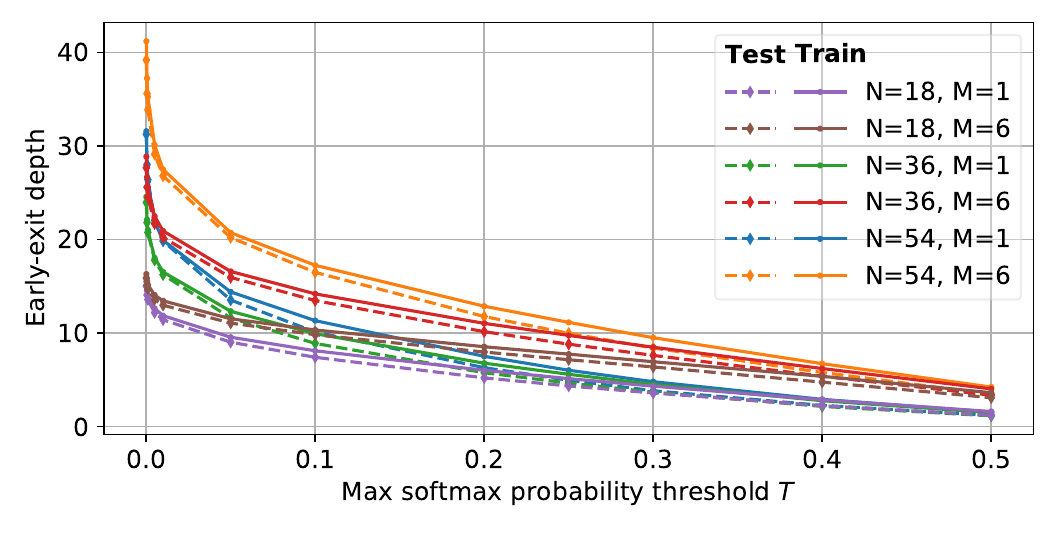}}
			\hbox{\includegraphics[width=\linewidth, trim=0 0 0 228, clip]{figure/plots_aggregated/cifar10_every1_6/aggregated_cifar10_Early_Exit_Every-1_6_accuracy__combined_full_max_softmax.pdf}}}
		\vspace{-0.7cm}
		\caption{Early-exit depths for varying thresholds}\label{fig:depth_by_threshold}
	\end{subfigure}
	\hfill
	\begin{subfigure}{0.48\linewidth}
		\centering
		\includegraphics[width=\linewidth]{figure/plots_aggregated/cifar10_every1_6/aggregated_cifar10_Early_Exit_Every-1_6_accuracy__combined_full_max_softmax.pdf}
		\vspace{-0.7cm}
		\caption{Early-exit accuracies for varying thresholds.}
		\label{fig:accuracy_by_threshold_2}
	\end{subfigure}
	\caption{Early-exiting results with the max softmax probability complement as exiting criterion on CIFAR-10 for ResNets with residual blocks $N \in \{18,36,54\}$ with two levels of early-exit supervision: every residual block ($M=1$) and every sixth ($M=6$).}
	\label{fig:depth_accuracy_by_threshold}
\end{figure*}

We now analyze the turnpike behavior of the early-exit depths, complementing our previous analysis of early-exit accuracies. \cref{fig:depth_accuracy_by_threshold} shows the exit depths and accuracies on CIFAR-10 using the max softmax probability complement as the exit criterion.
Most notably, however, the exit depth for low thresholds start close to the full backbone depths, with increasing exiting thresholds, they rapidly  decrease and converge toward one another. 
Notably, the trade-off between accuracy and computational cost is clearly visible when comparing the supervision levels. 
Networks with exits placed every $M=6$ blocks are higher than their counterparts with exits at every block $M=1$, the exit depth show that this comes at the price of an increased early-exiting depth and therefore higher computational demand for the inference.
Notably, the trade-off between accuracy and computational cost is clearly visible when comparing supervision frequencies. Networks with exits placed every $M=6$ residual blocks achieve higher early-exit accuracy than their $M=1$ counterparts, but at the cost of greater average exit depth, implying higher computational demand at inference time. 
In addition, the early-exit depths for training and test data are virtually indistinguishable, which further supports the generalization of the turnpike property to the inference time.

\begin{figure*}[h]
	\centering
	\begin{subfigure}{0.48\linewidth}
		\centering
		\includegraphics[width=\linewidth]{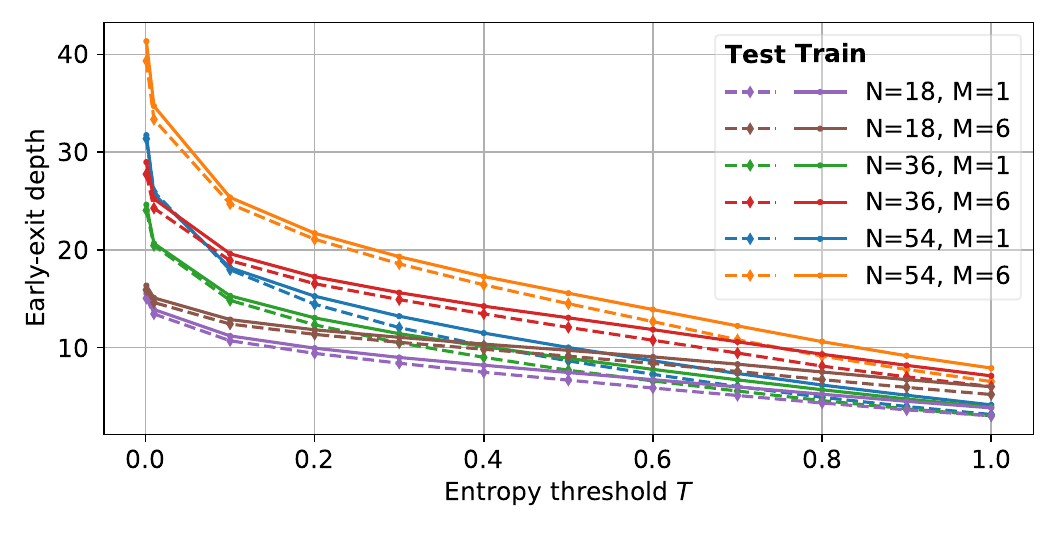}
		\vspace{-0.7cm}
		\caption{Early-exit depths for varying thresholds}\label{fig:depth_by_threshold_entropy}
	\end{subfigure}
	\hfill
	\begin{subfigure}{0.48\linewidth}
		\centering
		\includegraphics[width=\linewidth]{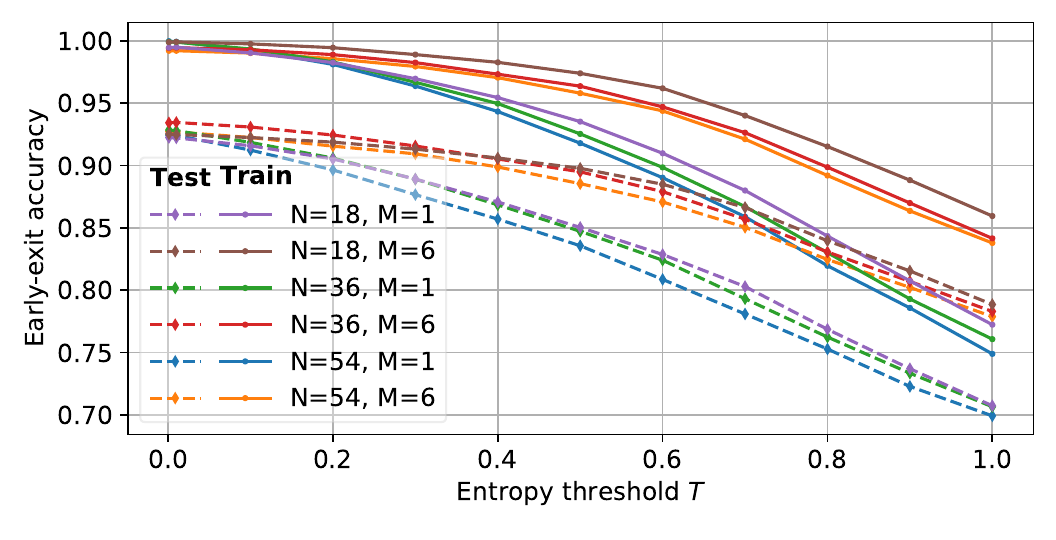}
		\vspace{-0.7cm}
		\caption{Early-exit accuracies for varying thresholds.}
		\label{fig:accuracy_by_threshold_2_entropy}
	\end{subfigure}
	\caption{Early-exiting results with the entropy as exiting criterion on CIFAR-10 for ResNets with residual blocks $N \in \{18,36,54\}$ with two levels of early-exit supervision: every residual block ($M=1$) and every sixth ($M=6$).}
	\label{fig:depth_accuracy_by_threshold_entropy}
\end{figure*}

Early-exiting results for the entropy as exiting criterion are show in \cref{fig:depth_accuracy_by_threshold_entropy}, which show the same turnpike behavior as the max softmax probability criterion.
From an optimal control perspective, both criteria robustly detect the turnpike at inference time. 
As the turnpike corresponds to the overthinking regime, this enables early exits upon reaching it, avoiding unnecessary computation.

\subsection{Shared Early-Exit Parameters}
\label{sec:early_exits_shared_parameters}
The early-exit architecture
\begin{equation*}
	\label{eq:proposed_early_exit}
	\begin{split}
		\outsample_k 
		= \outblock_k(\sample_k, \mathbf{w}_{\outblock,k}) 
		=\outblock(\skipterm_{N-1}({\scriptsize\cdots} \skipterm_{k}(\sample_k, \mathbf{w}_{\skipterm,k}), {\scriptsize\cdots} ,  \mathbf{w}_{\skipterm,N-1}), \mathbf{w}_{\outblock})
	\end{split}
\end{equation*}
admits two design choices regarding parameterization. 
Since the layers used by the early-exits are already part of the backbone ResNet, the corresponding parameters can either be shared with the backbone or instantiated separately for each early-exit. 
We note that reusing parameters of later skip connections in the stage cost of earlier residual blocks formally breaks the causal structure of the OCP, though this does not affect the training objective and its analysis, which is the focus of this work.
Sharing the early-exit parameters with the backbone network allows for earlier exiting without introducing additional trainable parameters.
However this comes at the cost of reduced model capacity.
\begin{figure*}[t]
	\centering
	\begin{subfigure}{0.48\linewidth}
		\centering
		\includegraphics[width=\columnwidth]{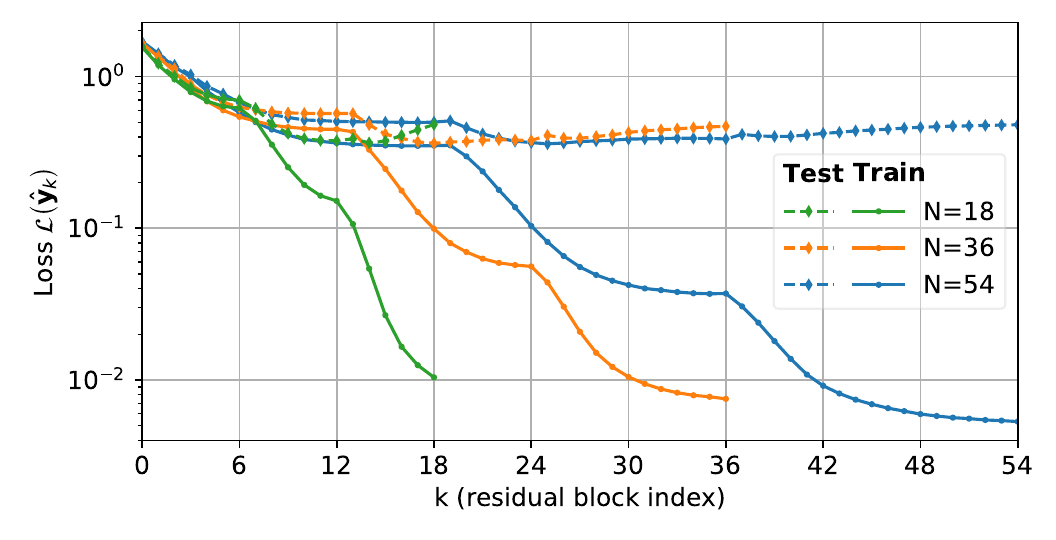}
		\caption{}\label{fig:loss_turnpike_cifar10_loss_sp}
	\end{subfigure}
	\hfill
	\begin{subfigure}{0.48\linewidth}
		\centering
		\includegraphics[width=\columnwidth]{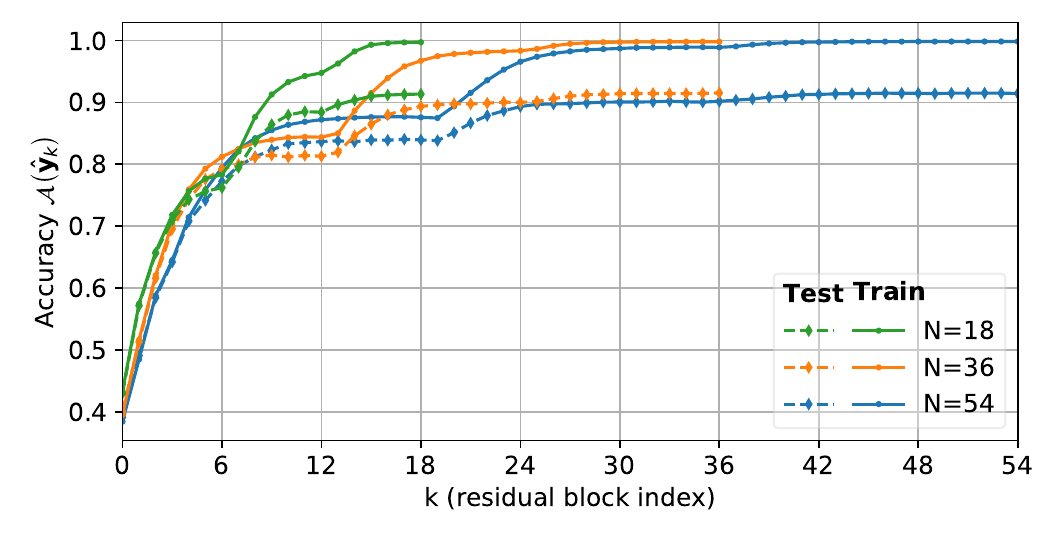}
		\caption{}\label{fig:loss_turnpike_cifar10_acc_sp}
	\end{subfigure}
	\caption{(a) Loss and (b) Accuracy trajectories for CIFAR-10 for the standard ResNet architecture with shared early-exit parameters over the depth in terms of residual blocks obtained for ResNets with residual blocks $N \in \{18,36,54\}$ and early-exits at every residual block ($M=1$).}
	\label{fig:loss_turnpike_cifar10_sp}
\end{figure*}
\cref{fig:loss_turnpike_cifar10_sp} shows the loss and accuracy trajectories on CIFAR-10 for the standard ResNet when parameters are shared between the backbone and the early-exits. 
While the training loss is comparable to that obtained with separate early-exit parameters as presented in the main \cref{fig:loss_turnpike_acc}, the test accuracy is significantly lower. 
This degradation in performance is consistent with the reduced model capacity induced by parameter sharing and indicates inferior generalization compared to the separate-parameter configuration.

\subsection{Effect of Stage-Cost Weight on the Turnpike Behavior? (cont. from \cref{sec:evaluation_stage_cost_weight})}
\label{sec:gamma_ablation}
We continue our ablation of the stage-cost weight from Section~\ref{sec:evaluation_stage_cost_weight}
using the same networks as before, ResNets with $N=54$ residual blocks and early-exits at every block $M=1$ trained using varying stage cost weights $\gamma \in \left\{0.0002, 0.002, 0.012, 0.02, 0.12, 0.2, 0.4, 0.6, 0.8, 1.0\right\}$.

\begin{figure*}[t]
	\centering
	\begin{subfigure}{0.48\linewidth}
		\centering
		\includegraphics[width=\columnwidth]{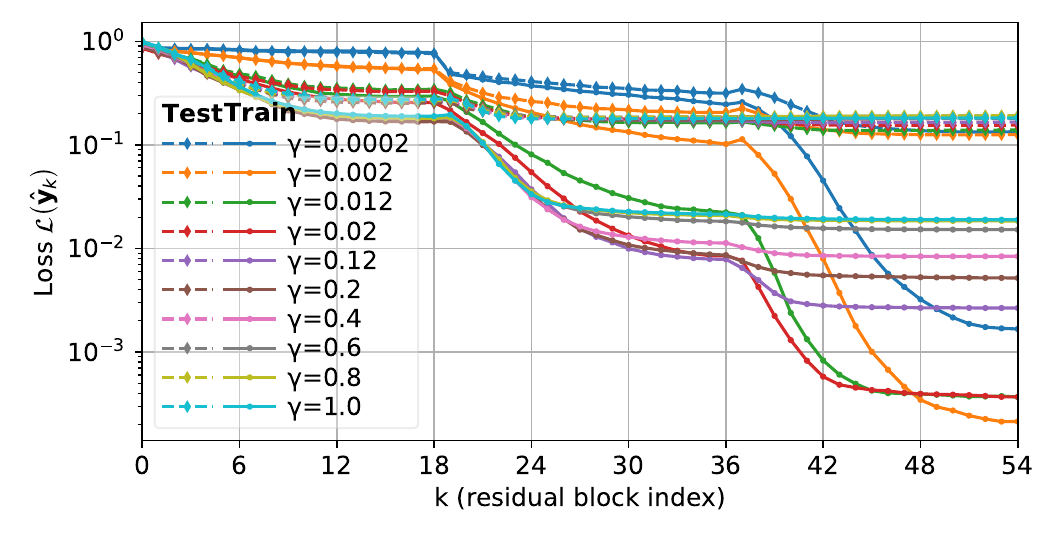}
		\caption{}\label{fig:loss_gamma_cifar10_loss}
	\end{subfigure}
	\hfill
	\begin{subfigure}{0.48\linewidth}
		\centering
		\includegraphics[width=\columnwidth]{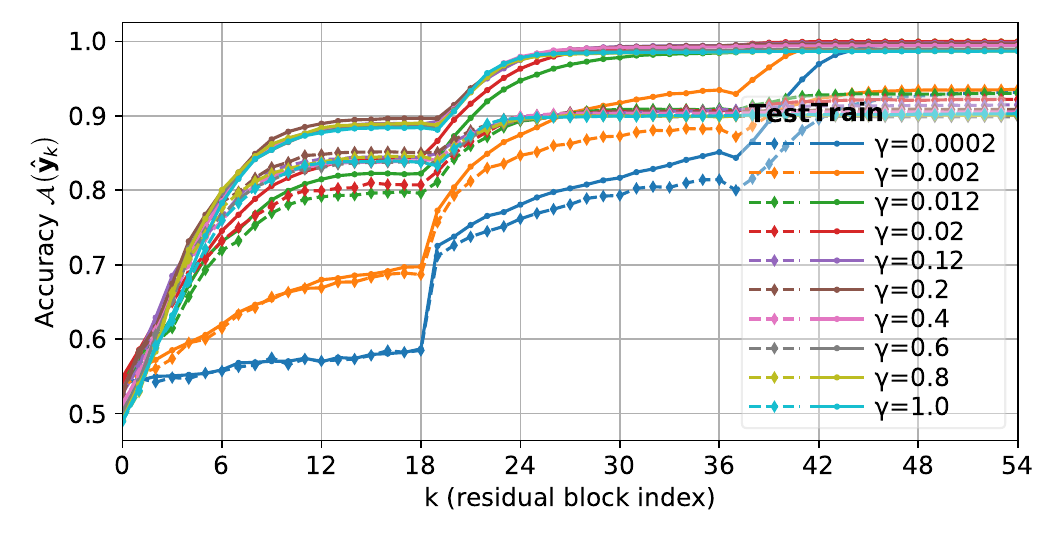}
		\caption{}\label{fig:loss_gamma_cifar10_acc}
	\end{subfigure}
	\caption{(a) Loss and (b) Accuracy trajectories for CIFAR-10 for the ResNet over the depth in terms of residual blocks obtained for ResNets with $N=54$ residual blocks trained with varying stage-cost weights and early-exits at every residual block ($M=1$).}
	\label{fig:loss_gamma_cifar10}
\end{figure*}

The resulting accuracies and loss curves in \cref{fig:loss_gamma_cifar10} show the trade-off between the intermediate outputs and the final classifier.
For weights $\gamma \geq 0.2$, the intermediate curves cluster tightly across early layers, but final-block performance deteriorates noticeably.
At the other extreme, small weights $\gamma \leq 0.002$ fail to achieve a meaningful convergence in the earlier layers, making them unsuitable for early-exiting.
The default values $\gamma=0.02$ used in our experiments achieves a good balance between intermediate and final-output performance.

\section{ResNet Basics}
\label{sec:resnet_basics}
We briefly revisit the core components of the ResNet-v2 architecture \citep{10.1007/978-3-319-46493-0_38,he2016deep} which is considered here.

\begin{figure}[h]
	\newif\ifincludestagecost
	\includestagecosttrue
	\input{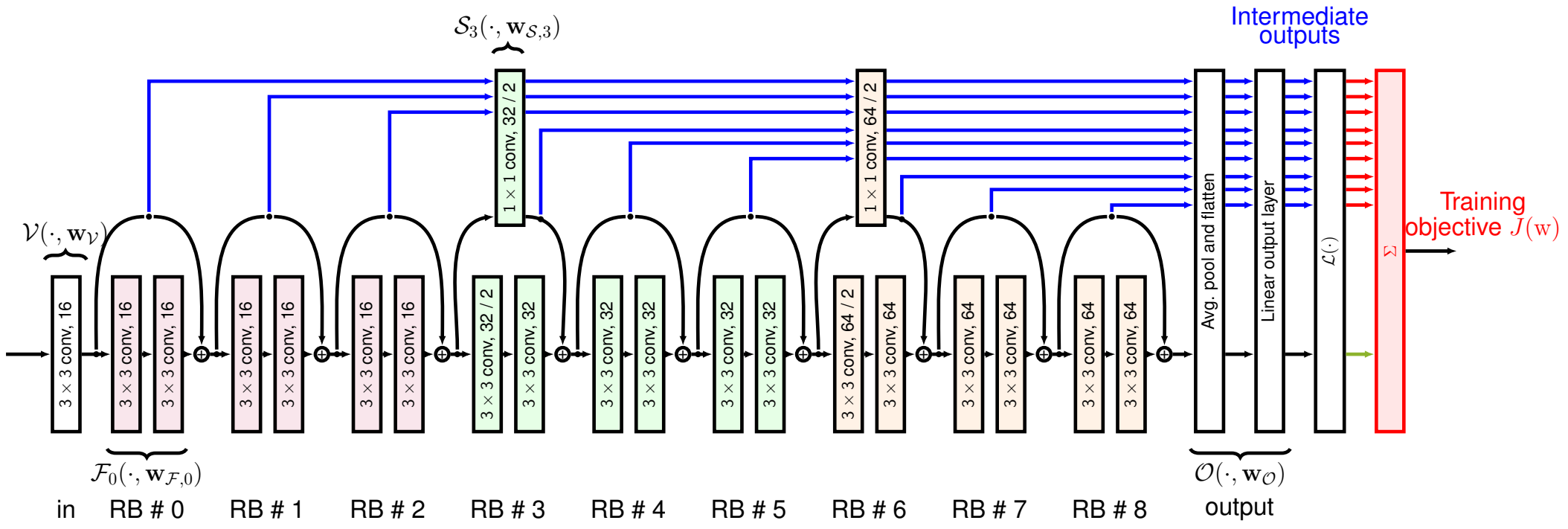}
	\caption{Illustration of a ResNet-v2 architecture while training with our proposed stage cost regularizer penalizing intermediate outputs. We propose to forward the hidden states through the network's non-trivial skip connections to account for dimensionality projections.}
	\label{fig:resnet_cifar}
\end{figure}

As illustrated in \cref{fig:resnet_cifar}, in the forward pass a sample $\sample$ is initially embedded into a latent space via $\inblock(\sample, \mathbf{w}_\inblock)$, then $N$ residual blocks process the input to eventually forward the final representation $\sample_N$ to the output head $\outblock(\sample_N, \mathbf{w}_\outblock)$.
We denote the trainable weights as $\mathbf{w} = \left[ \mathbf{w}_{\inblock},\mathbf{w}_{0},...,\mathbf{w}_{N-1}, \mathbf{w}_\outblock \right]$, where $\mathbf{w}_k = \left[\mathbf{w}_{\resterm,k},\mathbf{w}_{\skipterm,k}\right]$ corresponds to the $k$th residual block $f_k(\sample_k, \mathbf{w}_{k})$ that is composed of the convolutional layers $\mathcal{F}_k(\sample_k, \mathbf{w}_{F,k})$ and the skip connection $\skipterm_k (\sample_k, \mathbf{w}_{\skipterm,k})$ with optional parameters $\mathbf{w}_{\skipterm,k}$. If two consecutive blocks have the same dimension, an identity skip connection is employed, but if there is a change in dimensionality, a 1x1 convolution is used as a projection layer. For the ease of reading, we denote $\skiplayerset$ as the index set of blocks having non-trivial skip connections.
Overall, a ResNet with $N$ residual blocks comprises the following system equations:
\begin{equation}
	\label{eq:simple_resnet_dynamics}
	\begin{aligned}
		\sample_0 &= \inblock(\sample, \mathbf{w}_\inblock) \enspace, \\	 
		\sample_{k+1} 
		&= f_k(\sample_k, \mathbf{w}_k) =  \begin{cases}
			\sample_k &+ \resterm_k(\sample_k, \mathbf{w}_{\resterm,k}), \; k \notin \skiplayerset \\
			\skipterm_k (\sample_k, \mathbf{w}_{\skipterm,k}) &+ \resterm_k(\sample_k, \mathbf{w}_{\resterm,k}), \; k \in \skiplayerset, 
		\end{cases}
		\; &\forall k = 0, ..., N&-1 \enspace,\\
		\outsample &= \outblock(\sample_N, \mathbf{w}_\outblock) \enspace.
	\end{aligned}
\end{equation}

\subsection{Joint Training of Early-Exit ResNets}

In this work, we consider standard ResNet-v2 \citep{10.1007/978-3-319-46493-0_38}
trained on classification datasets consisting of $D$ samples of features $\sample$ and labels $y^{(i)}$
\begin{equation*}
	\dataset = \left\{\left(x^{(1)}, y^{(1)}\right), ..., \left(x^{(D)}, y^{(D)}\right)\right\}
\end{equation*}
with $C$ classes.

For training the ResNet on the entire dataset, we denote the propagation of all samples by boldface variables, e.g., $\mathbf{x} = \left[x^{(1)}, x^{(2)}, \ldots, x^{(D)}\right]$.
Assuming a constant state dimension in the hidden residual blocks, the propagation of the data in the residual blocks can be understood as a time-varying dynamical system: $\mathbf{x}_{k+1} = f_k(\mathbf{x}_k, \mathbf{w}_k) = \mathbf{x}_k + \resterm_k(\mathbf{x}_k, \mathbf{w}_{\resterm,k})$, with stacked data $\mathbf{x}_k$ as state variable and the trainable parameters $\mathbf{w}_k$ as control inputs.
The objective function of standard neural network training is the loss $\lossfunc$ evaluated at the network output, which corresponds to the terminal penalty in optimal control. 
In addition to common weight decay, optimal control approaches to ResNet training have suggested including a stage cost regularizer $\ell_k(\mathbf{x}_k, \mathbf{w}_k)$ that depends also on the states $\mathbf{x}_k$ \citep{esteve2021largetime, FAULWASSER2024100290, miao2024how}. This leads to the training objective 
\begin{equation*}
	J(\mathbf{w})=  \sum_{k = 0}^{N-1}  \ell_k(\mathbf{x}_k, \mathbf{w}_k) + \lossfunc (\hat{\mathbf{y}}).
\end{equation*}
This corresponds to the typical joint training formulations for the training of early-exits \citep{branchyNet, 10.1145/3527155}
\begin{equation*}
	\inf_{\mathbf{w}} \; J_N(\mathbf{w})=  \sum_{k\in \outset}\gamma \lossfunc (\hat{\mathbf{y}}_k) + \lossfunc (\hat{\mathbf{y}}) \quad  \text{subject to} \quad \text{\eqref{eq:simple_resnet_dynamics}, \eqref{eq:proposed_early_exit}},  \; \forall k = 0, ..., N-1,
\end{equation*}
if the stage cost for residual blocks with an early-exits defined as the loss of the intermediate prediction weighted by $\gamma > 0$,
$\ell_k(\mathbf{x}_k, \mathbf{w}_k) = \gamma \lossfunc(\hat{\mathbf{y}}_k)$, $k \in \outset$, and $\ell_k(\mathbf{x}_k, \mathbf{w}_k) = 0$ for residual blocks without early-exits.

For classification tasks, we first compute probabilities for each class using the softmax activation function from the output $\outsample$.
This gives a probability vector $\hat{p}(\hat{y}) \in (0,1)^C$, which is element-wise defined as
\begin{equation*}
	[\hat{p}(\hat{y})]_c = 
	\frac{
		e^{[\hat{y}]_c}
	}{
		\sum_{i = 1}^{C}	e^{[\hat{y}]_i},
	}
\end{equation*}
where by $\left[x\right]_j$ denotes the $j$-th element of the vector $x$.
With the softmax probability vector, we compute the cross-entropy loss for the label $y$ by taking the negative logarithm of the softmax probability for the class $y$
\begin{equation*}
	\lossfunc(\hat{y}) = -\log \left([\hat{p}(\hat{y})]_y\right).
\end{equation*}
Here, we omit the ground-truth label as an additional argument of the loss function.
For the full dataset, we compute the empirical loss by averaging the loss over the dataset
\begin{equation*}
	\lossfunc(\hat{\mathbf{y}}) = \frac{1}{D}\sum_{i = 1}^{D} \lossfunc(\outsample).
\end{equation*}

The entropy of the softmax probability distribution reads
\begin{equation*}
	\entropy(\hat{y}) =  -\sum_{c=1}^{C} \left[\hat{p}(\hat{y})\right]_c \log \left(\left[\hat{p}(\hat{y})\right]_c\right).
\end{equation*}
Similar to the cross-entropy loss, the entropy $\entropy$ of the output for all data samples $\hat{\mathbf{y}}$ is computed as
\begin{equation*}
	\entropy(\hat{\mathbf{y}}) =  \frac{1}{D}\sum_{i = 1}^{D} \entropy(\outsample).
\end{equation*}
The class with the maximum softmax probability gives the class prediction
\begin{equation*}
	\hat{c}(\hat{y}) = \argmax_c \left[\hat{p}(\hat{y})\right]_c.
\end{equation*}
For binary classification, $C = 2$, we adopt the convention that ties are broken in favor of the true class. Since the set of logits inducing a tie, i.e.\ $\{\hat{y}  : [\hat{p}(\hat{y})]_1 = [\hat{p}(\hat{y})]_2\} \subset \mathbb{R}^2$, is a null set with respect to the Lebesgue measure on $\mathbb{R}^2$, this convention has no effect for continuously distributed logits in practice and is merely introduced to make the threshold results in the following section well-defined at the boundary.
With this class prediction, we can compute the classification accuracy of the network over the entire dataset
\begin{equation*}
	\mathcal{A}(\hat{\mathbf{y}}) = 
	\frac{
		\# \left\{i = 1,..., D \middle| \hat{c}(\outsample) = y^{(i)} \right\}	
	}{
		D
	}.
\end{equation*}
For the early-exit placement and the final output $\outsetfinal = \{k_1, \dots, k_m\}$ with $k_1 < k_2 < \dots < k_m = N$, we denote by $\seq\left(\outsetfinal\right)_j = k_{\min\{j,\, m\}}$ the index of the residual block with the $j$-th output, truncated at the final block $N$ if $j > m$.

\section{Intermediate Output Architecture}
\label{sec:intermediate_outputs}
In early-exiting, intermediate outputs must remain shallow for efficiency. 
ResNets provide a natural mechanism for this: if the parameters of the residual branches for $M \leq k \leq N-1$ are set to zero $\mathbf{w}_{\resterm,k}=0$, then the residual blocks are zero $\resterm_k(\sample_k) = 0$ for all samples $\sample$. 
Such that the hidden state $\sample_M$ is forwarded only by the skip connections of the following residual blocks, resulting in the output
\begin{equation*}
	\outsample = \outblock(\skipterm_{N-1}(\cdots \skipterm_{M}(\sample_M, \mathbf{w}_{\skipterm,M}), \cdots,  \mathbf{w}_{\skipterm,N-1}), \mathbf{w}_{\outblock}).
\end{equation*}
Moreover, since most residual blocks have identity skip connections, the remaining number of layers the state $\sample_M$ gets propagated is comparatively low.
Hence, we can use those non-trivial skip connections as building blocks for obtaining the intermediate output of state $\sample_M$, since, similar to the standard design of early-exits, they consist of few layers for computational efficiency \citep{10.1145/3698767, 10.1145/3527155}.

In addition, since the layers contained in this intermediate output architecture are already part of the backbone network, their parameters can either be shared with the backbone network or can be additional parameters for increased representational capacity.
This way, we can re-interpret or fine-tune ResNets trained without our intermediate outputs, since all necessary parameters are at hand.

This leads to the definition of intermediate output heads $\outblock_k$, which predict $\outsample_k$ based on intermediate states $\sample_k$ via:
\begin{equation*}
	\outsample_k 
	= \outblock_k(\sample_k, \mathbf{w}_{\outblock,k}) 
	= \outblock(\skipterm_{N-1}(\cdots \skipterm_{k}(\sample_k, \mathbf{w}_{\skipterm,k}), \cdots,  \mathbf{w}_{\skipterm,N-1}), \mathbf{w}_{\outblock}), \quad\forall k = 0, \dots, N-1.
\end{equation*}
\cref{fig:resnet_cifar} illustrates the proposed flow of data through the skip connections of the original residual blocks.

This choice of intermediate output architecture enables the following theoretical guarantee, which establishes a connection to steady states (or controlled equilibria) of dynamic systems.
\begin{theorem}[Identity mapping]
	\label{lemma:identity_mapping}
	Consider the ResNet \eqref{eq:simple_resnet_dynamics}, with the intermediate outputs chosen according to \eqref{eq:proposed_early_exit}. 
	Let the intermediate outputs of residual block $M$ produce the prediction $\hat{\mathbf{y}}_M$, and let the parameters of all following residual terms be zero, $\mathbf{w}_{\resterm, k} = 0$, $k = M, ..., N-1$. Then all subsequent intermediate outputs and the final output are equal
	\begin{equation}
		\label{eq:identity_mapping}
		\hat{\mathbf{y}} = \hat{\mathbf{y}} _k = \hat{\mathbf{y}}_M,
	\end{equation}
	for all $k = M, ..., N$.
\end{theorem}
\begin{proof}
	If $\mathbf{w}_{\resterm,k}=0$, for the residual blocks $k= M, ..., N-1$, then $\resterm_k(\mathbf{x}_k, 0) = 0$ for each of these blocks.
	Hence, the state is being propagated by $\mathbf{x}_{k+1} = \skipterm_k(\mathbf{x}_k, \mathbf{w}_{\skipterm,k})$ for all $k= M, ..., N-1$. 
	This gives the output $\hat{\mathbf{y}} = \outblock(\mathbf{x}_N, \mathbf{w}_\outblock)$ and the intermediate outputs \eqref{eq:proposed_early_exit}, which corresponds to \eqref{eq:identity_mapping}.
\end{proof}

\section{Asymptotic Early-Exiting}
\label{sec:asymptotic_early_exiting}

In early-exit architectures, determining the placement of intermediate classifiers is a critical hyperparameter choice. 
Common practice is to distribute these exits uniformly across the network depth.

Similar, when considering the asymptotic-depth regime in our later  turnpike analysis, the question of  how early-exit placement should scale with network depth arises naturally.
One may fix the number of early-exits while letting the depth tend to infinity, which leads to asymptotically sparse exit positions. 
Alternatively, one may maintain a fixed spacing between exits, thereby preserving a dense placement; in this case, the number of early-exits increases asymptotically with the network depth.

Consider a sparse placement of $M$ early-exits, where $N$ is divisible by $M+1$, which leads to the joint training objective
\begin{equation*}
	\sum_{k = 1}^{M} \gamma \lossfunc\left(\hat{\mathbf{y}}_{k \frac{N}{M+1}}\right) + \lossfunc\left(\hat{\mathbf{y}}_N\right),
\end{equation*}
where the spacing $ \frac{N}{M+1}$ becomes sparse as the depth increases.
Alternatively, maintaining a fixed distance by placing early-exits every $M$ residual blocks
\begin{equation*}
	\sum_{k = 1}^{\frac{N}{M}-1} \gamma \lossfunc\left(\hat{\mathbf{y}}_{k M}\right) + \lossfunc\left(\hat{\mathbf{y}}_N\right).
\end{equation*}
In the first variant, the depth of the first early-exit is $\frac{N}{M+1}$ and decreases with the depth of the backbone networks.
In the second variant, the depth of the first early-exit is fixed at $M$.
Thus, the first variant is not suitable for early-exiting, since the depth of the exit would be unbounded.
Hence, we consider the second variant for the remainder of the paper.

\section{Turnpike Analysis of Early-Exiting}
\label{sec:turnpike_analysis}
In this section, we present our main theoretical results on the turnpike behavior of early-exiting ResNets and provide their proofs.

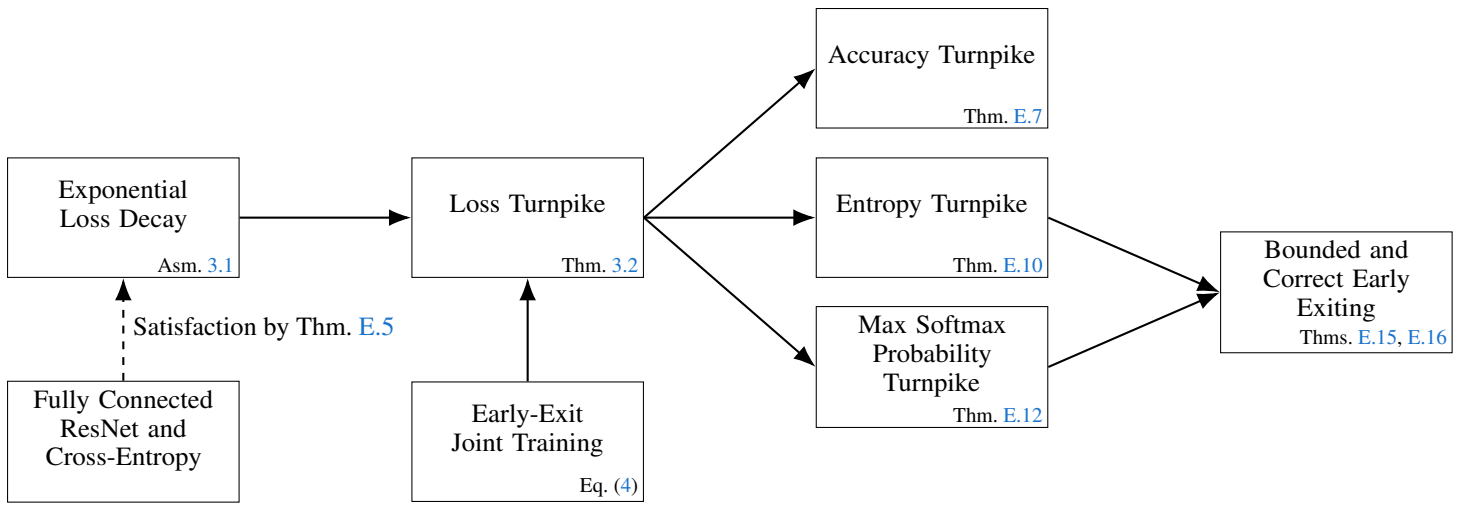
\begin{figure}[h]
    \centering
    \resizebox{0.7\columnwidth}{!}{\def\boxwidth{3.1cm}      
\def\boxheight{1.6cm}   
\def\colspacing{2.3cm}    
\def\rowspacing{2.0cm}    

\begin{tikzpicture}[
	box/.style = {
		draw, rectangle,
		minimum width=\boxwidth,
		minimum height=\boxheight,
		text width=\dimexpr\boxwidth-0.3cm\relax, 
		execute at end node={\endgraf\vspace{0.3cm}}, 
		align=center
	},
	thmref/.style={anchor=south east, font=\footnotesize, inner sep=2pt}, 
	arrow/.style={-{Latex[length=3mm]}, thick},
	dashedarrow/.style={-{Latex[length=3mm]}, thick, dashed}
	]

	\node[box] (exp) {Exponential\\Loss Decay};
	\node[thmref] at (exp.south east) {Asm.~\ref{ass:exp_reachability}};

	\node[box] (fcresnet) at ($(exp.center)+(0,-1.5*\rowspacing)$) {Fully Connected ResNet and Cross-Entropy};
	\draw[dashedarrow] (fcresnet.north) -- (exp.south)
	node[midway, right]{Satisfaction by Thm.~\ref{thm:exponential_decrease_cross_entropy}};

	\node[box, right=\colspacing of exp] (loss) {Loss Turnpike};

	\draw[arrow] (exp.east) -- (loss.west);
	\node[thmref] at (loss.south east) {Thm.~\ref{thm:turnpike_resnet}};

	\node[box] (ocp) at ($(loss.center)+(0,-1.5*\rowspacing)$) {Early-Exit Joint Training};
	\node[thmref] at (ocp.south east) {Eq.~\eqref{eq:fig_ee_training}};
	\draw[arrow] (ocp.north) -- (loss.south);

	\node[box, right=\colspacing of loss, yshift=\rowspacing] (accuracy) {Accuracy Turnpike};
	\node[box, right=\colspacing of loss] (entropy) {Entropy Turnpike};
	\node[box, right=\colspacing of loss, yshift=-\rowspacing] (maxsoft) {Max Softmax\\Probability\\Turnpike};

	\draw[arrow] (loss.east) -- (accuracy.west);
	\node[thmref] at (accuracy.south east) {Thm.~\ref{thm:accuracy_turnpike_resnet}};

	\draw[arrow] (loss.east) -- (entropy.west);
	\node[thmref] at (entropy.south east) {Thm.~\ref{thm:entropy_turnpike}};

	\draw[arrow] (loss.east) -- (maxsoft.west);
	\node[thmref] at (maxsoft.south east) {Thm.~\ref{thm:maximum_softmax_probability_turnpike}};

	\node[box, right=\colspacing of entropy, yshift=-0.5*\rowspacing] (finite) {Bounded and\\Correct Early\\Exiting};

	\draw[arrow] (entropy.east) -- (finite.west);
	\draw[arrow] (maxsoft.east) -- (finite.west);
	\node[thmref] at (finite.south east) {Thms.~\ref{thm:turnpikes_in_early_exiting_entropy}, \ref{thm:turnpikes_in_early_exiting_max_softmax}};

\end{tikzpicture}}
    \caption{Overview of the main theoretical contributions.}
    \label{fig:overview_main_contributions}
\end{figure}

In deep ResNets the later layers primarily act as refinements of the features.  
This behavior is further emphasized when using a stage cost loss, i.e., a regularization for the hidden states dependent on the considered loss function.
Below, we prove specifically that a ResNet trained with a stage cost loss exhibits the turnpike phenomenon with respect to the loss from optimal control, i.e., for any given loss threshold, only a bounded number of intermediate outputs exceed this threshold.
We establish this \textit{loss turnpike} behavior under the assumption of exponential loss decay, which we subsequently verify for a specific ResNet architecture in \cref{sec:exponential_loss_decay}.
This implies that, for most of the depth of the network, the hidden states remain near an optimal representation resulting in near zero loss for their intermediate outputs.
Moreover, the loss turnpike directly implies that a trained ResNet also exhibits an \textit{accuracy turnpike}, i.e., the number of intermediate outputs with less than perfect accuracy is bounded from above.
These implications are illustrated in \cref{fig:overview_main_contributions}.

Finally, in Appendix~\ref{sec:implications_turnpikes_early_exiting}, we show that combining accuracy turnpikes with the early-exit criterion guarantees the existence of an early-exit threshold that ensures both bounded early-exit depth and correct classification on the training data.

\assexpreachability*
In essence, this assumption requires that any finite positive loss can be achieved with some network of finite depth. 
Next, we will show that under this assumption, the ResNet trained with stage cost loss exhibits the turnpike phenomenon. 
In Appendix~\ref{sec:exponential_loss_decay}, we show that this Assumption holds for a specific ResNet architecture with fully connected layers.
Moreover, a similar result for neural ODEs can be found in \citep[Proposition 7.4]{geshkovski2021control}.

\thmturnpikeresnet*
\begin{proof}
	First, Assumption~\ref{ass:exp_reachability} gives that there exist ResNet parameters $\tilde{\mathbf{w}}$ and constants $\beta$ and $\rho \in [0,1)$, such that the training losses of the intermediate outputs $\tilde{\mathbf{y}}_k$ decays exponentially 
	\[
		\lossfunc(\tilde{\mathbf{y}}_k) \leq \beta \rho^k,
	\]
	for early-exits attached to every residual block $k \in \outsetfinaltilde = \left\{0,..., N\right\}$.
	Hence, the infimum of the joint training objective $J_N$ is upper bounded by the corresponding geometric series
	\begin{equation}
		\inf_{\mathbf{w}'} \;J_N(\mathbf{w}') 
		\leq 
		\sum_{k \in \mathcal{K}_\outblock} \gamma\lossfunc(\tilde{\mathbf{y}}_k)  + 
		\lossfunc(\tilde{\mathbf{y}})         
        \leq 
		\sum_{k \in \mathcal{K}_\outblock} \gamma\beta \rho^k  + 
		\beta \rho^N
		\leq 
		\sum_{k = 0}^{N-1} \gamma\beta \rho^k  + 
		\beta \rho^N,
	\end{equation}
	where $\tilde{\mathbf{y}}_k$ and $\tilde{\mathbf{y}}$ are the intermediate and final outputs of the ResNet with parameters $\tilde{\mathbf{w}}$ for the training data.
	Since $\rho \in [0,1)$ and $N\geq 0$, we have that $\rho^N\leq 1$ and therefore
	\begin{equation*}
		\sum_{k \in \mathcal{K}_\outblock} \gamma\beta \rho^k  + 
		\beta \rho^N
		\leq
		\sum_{k = 0}^{N-1} \gamma\beta \rho^k  + 
		\beta \rho^N 
		\leq
		\gamma \beta \frac{1 - \rho^N}{1-\rho} + \beta \rho^N
		\leq
		\frac{\gamma\beta}{1-\rho} + \beta \doteq \bar{J}_{\mathrm{inf}}.
	\end{equation*}
	Hence, the infimum of the training objective satisfies $\inf_{\mathbf{w}'} J_N(\mathbf{w}') \leq \bar{J}_{\mathrm{inf}}$ for all depths $N$ and all early-exit placements $\mathcal{K}_\outblock$.
	Therefore, for the $\delta$-suboptimal parameters $\mathbf{w}$, we have $J_N(\mathbf{w}) \leq \inf_{\mathbf{w}'} J_N(\mathbf{w}') + \delta \leq \bar{J}_{\mathrm{inf}} + \delta \doteq \bar{J}$.
	Moreover, from $0< \gamma \leq 1$, we have $\lossfunc(\hat{\mathbf{y}})  \geq \gamma\lossfunc(\hat{\mathbf{y}}) = \gamma\lossfunc(\hat{\mathbf{y}}_N)$ and therefore
	\begin{equation*}
		\sum_{k \in \outsetfinal} \gamma\lossfunc(\hat{\mathbf{y}}_k)  
        \leq \sum_{k \in \mathcal{K}_\outblock} \gamma\lossfunc(\hat{\mathbf{y}}_k)  + \lossfunc(\hat{\mathbf{y}}) 
        = J_N(\mathbf{w})
        \leq \bar{J}.
	\end{equation*}
	Moreover, with $\hat{\mathcal{Q}}_\varepsilon\doteq \left\{ k \in \outsetfinal \middle| \lossfunc(\hat{\mathbf{y}}_k) > \varepsilon\right\}$, each $k \in \hat{\mathcal{Q}}_\varepsilon$ contributes at least $\lossfunc(\hat{\mathbf{y}}_k) > \varepsilon$ to the joint training objective and hence $\gamma \varepsilon\#\hat{\mathcal{Q}}_\varepsilon  < \sum_{k \in \outsetfinal } \gamma\lossfunc(\hat{\mathbf{y}}_k)$.
	With this we obtain a bound on the cardinality of $\hat{\mathcal{Q}}_\varepsilon$
	\begin{equation}
		\gamma \varepsilon \#\hat{\mathcal{Q}}_\varepsilon  
		<
		\sum_{k \in \outsetfinal} \gamma\lossfunc(\hat{\mathbf{y}}_k) 
		\leq 
		\bar{J},
	\end{equation}
	and hence $ \#\hat{\mathcal{Q}}_\varepsilon \leq \kappa(\varepsilon) \doteq \frac{\bar{J}}{\gamma \varepsilon} = \frac{1}{\gamma\varepsilon} \left[\frac{\gamma\beta}{1-\rho} + \beta {} + \delta\right]$. This finishes the proof.
\end{proof}

Next, we consider the influence of our proposed architecture for the loss trajectories of the intermediate outputs~\eqref{eq:proposed_early_exit}.
\begin{theorem}[Decreasing loss behavior]
	\label{thm:turnpikes_without_leaving_arc}
	Consider a ResNet of depth $N$ be trained with stage cost loss weighted by $\gamma >0$ and the joint training formulation \eqref{eq:fig_ee_training} and the intermediate outputs chosen as~\eqref{eq:proposed_early_exit}, and let its parameters $\mathbf{w}$ be $\delta$-suboptimal for some fixed $0 \leq \delta < \infty$, i.e., $J_N(\mathbf{w}) \leq \inf_{\mathbf{w}'} J_N(\mathbf{w}') + \delta$, then
    the loss of the final output is bounded by
	\[
		\lossfunc(\hat{\mathbf{y}}) \leq \lossfunc(\hat{\mathbf{y}}_k) {} + \delta,
 	\]
 	compared to all intermediate outputs $k \in \mathcal{K}_\outblock$.
\end{theorem}
\begin{proof}
	Consider a ResNet trained by joint training with the $\delta$-suboptimal parameters $\mathbf{w} = \left[ \mathbf{w}_{\inblock},\mathbf{w}_{0},...,\mathbf{w}_{N-1}, \mathbf{w}_\outblock \right]$.
	Let $j = \max \argmin_{k \in \mathcal{K}_\outblock}\, \lossfunc(\hat{\mathbf{y}}_k)$ be the index of last intermediate output with the smallest loss value.

	Next, we construct parameters $\tilde{\mathbf{w}}$ that contradict the $\delta$-suboptimality of the parameters $\mathbf{w}$.
	Consider the parameters $\tilde{\mathbf{w}}$ which are equal to $\mathbf{w}$, except for the parameters of the  residual terms following $j$, which are chosen as zero $\tilde{\mathbf{w}}_{\resterm,k}$ for $k = j,\dots, N-1$.
	By Theorem~\ref{lemma:identity_mapping} the zero parameters for the residual blocks leads to an identity mapping for the following intermediate outputs $\tilde{\mathbf{y}}_k = \hat{\mathbf{y}}_j$ for $k \in \left\{ k \in \mathcal{K}_\outblock| k \geq j \right\}$ and the output $\tilde{\mathbf{y}} = \hat{\mathbf{y}}_j$.
	While the intermediate outputs before $j$ are the equal to the original intermediate outputs, $\tilde{\mathbf{y}}_k = \hat{\mathbf{y}}_k$ for $k \in \left\{ k \in \mathcal{K}_\outblock| k < j \right\}$.
	This leads to the joint training objective
	\[
		J_N(\tilde{\mathbf{w}}) = \sum_{k \in \mathcal{K}_\outblock} \gamma \lossfunc(\tilde{\mathbf{y}}_k) + \lossfunc(\tilde{\mathbf{y}})
		=
		\sum_{\substack{k \in \mathcal{K}_\outblock\\ k\leq j}} \gamma \lossfunc(\hat{\mathbf{y}}_k) +
		\sum_{\substack{k \in \mathcal{K}_\outblock\\ k> j}} \gamma \lossfunc(\hat{\mathbf{y}}_j) +
		 \lossfunc(\hat{\mathbf{y}}_j).
	\]
	The difference between the joint training objectives of the parameters $\mathbf{w}$ and $\tilde{\mathbf{w}}$ is
	\[
		J_N(\mathbf{w}) - J_N(\tilde{\mathbf{w}})
		= \lossfunc(\hat{\mathbf{y}}) - \lossfunc(\hat{\mathbf{y}}_j)
		+ \sum_{\substack{k \in \mathcal{K}_\outblock\\ k> j}} \gamma \left(\lossfunc(\hat{\mathbf{y}}_k) - \lossfunc(\hat{\mathbf{y}}_j)\right).
	\]
	Since $\inf_{\mathbf{w}'} J_N(\mathbf{w}') \leq J_N(\tilde{\mathbf{w}})$, the $\delta$-suboptimality of the parameters $\mathbf{w}$ gives
	$
		J_N(\mathbf{w}) - J_N(\tilde{\mathbf{w}}) \leq \inf_{\mathbf{w}'} J_N(\mathbf{w}') + \delta - J_N(\tilde{\mathbf{w}}) \leq \delta.
	$
	Since $j$ is the early-exit with the smallest loss, we have $\lossfunc(\hat{\mathbf{y}}_k) - \lossfunc(\hat{\mathbf{y}}_j) \geq 0$ for all $k \in \mathcal{K}_\outblock$ and all summands are non-negative such that
	$
		\lossfunc(\hat{\mathbf{y}}) - \lossfunc(\hat{\mathbf{y}}_j) \leq \delta.
	$
	Finally, $\lossfunc(\hat{\mathbf{y}}_j) \leq \lossfunc(\hat{\mathbf{y}}_k)$ for all $k \in \mathcal{K}_\outblock$ yields $\lossfunc(\hat{\mathbf{y}}) \leq \lossfunc(\hat{\mathbf{y}}_k) + \delta$.
\end{proof}
From the optimal control point of view, together with the turnpike results from Theorem~\ref{thm:turnpike_resnet}, the immediate implication of the previous result is that the leaving arc of the ResNet is bounded by its suboptimality $\delta$.
This is in contrast to the optimal trajectories of the example OCP show in \cref{fig:example_ocp}, where the trajectories towards the end of the horizon depart from the optimal steady state.
Fore more details on turnpike and leaving arcs, we refer to \citet{tudo:faulwasser22a}.

\subsection{Exponential Loss Decay}
\label{sec:exponential_loss_decay}
We now consider a specific ResNet architecture for which the exponential decay of the training loss, as stated in Assumption~\ref{ass:exp_reachability}, can be guaranteed.
This ResNet architecture consists of fully connected layers with ReLU activations, $\sigma(x) = \max \left\{0,x\right\}$, and is designed to transfer established results by~\citet{hardt2017identity} to the intermediate outputs 
\begin{subequations}
	\label{eq:resnet_reachability}
	\begin{align}
		\sample_0  &= \inblock(\sample, \mathbf{w}_\inblock) &&= E \sample, \\
		\sample_{k+1} &= f_k(\sample_k, \mathbf{w}_k) &&= \sample_{k} + W_k \sigma(A_k \sample_k + b_k), \qquad &k = 0, \ldots, N-1, \\
		\outsample_k &= \outblock_k(\sample_k, \mathbf{w}_{\outblock, k}) &&= T_k \sample_k , \qquad &k = 0, \ldots, N.
	\end{align}
\end{subequations}
The ResNet starts with the input layer $\inblock$, with parameters $\mathbf{w}_\inblock = \left[E\right]$, which embeds the input sample $\sample \in \mathbb{R}^{d_{\mathrm{in}}}$ into the hidden state $\sample_0 \in \mathbb{R}^d$ by the linear layer $E \in \mathbb{R}^{d \times d_{\mathrm{in}}}$.
The hidden state dimension $d$ is constant across all residual blocks, $\sample_k \in \mathbb{R}^d$ for all $k = 0, \ldots, N$.
Each residual block $f_k$ consists of two linear layers with parameters $A_k$, $b_k$ and $W_k$ collected into $\mathbf{w}_k = [A_k, b_k, W_k]$. 
The intermediate outputs $\outblock_k$ with parameters $\mathbf{w}_{\outblock, k} = \left[T_k\right]$ predict $\outsample_k \in \mathbb{R}^{C}$, $k = 0, \ldots, N$, via a linear layers $T_k \in \mathbb{R}^{C\times d}$, and the final prediction is given by the last intermediate output $\hat{y} = \hat{y}_N$.

The realizability results for this ResNet rely on the following assumption on the regularity of the training dataset.
\begin{assumption}[Regular dataset]
	\label{ass:distinct_data samples}
	Consider the training dataset $\dataset= \left\{ \left(x^{(1)}, y^{(1)}\right), ..., \left(x^{(D)}, y^{(D)}\right)\right\}$, then for any two distinct samples $i,j = 1,.., D$ with $i \neq j$, the distance between their features is lower bounded, i.e.,  $\lVert x^{(i)} - x^{(j)}  \rVert\geq \rho$ for some $\rho>0$. 
\end{assumption}

Given this assumption, we now prove a slightly adapted version of the realizability results by \citet{hardt2017identity}.
\begin{theorem}[Finite depth reachability for ResNets]
	\label{thm:exact_reachability_resnet}
	Let the training data $\dataset$ satisfy Assumption~\ref{ass:distinct_data samples} and consider the ResNet architecture~\eqref{eq:resnet_reachability}. 
	Moreover, for each class $c = 1, \dots, C$, we consider target vectors $\bar{y}_c \in \mathbb{R}^C$ in the output space.
	Then, there exist a finite hidden state dimension $d<\infty$ and depth $N<\infty$ and parameters $\mathbf{w}$ such that, for every sample $i =1,..., D$, the network output $\outsample$ matches the target corresponding to its label
	\begin{equation*}
		\hat{y}^{(i)}_N = \bar{y}_{y^{(i)}}.
	\end{equation*}
	Furthermore, the first $C$ components of the final hidden states $\sample_N$ represent a one-hot encoding of the labels
	\begin{equation*}
		[\sample_N]_{j} = 
		\begin{cases}
			1\quad & \text{if } y^{(i)} = j\\
			0 & \text{if } y^{(i)} \neq j,
		\end{cases}  \qquad j = 1, \dots, C.
	\end{equation*}
\end{theorem}
\begin{proof}
	\newcommand{\tildesample}{\tilde{x}^{(i)}}
	\newcommand{\tildesampleout}{\tilde{y}^{(i)}}
	Consider the alternative ResNet with width $\tilde{d}$ and depth $\tilde{N}$, given by
	\begin{subequations}
		\label{eq:alternative_resnet_reachability}
		\begin{align}
			\tildesample_0  &= \tilde{\inblock}(\sample, \tilde{\mathbf{w}}_\inblock) &&= \tilde{E} \sample, \\
			\tildesample_{k+1} &= \tilde{f}_k(\tildesample_k, \tilde{\mathbf{w}}_k) &&= \tildesample_{k} + \tilde{W}_k \sigma(\tilde{A}_k \tildesample_k + \tilde{b}_k), \qquad k = 0, \ldots, \tilde{N}-1, \\
			\tildesampleout &= \tilde{\outblock}(\tildesample_{\tilde{N}}, \tilde{\mathbf{w}}_{\outblock}) &&= \tilde{W}_{\tilde{N}} \sigma(\tilde{A}_{\tilde{N}} \tildesample_{\tilde{N}} + \tilde{b}_{\tilde{N}}),
		\end{align}
	\end{subequations}
	where the output block projects the hidden state dimension to the number of classes $C$
	by $\tilde{A}_{\tilde{N}} \in \mathbb{R}^{C\times \tilde{d}}$, $\tilde{b}_{\tilde{N}} \in \mathbb{R}^{C}$, and $\tilde{W}_{\tilde{N}} \in \mathbb{R}^{C\times C}$, which we aggregate into
	$\tilde{\mathbf{w}}_\outblock = \left[\tilde{A}_{\tilde{N}}, \tilde{b}_{\tilde{N}}, \tilde{W}_{\tilde{N}} \right]$.
	
	By the reachability results of \citet[Theorem 3.2]{hardt2017identity}, 
 	\footnote{Notice that the published paper erroneously shows a skip connection despite a dimensionality mismatch. In their rebuttal the authors clarified to Reviewer~3 that this skip connection should be omitted; see \url{https://openreview.net/forum?id=ryxB0Rtxx}.
	}
	there exists a finite width $\tilde{d}$ and depth $\tilde{N}$
	and parameters $\tilde{\mathbf{w}} = \left[ \tilde{\mathbf{w}}_{\inblock},\tilde{\mathbf{w}}_{0},...,\tilde{\mathbf{w}}_{\tilde{N}-1}, \tilde{\mathbf{w}}_\outblock \right]$ such that, for all training samples $i = 1,..,D$, the alternative network outputs a one-hot encoding of the label
	\begin{equation}
		\label{eq:alternative_resnet_one_hot}
		[\tildesampleout]_{j} = 
		\begin{cases}
			1\quad & \text{if } y^{(i)} = j\\
			0 & \text{if } y^{(i)} \neq j.
		\end{cases}.
	\end{equation}
	
	To transfer these reachability results to the architecture~\eqref{eq:resnet_reachability}, we embed the dynamics of the alternative ResNet~\eqref{eq:alternative_resnet_reachability} into~\eqref{eq:resnet_reachability}.
	In particular, we introduce an additional residual block to account for the output residual block, leading to the total depth $N = \tilde{N}+1$, and choose the hidden state dimension based on the number of classes and the hidden state dimension of the alternative ResNet $d = 2\tilde{d}+ C$.
	Given the parameters $\tilde{\mathbf{w}}$ from~\citet[Theorem 3.2]{hardt2017identity}, we construct parameters $\mathbf{w}$ for the ResNet~\eqref{eq:resnet_reachability} such that the first $C$ components of the final state of the ResNet match the output of the alternative ResNet, i.e., $[\sample_N]_{1:C} = \tilde y^{(i)}$ for all samples $i = 1, \dots, D$.
	The parameters $\mathbf{w}$ for ResNet~\eqref{eq:resnet_reachability} are
	\begin{enumerate}
		\item Input embedding
		\begin{equation}
			E = 
			\begin{bmatrix}
				\tilde{E} \\
				\mathbf{0}_{d-\tilde{d} \times d_\mathrm{in}}
			\end{bmatrix}.
		\end{equation}
		\item  First residual blocks $k = 0,..., N-2$   
		\begin{equation*}
			A_k = \begin{bmatrix}
				\tilde{A}_k & \mathbf{0}_{\tilde{d}\times d - \tilde{d}} \\
				\mathbf{0}_{d- \tilde{d}\times \tilde{d}} & \mathbf{0}_{d- \tilde{d}\times d- \tilde{d}}
			\end{bmatrix} 
			\quad
			b_k = 
			\begin{bmatrix}
				\tilde{b}_k \\
				\mathbf{0}_{d-\tilde{d}}
			\end{bmatrix}
			\quad
			W_k = \begin{bmatrix}
				\tilde{W}_k & \mathbf{0}_{\tilde{d}\times d - \tilde{d}} \\
				\mathbf{0}_{d- \tilde{d}\times \tilde{d}} & \mathbf{0}_{d- \tilde{d}\times d- \tilde{d}}
			\end{bmatrix}.
		\end{equation*}
		\item Last residual block $k = N-1= \tilde{N}$ from output residual block
		\begin{equation*}
			A_{N-1} = \begin{bmatrix}
				\tilde{A}_{\tilde{N}} 					& \mathbf{0}_{C\times d- \tilde{d}} 	\\
				\mathbf{I}_{\tilde{d}\times \tilde{d}} 	& \mathbf{0}_{\tilde{d}\times d-\tilde{d}}\\
				-\mathbf{I}_{\tilde{d}\times \tilde{d}} & \mathbf{0}_{\tilde{d}\times d-\tilde{d}} 
			\end{bmatrix} 
			\quad
			b_{N-1} = 
			\begin{bmatrix}
				\tilde{b}_{\tilde{N}} \\
				\mathbf{0}_{\tilde{d}} \\
				\mathbf{0}_{\tilde{d}} 
			\end{bmatrix}
			\quad
			W_{N-1} = \begin{bmatrix}
				\tilde{W}_{\tilde{N}} & -\mathbf{I}_{C\times \tilde{d}} & \mathbf{I}_{C\times \tilde{d}}  \\
				\mathbf{0}_{\tilde{d}\times C} & \mathbf{0}_{\tilde{d}\times \tilde{d}} & \mathbf{0}_{\tilde{d}\times \tilde{d}} \\
				\mathbf{0}_{\tilde{d}\times C}  & \mathbf{0}_{\tilde{d}\times \tilde{d}} & \mathbf{0}_{\tilde{d}\times \tilde{d}} \\
			\end{bmatrix},
		\end{equation*}
		where we denote by $\mathbf{I}_{C\times \tilde{d}} \in \mathbb{R}^{C\times \tilde{d}}$, the rectangular identity matrix defined by $[\mathbf{I}_{C\times \tilde{d}}]_{i,j}=1$ if $i=j$ and $[\mathbf{I}_{C\times \tilde{d}}]_{i,j}=0$ otherwise, for $i=1,\dots,C$ and $j=1,\dots,\tilde d$.
		In this construction, we use that for the ReLU activation function $x=\sigma(x)-\sigma(-x)$	for all $x\in\mathbb R$.
		\item Output projector matrix
		\begin{equation*}
			T_N = \begin{bmatrix}
				\bar{y}_1 & ...& \bar{y}_C & \mathbf{0}_{C\times d-C}
			\end{bmatrix}.
		\end{equation*} 
	\end{enumerate}
	
	These parameters embed the dynamics of the alternative ResNet~\eqref{eq:alternative_resnet_reachability} into ResNet~\eqref{eq:resnet_reachability}. 
	The resulting hidden states are
	\begin{equation*}
		\sample_k = \begin{bmatrix}
			\tildesample_k \\
			\mathbf{0}_{d-\tilde{d}}
		\end{bmatrix}
	\end{equation*}
	for every data sample $i = 1, \dots, D$ and each residual block $k = 0, \dots, N-1$.
	Moreover, for this terminal state this implies
	\begin{equation*}
		[\sample_N]_{1,\dots,C} = \tildesampleout.
	\end{equation*}
	Hence, by~\eqref{eq:alternative_resnet_one_hot}, the first $C$ components of the state after the last residual block coincide with the output of the alternative ResNet, which is the one-hot encoding of the label
	\begin{equation*}
		[\sample_N]_{j} = 
		\begin{cases}
			1\quad & \text{if } j = y^{(i)}\\
			0 & \text{if } j \in \left\{1,.., C\right\}\setminus \left\{y^{(i)} \right\}
		\end{cases}, \qquad j=1,\dots,C.
	\end{equation*}
	Multiplication with the output projector matrix $T_N$ yields the selected target for the label $\bar{y}_{y^{(i)}}$
	\begin{equation*}
		\hat{y}^{(i)}_N =
		T_N \sample_N = \bar{y}_{y^{(i)}}.
	\end{equation*}
\end{proof}
This result shows that any label-dependent targets $\bar{y}_y$ can be reached.
Moreover, selecting $\bar{y}_{y}$ as the output minimizing the loss function guarantees zero loss at finite depth and also ensures exponential loss decay—for instance, for the $L_2$ loss~\citep{FAULWASSER2024100290}.
However, the most commonly used loss, the cross-entropy with one-hot labels, has no finite minimizer; rather, it approaches its infimum at zero only as the network outputs (logits) grow exponentially in absolute value, thereby asymptotically ensuring maximal separation between the classes.
We now extend these finite depth reachability results as a starting point to design parameters that ensure exponentially diverging states 
such that the cross-entropy loss decays exponentially.
\begin{theorem}[Exponential decrease of cross-entropy]
	\label{thm:exponential_decrease_cross_entropy}
	Consider the ResNet architecture~\eqref{eq:resnet_reachability}, and let the training data $\dataset$ satisfy Assumption~\ref{ass:distinct_data samples}.
	Let $\lossfunc$ be the cross-entropy loss. Then, there exists an infinite-depth ResNet with finite width $d<\infty$ and parameters such that the resulting intermediate outputs $\tilde{\mathbf{y}}_k$ satisfy the exponential loss decrease from Assumption \ref{ass:exp_reachability}.
\end{theorem}
\begin{proof}
	The cross-entropy loss has no finite minimizer but converges to zero as the network states diverge to infinity~\citep[Lemma 2]{Puettschneider2026}.
	To establish exponential loss decay, we construct an infinite-depth ResNets for which the states diverge exponentially.
	This construction leverages the finite-depth reachability results from Theorem~\ref{thm:exact_reachability_resnet}.
	The finite-depth ResNet are extended to an infinite-depth ResNet using a simple choice of parameters that induces the required exponential divergence of the states for cross-entropy loss decay.
	The construction of the infinite-depth ResNet is illustrated in \cref{fig:illustration_proof_exponential_decrease_cross_entropy}.
	
	\begin{figure}[h]
		\centering
		\begin{tikzpicture}[scale=1.5]
	
	\draw[->] (-0.5,0) -- (2.5,0) node[anchor=west] {$[x]_1$};
	\draw[->] (0,-0.5) -- (0,2.5) node[anchor=south] {$[x]_2$};
	
	\foreach \x in {0.5,1.0}
	\draw (\x,0.05) -- (\x,-0.05) node[below] {\x};
	
	\foreach \y in {0.5,1.0}
	\draw (0.05,\y) -- (-0.05,\y) node[left] {\y};
	
	\fill[blue!30,opacity=0.4] (-0.5,0.5) -- (0.5,0.5) -- (0.5,2.0) -- (-0.5,2.0) -- cycle;
	
	\fill[red!30,opacity=0.4] (0.5,-0.5) -- (2.0,-0.5) -- (2.0,0.5) -- (0.5,0.5) -- cycle;
	
	\fill[red] (1.0,0) circle (2pt);
	
	\fill[blue] (0,1.0) circle (2pt);
	
	\draw[->,thick,red!70] (1.3,0.15) -- ++(0.4,0);
	\draw[->,thick,red!70] (1.3,-0.15) -- ++(0.4,0);
	
	\draw[->,thick,blue!70] (0.15, 1.3) -- ++(0,0.4);
	\draw[->,thick,blue!70] (-0.15, 1.3) -- ++(0,0.4);
	
 	\node[red] at (1.0,0.35) {Class 1};
	\node[blue, rotate=90] at (0.35, 1.0) {Class 2};
\end{tikzpicture}
		\caption{Illustration construction of the infinite-depth ResNet that ensures  exponential decay of the cross-entropy loss.
			First, Theorem~\ref{thm:exact_reachability_resnet} guarantees that all hidden states can be moved to the illustrated configuration in finite depth, where the state of each data point corresponds to the unity vector of its label.
			Then, the infinite-depth ResNet causes the hidden states to diverge exponentially along the axes of their label thereby ensuring an exponential decay of the cross-entropy loss.}
		\label{fig:illustration_proof_exponential_decrease_cross_entropy}
	\end{figure}
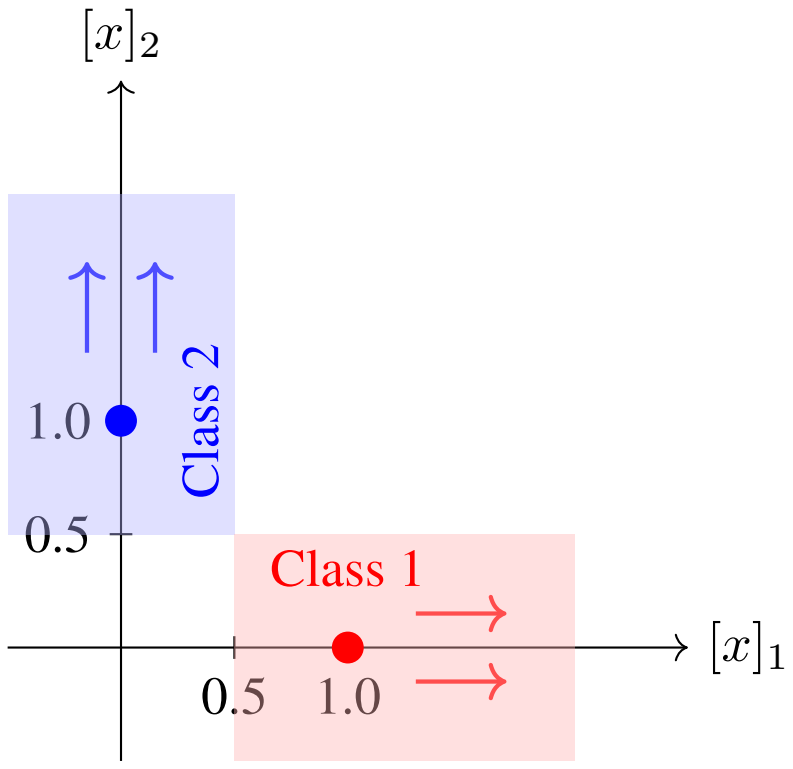 
	
	Throughout the remainder of this proof, we use the constant output projector matrix $\tilde{T}_k$ for all layers $k = P, ..., \infty$.
	\begin{equation*}
		\tilde{T}_k = 
		\begin{bmatrix}
			I_{C \times C} & 0_{C \times( d- C)}
		\end{bmatrix}.
	\end{equation*}
	With this choice, the network output corresponds to the first $C$ components of the hidden state.
	
	By Theorem \ref{thm:exact_reachability_resnet}, there exists a ResNet with finite width $d<\infty$, depth $P < \infty$, and parameters $\tilde{\mathbf{w}}$, such that all samples are separated according to their label.
	After these $P$ residual blocks, for every sample $i = 1, \ldots, D$, the hidden state is for $j = 1, \dots, C$
	\[
	[\tilde{x}^{(i)}_P]_j = 
	\begin{cases}
		1\quad & \text{if } j = y^{(i)} \\
		0 & \text{if } j \neq y^{(i)} .
	\end{cases}
	\]
	That is, after $P$ residual blocks, the state of datapoint lies on the coordinate axis corresponding to its label with a value of one.
	
	On the finite horizon $k = 0,..., P$ all intermediate output $\tilde{\mathbf{y}}_k$ are finite, leading to finite loss values $\lossfunc(\tilde{\mathbf{y}}_k)$, and hence, there exists $\beta_1 \geq 0$ and $\rho_1 \in [0,1)$ such that 
	\begin{equation*}
		\lossfunc(\tilde{\mathbf{y}}_k) \leq \beta_1\rho_1^k.
	\end{equation*} 
	
	With the states $\tilde{x}^{(i)}_P$ separated by their labels after $P$ residual blocks, we design simple infinite-depth ResNet parameters that result in an exponential growth of the state and thus prove the exponential decay of the cross-entropy.
	For $k = P,..., \infty$ we choose the parameters $\tilde{W}_k = I$, $\tilde{A}_k = 2I$, $\tilde{b}_k = -0.5\mathbf{1}$.
	This results in the ResNet dynamics
	\begin{equation*}
		\tilde{x}^{(i)}_{k+1} = \tilde{x}^{(i)}_{k} + \sigma(2 \tilde{x}_{k}^{(i)} - 0.5),
	\end{equation*}
	for $k \geq P$ and all samples $i = 1,.., D$.
	
	To show the exponential growth of the states, we consider the dynamics component-wise, which for the $j$-th state component leads to 
	\begin{equation*}
		[\tilde{x}^{(i)}_{k+1}]_{j} = [\tilde{x}^{(i)}_{k}]_{j} + \sigma(2 [\tilde{x}^{(i)}_k]_{j} - 0.5),
	\end{equation*}
	for all residual blocks $k \geq P$ and all samples $i=1,\dots,D$.
	We then show that the state component corresponding to the label growth exponentially, $[\tilde{x}^{(i)}_{k+1}]_j \geq 2 [\tilde{x}^{(i)}_{k}]_j$ for $j = y^{(i)}$, while all other components remain at zero $[\tilde{x}^{(i)}_{k+1}]_j = 0$ for $j \neq  y^{(i)}$.
	
	First, for the state dimension corresponding to the label, $j = y^{(i)}$, we have $[\tilde{x}^{(i)}_k]_{j} \geq 1$, which implies $2 [\tilde{x}^{(i)}_k]_{j} - 0.5 \geq 0$, so that for the ReLU activation $\sigma(2 [\tilde{x}^{(i)}_k]_{j} - 0.5) = 2 [\tilde{x}^{(i)}_k]_{j} - 0.5$. 
	Moreover, since $2 [\tilde{x}^{(i)}_k]_{j} - 0.5 \geq [\tilde{x}^{(i)}_k]_{j}$, it follows that
	\begin{equation*}
		[\tilde{x}^{(i)}_{k+1}]_{j} = [\tilde{x}^{(i)}_{k}]_{j} + \sigma(2 [\tilde{x}^{(i)}_k]_{j} - 0.5) =  [\tilde{x}^{(i)}_{k}]_{j} + 2 [\tilde{x}^{(i)}_k]_{j} - 0.5 \geq [\tilde{x}^{(i)}_{k}]_{j} +  [\tilde{x}^{(i)}_k]_{j}  = 2[\tilde{x}^{(i)}_k]_{j}.
	\end{equation*}
	For the other state dimensions, $j \neq y^{(i)}$, we have $[\tilde{x}^{(i)}_k]_{j} = 0$, so that the ReLU argument is negative, and therefore $\sigma(2 [\tilde{x}_k^{(i)}]_{j} - 0.5) = 0$.
	So, the off-label state components $j\neq \labelsample$ remain at zero
	\begin{equation*}
		[\tilde{x}^{(i)}_{k+1}]_{j} = [\tilde{x}^{(i)}_{k}]_{j} + \sigma(2 [\tilde{x}_k^{(i)}]_{j} - 0.5) = [\tilde{x}^{(i)}_{k}]_{j} + 0 = 0.
	\end{equation*}
	
	With the constant output projector $\tilde{T}_k = 	\begin{bmatrix} 		I_{C \times C} & 0_{C \times( d- C)} 	\end{bmatrix}$, the output is equal to the $C$ components of the state corresponding to the number of classes
	\begin{equation*}
		\tilde{y}^{(i)}_{k+1} = [\tilde{x}^{(i)}_{k+1}]_{1:C}.
	\end{equation*}
	This implies that for $k \geq P$, the intermediate outputs satisfy $[\tilde{y}^{(i)}_{k+1}]_j \geq 2 [\tilde{y}^{(i)}_{k}]_j$ and $[\tilde{y}^{(i)}_{k}]_j \geq 1$ for the component corresponding to the label $j = y^{(i)}$, while keeping the other components at zero, $[\tilde{y}^{(i)}_{k+1}]_j = 0$ for $j \neq y^{(i)}$.
	
	Now we consider the effect on the cross-entropy loss and show that this leads to exponentially decreasing loss, i.e., there exists $\rho_2 \in [0, 1)$ such that $\tilde{\lossfunc} ([\tilde{y}^{(i)}_{k+1}]_{y^{(i)}})  \leq \rho_2\tilde{\lossfunc} ([\tilde{y}^{(i)}_{k}]_{y^{(i)}})$ for all residual blocks $k = P, ..., \infty$ and all samples $i=1,...,D$.
	First, using that $[\tilde{y}^{(i)}]_j = 0$ for $j \neq y^{(i)}$, we rewrite the cross-entropy loss $\tilde{\lossfunc}$ only as a function of the component corresponding to the label $[\tilde{y}^{(i)}]_{y^{(i)}}$
	\begin{equation*}
		\begin{split}
			\lossfunc(\tilde{y}^{(i)})
			&= -\log \left( \frac{e^{[\tilde{y}^{(i)}]_{y^{(i)}}}}{ \sum_{j = 1}^{C} e^{[\tilde{y}^{(i)}]_j} }\right) 
			= \log \left(  \sum_{j = 1}^{C} e^{[\tilde{y}^{(i)}]_j - [\tilde{y}^{(i)}]_{y^{(i)}}} \right) \\ 
			&= \log \left( 1 +  \sum_{\begin{subarray}{l} j = 1 \\ j \neq y^{(i)} \end{subarray}}^{C} e^{[\tilde{y}^{(i)}]_j - [\tilde{y}^{(i)}]_{y^{(i)}}} \right)
			= \log \left(1 + \sum_{\begin{subarray}{l} j = 1 \\ j \neq y^{(i)} \end{subarray}}^{C} e^{-[\tilde{y}^{(i)}]_{y^{(i)}}} \right) \\
			&= \log \left(1 + (C-1)  e^{-[\tilde{y}^{(i)}]_{y^{(i)}}} \right) := 			\tilde{\lossfunc}([\tilde{y}^{(i)}]_{y^{(i)}}).
		\end{split}
	\end{equation*}

	The loss $\tilde{\lossfunc}([\tilde{y}^{(i)}]_{y^{(i)}}) $ is strictly decreasing in the correct class score $[\tilde{y}^{(i)}]_{y^{(i)}}$, since the logarithm is increasing and the negative exponential function decreases with $[\tilde{y}^{(i)}]_{y^{(i)}}$.
	Since $\tilde{\lossfunc}$ is decreasing, the inequality $[\tilde{y}^{(i)}_{k+1}]_{y^{(i)}} \geq 2 [\tilde{y}^{(i)}_{k}]_{y^{(i)}}$, implies that
	\begin{equation*}
		\tilde{\lossfunc}([\tilde{y}^{(i)}_{k+1}]_{y^{(i)}})   \leq \tilde{\lossfunc} (2[\tilde{y}^{(i)}_{k}]_{y^{(i)}}) .
	\end{equation*}
	Next we show that there exists $\rho_2 \in [0, 1)$ such that $\tilde{\lossfunc} ([\tilde{y}^{(i)}_{k+1}]_{y^{(i)}})  \leq \rho_2\tilde{\lossfunc} ([\tilde{y}^{(i)}_{k}]_{y^{(i)}})$ for $k = P, ..., \infty$. To this end, consider the ratio
	\begin{equation*}
		\frac{\tilde{\lossfunc} ([\tilde{y}^{(i)}_{k+1}]_{y^{(i)}})}{\tilde{\lossfunc} ([\tilde{y}^{(i)}_{k}]_{y^{(i)}})} \leq 
		\frac{\tilde{\lossfunc} (2[\tilde{y}^{(i)}_{k}]_{y^{(i)}})}{\tilde{\lossfunc} ([\tilde{y}^{(i)}_{k}]_{y^{(i)}})}  
		= \frac{
			\log \left(1 + (C-1) e^{-2[\tilde{y}^{(i)}_k]_{y^{(i)}}} \right)
		}{
			\log \left(1 + (C-1) e^{-[\tilde{y}^{(i)}_k]_{y^{(i)}}} \right)
		}.
	\end{equation*}
	The numerator and denominator only differ by the additional factor 2 in the exponent of the numerator, which causes the numerator to decrease faster than the denominator, hence the ratio attains its maximum for the smallest possible $[\tilde{y}^{(i)}_k]_{y^{(i)}}$, which is $[\tilde{y}^{(i)}_k]_{y^{(i)}} = 1$ and hence for all $k = P,..., \infty$
	\begin{equation*}
		\frac{\tilde{\lossfunc} ([\tilde{y}^{(i)}_{k+1}]_{y^{(i)}})}{\tilde{\lossfunc} ([\tilde{y}^{(i)}_{k}]_{y^{(i)}})} \leq 
		\frac{
			\log \left(1 + (C-1) e^{-2[\tilde{y}^{(i)}_k]_{y^{(i)}}} \right)
		}{
			\log \left(1 + (C-1) e^{-[\tilde{y}^{(i)}_k]_{y^{(i)}}} \right)
		} 
		\leq
		\frac{
			\log \left(1 + (C-1) e^{-2} \right)
		}{
			\log \left(1 + (C-1) e^{-1} \right)
		} = \rho_2 
		< 1.
	\end{equation*}
	
	We obtain an exponential bound on the deeper residual blocks as $\beta_2 = \lossfunc (\tilde{y}^{(i)}_{P}) \rho_2^{-P}$, $\lossfunc (\tilde{y}^{(i)}_{k}) \leq \beta_2 \rho_2^{k}$ for $k = P,..., \infty$ for all data samples $i = 1,..., D$.
	Moreover, since the decay holds for all data samples it also hold for the empirical risk
	\begin{equation*}
		\lossfunc(\tilde{\mathbf{y}}_{k})
		= \frac{1}{D} \sum_{i=1}^{D} \lossfunc(\tilde{y}^{(i)}_{k})
		\leq 
		\frac{1}{D} \sum_{i=1}^{D}\beta_2 \rho_2^k
		= \beta_2\rho_2^k.
	\end{equation*}
	
	For the entire depth, $k = 0,..., \infty$, we obtain an exponentially decaying bound by taking the maximum of the two bounds $\lossfunc(\tilde{\mathbf{y}}_{k}) \leq \beta_1 \rho_1^{k}$ for $k = 0,..., P$ and $\lossfunc(\tilde{\mathbf{y}}_{k}) \leq \beta_2 \rho_2^{k}$ for $k = P,..., \infty$. 
	Then with  $\beta = \max \left\{\beta_1, \beta_2\right\}$ and $\rho = \max \left\{\rho_1, \rho_2\right\} < 1$, we get the exponentially decaying bound
	\begin{equation*}
		\lossfunc(\tilde{\mathbf{y}}_k) \leq \beta\rho^k,
	\end{equation*} 
	for the entire depth $k = 0,..., \infty$.
	Such that the exponential loss decay from Assumption~\ref{ass:exp_reachability} is satisfied.
\end{proof}

\subsection{Accuracy Turnpike}
\label{sec:accutracy_turnpike}
After showing that turnpikes exist for the cross-entropy loss, we now show turnpikes also occur for other metrics, in this section, we start with the accuracy.
We start by showing that an upper bound on the loss provides a lower bound on the accuracy.

\begin{lemma}[Loss bound implies accuracy  bound]
	\label{thm:accuracy_bound_by_loss_bound}
	Let $\dataset$ be a dataset of size $D$, and let $\hat{\mathbf{y}}$ denote the corresponding output.
	If the average cross-entropy loss satisfies for some number of samples $B \in \mathbb{N}_{[0,D]}$
	\begin{equation*}
		\lossfunc(\hat{\mathbf{y}}) 
		\leq \frac{B+1}{2D} \log(2),
	\end{equation*}
	then at most $B$ samples are misclassified, and hence the accuracy is lower bounded by
	$\mathcal{A}(\hat{\mathbf{y}}) \geq 1 - \frac{B}{D}$.
\end{lemma}
\begin{proof}
	First, we show that, for a single sample with prediction $\hat{y}$ and label $y$ the upper bound on the loss
	\begin{equation*}
		\lossfunc(\hat{y}) < \log(2).
	\end{equation*}
	ensures correct classification of this sample, i.e., $\argmax_{c} \left[\hat{p}(\hat{y})\right]_c  = y$.
	If $\lossfunc(\hat{y}) = -\log([\hat{p}(\hat{y})]_y)< \log(2)$, then $[\hat{p}(\hat{y})]_y > 0.5$.
	Moreover, since the softmax probabilities are non-negative and sum up to one, it follows that $[\hat{p}(\hat{y})]_c<0.5$ for any $c \in \left\{1,..., C\right\}\setminus \left\{y\right\}$ 
	Therefore $[\hat{p}(\hat{y})]_y>[\hat{p}(\hat{y})]_c$ for any $c \in \left\{1,..., C\right\} \setminus \left\{y\right\}$ and therefore the sample is classified correctly.
	
	For the empirical risk on the full dataset, this implies that each misclassified sample must incur a loss larger or equal to $\log(2)$.
	Hence, if the average cross-entropy satisfies
	\begin{equation*}
		\lossfunc(\hat{\mathbf{y}}) = \frac{1}{D} \sum_{i = 1}^{D} \lossfunc(\outsample) < \frac{B+1}{D} \log(2),
	\end{equation*}
	then at most $B$ samples are misclassified.
	Moreover, since $\frac{B+1}{2D} < \frac{B+1}{D}$ for all $B \geq 0$ and $D \geq 1$, this holds in particular if $\lossfunc(\hat{\mathbf{y}}) \leq \frac{B+1}{2D} \log(2)$.
\end{proof}
With this relation between the loss and the accuracy, we now show that the loss turnpike implies an accuracy turnpike.
\begin{theorem}[Accuracy turnpike]
	\label{thm:accuracy_turnpike_resnet}
	Suppose that Assumption \ref{ass:exp_reachability} holds, and consider a ResNet of depth $N$ with early-exits attached to set the set of residual blocks $\outset$ trained with stage cost loss and weight $0 < \gamma \leq 1$ by the optimal control training~\eqref{eq:fig_ee_training}, and let its parameters $\mathbf{w}$ be $\delta$-suboptimal for some fixed $0 \leq \delta < \infty$, i.e., $J_N(\mathbf{w}) \leq \inf_{\mathbf{w}'} J_N(\mathbf{w}') + \delta$.
	Then the following hold:
	\begin{enumerate}
		\setlength{\itemsep}{0pt}
		\setlength{\topsep}{0pt}
		\setlength{\parsep}{0pt}
		\setlength{\parskip}{0pt}
		\item[(i)] 	Then the ResNet exhibits an accuracy turnpike on the training data: for any number of misclassified samples $B\in \mathbb{N}_{[0,D]}$, the number of intermediate outputs which misclassify more than $B$ samples is bounded by some $\kappa<\infty$ independent of the depth of the ResNet $N$ and the early-exit placement $\outset \subseteq \mathbb{N}_{[0,N-1]}$.
		\item[(ii)] 	For the set of early-exits with accuracy $\mathcal{A}(\hat{\mathbf{y}}_k)  < 1 -\frac{B}{D} $ this implies
		\begin{equation*}
			\#
			\left\{ k \in \outsetfinal \middle| 
			\mathcal{A}(\hat{\mathbf{y}}_k)  < 1-\frac{B}{D} 
			\right\} \leq \kappa,
		\end{equation*}
		where $\hat{\mathbf{y}}_k$, $k = 0,..., N$, denotes the intermediate outputs on the training data.
		\end{enumerate}
\end{theorem}
\begin{proof}
	By Lemma \ref{thm:accuracy_bound_by_loss_bound}, $\lossfunc(\hat{\mathbf{y}}_k) 
	\leq \frac{B+1}{2D} \log(2)$ implies $\mathcal{A}(\hat{\mathbf{y}}_k)  \geq 1 -\frac{B}{D} $
	and therefore $\mathcal{A}(\hat{\mathbf{y}}_k)  < 1- \frac{B}{D}$ requires $\lossfunc(\hat{\mathbf{y}}_k)  > \frac{B+1}{2D} \log(2)$.
    For the set of early-exits with accuracy less than $1- \frac{B}{D}$ this implies
	\begin{equation*}
		\#
		\left\{ k \in \outset \middle| 
		\mathcal{A}(\hat{\mathbf{y}}_k)  < 1- \frac{B}{D}
		\right\} \leq\# \left\{ k \in \outset \middle| 
		\lossfunc(\hat{\mathbf{y}}_k) > \frac{B+1}{2D} \log(2)
		\right\}.
	\end{equation*}
	Under Assumption~\ref{ass:exp_reachability}, by Theorem~\ref{thm:turnpike_resnet} the jointly trained ResNet exhibits a loss turnpike and therefore, there exists $\kappa< \infty$ such that
	\begin{equation*}
		\#
		\left\{ k \in \outset \middle| 
		\lossfunc(\hat{\mathbf{y}}_k) > \frac{B+1}{2D} \log(2)
		\right\}
		\leq \kappa,
	\end{equation*}
    and therefore $\left\{ k \in \outset \middle| 
		\mathcal{A}(\hat{\mathbf{y}}_k)  < 1- \frac{B}{D}
		\right\} \leq \kappa$.
\end{proof}
\subsection{Entropy Turnpike}
\label{sec:entropy_turnpike}
In this section, we show that the cross-entropy loss turnpike also induces a turnpike in the entropy, a commonly used criterion for early-exiting \citep{10.1145/3527155}.
We begin by establishing a relationship between the the cross-entropy loss and the entropy.

\begin{theorem}[Bound on cross-entropy implies bound on entropy]
	\label{thm:bound_on_loss_implies_bound_entropy}
	Let $\dataset$ be any dataset of size $D$, and let $\hat{\mathbf{y}}$ denote the predicted output.
	Suppose that the cross-entropy loss satisfies the bound 
	\[
	\lossfunc(\hat{\mathbf{y}}) \leq \bar{\mathcal{L}}
	\]
	for $\bar{\lossfunc}>0$, then the entropy  $\entropy$ is bounded by
	\[
	\entropy(\hat{\mathbf{y}}) \leq \bar{\entropy}(D\bar{\mathcal{L}}),
	\]
	where 
	\begin{equation}
		\label{eq:max_entropy}
		\bar{\entropy}(\bar{\mathcal{L}}) = 
		\begin{cases} 
			\log(C), & \bar{\mathcal{L}} \ge \log(C),\\[1mm]
			\bar{\mathcal{L}} e^{-\bar{\mathcal{L}}} - \left(1 - e^{-\bar{\mathcal{L}}}\right) \log\Big(\frac{1 - e^{-\bar{\mathcal{L}}}}{C-1}\Big), &  \bar{\mathcal{L}} < \log(C).
		\end{cases}
	\end{equation}
	Moreover, the entropy for every sample $i=1,...,D$ satisfies
	\[
	\entropy(\outsample) \leq \bar{\entropy}(D\bar{\mathcal{L}}).
	\]
\end{theorem}
\begin{proof}
	First, we consider a single sample with label $y$ and the cross-entropy and entropy as a function of the probabilities $p \in [0,1]^{C}$.
	We show that if the cross-entropy loss satisfies
	\[
	\lossfunc_{p}(p) = -\log(\left[p\right]_y) \leq \bar{\mathcal{L}}
	\]  
	with \(\bar{\mathcal{L}} \geq 0\), then there exist a corresponding bound on the entropy $\bar{\entropy}(\bar{\mathcal{L}})$ such that
	\[
	\entropy_p(p) = -\sum_{c = 1}^{C} \left[p\right]_c\log(\left[p\right]_c) \leq \bar{\entropy}(\bar{ \mathcal{L}}).
	\]
	
	First, we rewrite the upper bound on the cross-entropy $\lossfunc_p(p) = -\log([p]_y) \leq \bar{\mathcal{L}}$ as a lower bound on the softmax probability of the label class $y$, $[p]_y \geq e^{-\bar{\mathcal{L}}}$.
	Next, consider the maximization of the entropy on the probability simplex subject to the label probability constraint $[p]_y \ge e^{-\bar{\mathcal{L}}}$
	\begin{subequations}
		\begin{align}
			\bar{\entropy}(\bar{ \mathcal{L}}) = \max_{p \in \mathbb{R}^C} \quad & \entropy_p(p) = - \sum_{c=1}^{C} [p]_c \log([p]_c) \\
			\text{subject to} \quad & \sum_{c=1}^{C} [p]_c = 1,\\
			& [p]_c \ge 0,  &c=1,\dots,C,\\
			& [p]_y \ge e^{-\bar{\mathcal{L}}}.
		\end{align}
		\label{eq:entropy_maximization_loss_constraint}
	\end{subequations}
	The feasible set of this optimization problem is
	\begin{equation*}
		\mathcal{P} = \left\{ p \in [0,1]^C\middle| [p]_y \ge e^{-\bar{\mathcal{L}}}, \sum_{c=1}^{C}  [p]_c = 1 \right\}.
	\end{equation*}
	We define a candidate probability vector $p \in \mathcal{P}$
	\begin{equation*}
		[p]_i = 
		\begin{cases} 
			\frac{1}{C} 							&\quad\text{if $\bar{\mathcal{L}} \ge \log(C)$} ,\\
			e^{-\bar{\mathcal{L}}} 					&\quad\text{if $\bar{\mathcal{L}} < \log(C)$ and $i = y$}, \\
			\frac{ 1- e^{-\bar{\mathcal{L}}}}{C-1} 	&\quad\text{if $\bar{\mathcal{L}} < \log(C)$  and $i \neq y$}.
		\end{cases}
	\end{equation*}
	The probability distribution vector $p$ is being majorized by any other feasible probability distribution $q \in \mathcal{P}$.
	That is, if $[q]_c^\downarrow$ and $[p]_c^\downarrow$ denote the components of $q$ and $p$ sorted in decreasing order, respectively, then
	\[
	\sum_{c=1}^j [q]_c^\downarrow \ge \sum_{c=1}^j [p]_c^\downarrow \quad \text{for all } j = 1,\dots,C-1,
	\]
	with equality for $j = C$ due to normalization.
	And since the entropy $\entropy_p$ is Schur-concave \citep{marshall2010inequalities} it follows that $\entropy_p(p) \geq \entropy_p(q)$ for all $ q \in \mathcal{P}$, so $p$ is a maximizer to \eqref{eq:entropy_maximization_loss_constraint}.
	The resulting maximum value is $\bar{\entropy}(\bar{\lossfunc})$ from \eqref{eq:max_entropy}.
	
	With the softmax probability distribution $\hat{p}(\hat{y})$ of the network output $\hat{y}$ for a single sample, the relationship between the loss bound $\lossfunc(\hat{y}) = \lossfunc_p \left( \hat{p}(\hat{y}) \right) \leq \bar{\lossfunc}$ and $\entropy(\hat{y}) = \entropy_p \left( \hat{p}(\hat{y}) \right) \leq \bar{\entropy}(\bar{\lossfunc})$ directly translates to the loss and entropy as a function of the network output $\hat{y}$.
	
	Next, we prove that an upper bound on the empirical loss
	\begin{equation*}
		\lossfunc(\hat{\mathbf{y}}) 
		= \frac{1}{D}\sum_{i=1}^{D} \lossfunc(\outsample) \leq \bar{\mathcal{L}}
	\end{equation*}
	implies a corresponding bound on the entropy averaged over the dataset
	\begin{equation*}
		\entropy(\hat{\mathbf{y}}) 
		= \frac{1}{D}\sum_{i=1}^{D} \entropy(\outsample) \leq \bar{\entropy}(D \bar{ \mathcal{L}}). 
	\end{equation*}
	Since the loss function is non-negative, it follows that for each sample, $i = 1,..., D$, $\lossfunc(\outsample) \leq D\bar{\mathcal{L}}$.
	And thus, the entropy for each sample, $i = 1,..., D$, is bounded by $\entropy(\outsample) \leq \bar{\entropy}(D \bar{\lossfunc}) $. 
	Hence, the averaged entropy suffices
	\begin{equation*}
		\entropy(\hat{\mathbf{y}}) 
		= \frac{1}{D}\sum_{i=1}^{D} \entropy(\outsample)
		\leq 
		\frac{1}{D}\sum_{i = 1}^{D} \bar{ \entropy}(D\bar{\lossfunc})
		= \bar{ \entropy}(D\bar{\lossfunc}).
	\end{equation*}
\end{proof}

\begin{theorem}[Entropy bound from loss bound]
	\label{thm:entropy_bound_by_loss_bound}
	For every $T>0$, there exists $\varepsilon>0$, such that
	$\lossfunc(\hat{\mathbf{y}}) \leq \varepsilon$ implies 
	$\entropy(\hat{\mathbf{y}}) \leq T$.
\end{theorem}
\begin{proof}
	By Theorem~\ref{thm:bound_on_loss_implies_bound_entropy}, the entropy satisfies the bound
	\[
	\entropy(\hat{\mathbf{y}}) \leq \bar{\entropy}\!\left(D\lossfunc(\hat{\mathbf{y}})\right),
	\]
	where $\bar{\entropy}$ is a continuous, nonnegative function with  
	\[
	\lim_{\bar{\mathcal{L}} \to 0} \bar{\entropy}(D\bar{\mathcal{L}}) = 0.
	\]
	Therefore, for any $T > 0$, continuity of $\bar{\entropy}$ guarantees the existence of $\varepsilon > 0$ such that $\lossfunc(\hat{\mathbf{y}}) \leq \varepsilon$ implies $\bar{\entropy}\!\left(D\lossfunc(\hat{\mathbf{y}})\right) \leq T$.
\end{proof}

\begin{theorem}[Entropy turnpike]
	\label{thm:entropy_turnpike}
	Suppose that Assumption \ref{ass:exp_reachability} holds, and consider a ResNet of depth $N$ with early-exit attached to the set of residual blocks $\outset$ trained with stage cost loss $0 < \gamma \leq 1$ by the optimal control training~\eqref{eq:fig_ee_training} and $\lossfunc$ as the cross-entropy loss, and let its parameters $\mathbf{w}$ be $\delta$-suboptimal for some fixed $0 \leq \delta < \infty$, i.e., $J_N(\mathbf{w}) \leq \inf_{\mathbf{w}'} J_N(\mathbf{w}') + \delta$.
	Then, the ResNet exhibits an entropy turnpike on the training data, i.e., for any entropy threshold $T>0$, there exists a depth bound $\kappa<\infty$ such that 
	\begin{equation}
		\#
		\left\{ k \in \outsetfinal \middle| 
		\entropy(\hat{\mathbf{y}}_k)  > T
		\right\} \leq \kappa,
		\label{eq:entropy_turnpike}
	\end{equation}
	where $\hat{\mathbf{y}}_k$ denotes the intermediate outputs on the training data.
\end{theorem}
\begin{proof}
	By Theorem \ref{thm:entropy_bound_by_loss_bound}, for any $T> 0$, there exists $\varepsilon>0$, such that $\lossfunc(\hat{\mathbf{y}}) \leq \varepsilon$ implies $\entropy(\hat{\mathbf{y}}) \leq T$, hence $\entropy(\hat{\mathbf{y}}_k)  > T$ requires $\lossfunc(\hat{\mathbf{y}}_k) > \varepsilon$ and hence
	\begin{equation*}
		\#
		\left\{ k \in \outsetfinal \middle| 
		\entropy(\hat{\mathbf{y}}_k)  > T
		\right\}
		\leq 
		\#
		\left\{ k \in \outsetfinal \middle| 
		\lossfunc(\hat{\mathbf{y}}_k) > \varepsilon
		\right\}.
	\end{equation*}
	Moreover, under Assumption~\ref{ass:exp_reachability} by Theorem~\ref{thm:turnpike_resnet}, the ResNet trained with stage cost loss weight $0 < \gamma \leq 1$ exhibits a loss turnpike, i.e.,	for every $\varepsilon>0$, there exists $\kappa< \infty$ such that 
	\begin{equation*}
		\#
		\left\{ k \in \outsetfinal \middle| 
		\lossfunc(\hat{\mathbf{y}}_k) > \varepsilon
		\right\}
		\leq \kappa,
	\end{equation*}
    and hence $		\#
		\left\{ k \in \outsetfinal \middle| 
		\entropy(\hat{\mathbf{y}}_k)  > T
		\right\}
		\leq \kappa$.
\end{proof}

\subsection{Maximum Softmax Probability Turnpike}
\label{sec:maximum_softmax_probability_turnpike}
In this section, we show that the cross-entropy loss turnpike also induces a turnpike in the maximum softmax probability, another commonly used criterion for early-exiting \cite{10.1145/3527155}.
We begin by establishing a relationship between the cross-entropy loss and the maximum softmax probability.

\begin{theorem}[Maximum softmax probability bound by loss bound]
	\label{thm:max_softmax_proabability_by_loss_bound}
	For every $\underline{p} \in (0,1)$, if the cross-entropy loss $\lossfunc(\hat{\mathbf{y}})$ for some network output $\hat{\mathbf{y}}$ satisfies
	\[
	\lossfunc(\hat{\mathbf{y}}) \leq \bar{\lossfunc} = \frac{1}{D}\min \left\{\log(2),  -\log\left(1-\underline{p}\right)\right\},
	\]
	then for every sample $i = 1, \dots, D$, the maximum predicted class probability is bounded from below by
	\[
	\max_{c} \left[\hat{p}(\outsample)\right]_c \;\geq\; 1 - \underline{p},
	\]
	and moreover, the predicted class is the label class, i.e.,
	\[
	\argmax_{c} \left[\hat{p}(\outsample)\right]_c = \labelsample.
	\]
\end{theorem}
\begin{proof}
	Suppose that the averaged cross-entropy loss satisfies
	\begin{equation*}
		\lossfunc(\hat{\mathbf{y}}) = 
		\frac{1}{D} \sum_{i = 1}^{D} \lossfunc(\outsample)
		\leq \bar{\lossfunc}.
	\end{equation*}
	Since the cross-entropy is nonnegative for each sample, $\lossfunc(\outsample) \geq 0$, it follows that
	\begin{equation*}
		\lossfunc(\outsample) = -\log \left( \left[\hat{p}(\outsample)\right]_{y^{(i)}}\right) \leq D \bar{\mathcal{L}},
	\end{equation*}
	and hence
	\begin{equation*}
		\left[\hat{p}(\outsample)\right]_{y^{(i)}} \geq e^{-D\bar{\lossfunc}}.
	\end{equation*}
	Since $D\bar{\lossfunc} \leq \log(2)$, we have that the probability of the label class is lower bounded by $\left[\hat{p}(\outsample)\right]_{y^{(i)}}  \geq \frac{1}{2}$.
	The softmax probabilities for all other classes are strictly larger than zero, i.e. $\left[\hat{p}(\outsample)\right]_c>0$ for all classes $c = 1,...,C$.
	Hence, we have that $\argmax_c \left[\hat{p}(\outsample)\right]_{c} = y^{(i)}$ and $\left[\hat{p}(\outsample)\right]_{y^{(i)}} =  \max_c \left[\hat{p}(\outsample)\right]_c$.
	With $\bar{\lossfunc} = \frac{1}{D}\min \left\{\log(2),  -\log\left(1-\underline{p}\right)\right\}$, we have that for each data sample $i = 1,..., D$
	\begin{equation*}
		\max_c \left[\hat{p}(\outsample)\right]_c 
		= \left[\hat{p}(\outsample)\right]_{y^{(i)}} 
		\geq e^{ -D\bar{\lossfunc} } = \max \left\{\frac{1}{2}, 1-\underline{p} \right\} \geq 1 - \varepsilon.
	\end{equation*}
\end{proof}

\begin{theorem}[Maximum softmax probability turnpike]
	\label{thm:maximum_softmax_probability_turnpike}
	Let Assumption \ref{ass:exp_reachability} hold.
	Consider the ResNet of depth $N$ with the early-exit placements $\outset$ trained by \eqref{eq:fig_ee_training} with stage cost weight $0 < \gamma \leq 1$ and $\lossfunc$ as the cross-entropy loss, and let its parameters $\mathbf{w}$ be $\delta$-suboptimal for some fixed $0 \leq \delta < \infty$, i.e., $J_N(\mathbf{w}) \leq \inf_{\mathbf{w}'} J_N(\mathbf{w}') + \delta$.
	Then, the ResNet exhibits an max softmax probability on the training data, i.e., for any threshold $\underline{p}  \in (0,1)$, there exists a bound $\kappa<\infty$ on the number of early-exits for which a data sample has a softmax probability less than $1-\underline{p}$, i.e.,
	\begin{equation}
		\#
		\left\{ k \in \outsetfinal \middle| 
		    \exists i = 1,..., D, \; \max_c \left[\hat{p}(\outsample_k)\right]_c  < 1-\underline{p}
		\right\} \leq \kappa.
		\label{eq:maximum_softmax_probability_turnpike}
	\end{equation}
\end{theorem}
\begin{proof}
	By Theorem \ref{thm:max_softmax_proabability_by_loss_bound}, for any $\underline{p}> 0$, there exists $\varepsilon>0$, such that $\lossfunc(\hat{\mathbf{y}}_k) \leq \varepsilon$ implies $\max_c \left[\hat{p}(\outsample_k)\right]_c  \geq 1-\underline{p}$ for all samples $i = 1,..., D$, 
    therefore, if there exists $i=1,..., D$, $\max_c \left[\hat{p}(\outsample_k)\right]_c  < 1-\underline{p}$, then $\lossfunc(\hat{\mathbf{y}}_k) > \varepsilon$ and thus
    and hence 
	\begin{equation*}
		\#
		\left\{ k \in \outsetfinal \middle| 
		\exists i = 1,..., D, \; \max_c \left[\hat{p}(\outsample_k)\right]_c  < 1-\underline{p}
		\right\}
		\leq 
		\#
		\left\{ k \in \outsetfinal \middle| 
		\lossfunc(\hat{\mathbf{y}}_k) > \varepsilon
		\right\}.
	\end{equation*}
	Moreover, under Assumption~\ref{ass:exp_reachability} by Theorem~\ref{thm:turnpike_resnet}, the ResNet trained with stage cost weight $0 < \gamma \leq 1$ exhibits a loss turnpike, i.e.,
	for every $\varepsilon>0$, there exists $\kappa< \infty$ such that 
	\begin{equation*}
\resizebox{\linewidth}{!}{$\displaystyle
    \#\left\{ k \in \outsetfinal \middle|
    \exists i = 1,\dots, D, \; \max_c \left[\hat{p}(\hat{y}^{(i)}_k)\right]_c < 1-\underline{p}
    \right\}
    \leq
    \#\left\{ k \in \outsetfinal \middle| \lossfunc(\hat{\mathbf{y}}_k) > \varepsilon \right\}
    \leq \kappa,
$}
\end{equation*}
	which gives \eqref{eq:maximum_softmax_probability_turnpike}.
\end{proof}

\subsection{Implication of Turnpikes on Early-Exiting}
\label{sec:implications_turnpikes_early_exiting}
Next, we analyze the implications of the turnpikes in the entropy and the maximum softmax probabilities on the early-exiting schemes, which use entropy and maximum softmax probability threshold as the early-exit criterion.
First we show that under an upper bound on the loss, there exists an entropy threshold that guarantees correct classification. 

\begin{theorem}[Entropy threshold for correct classification under loss bound]
	\label{thm:entropy_threshold_correct_classification}
	Suppose that the loss function satisfies 
	\begin{equation*}
		\lossfunc(\hat{\mathbf{y}}) \leq \bar{\lossfunc},
	\end{equation*}
	for $\bar{\mathcal{L}}>0$, then the early-exiting threshold
	\begin{equation}
		T(\bar{\mathcal{L}}) = 
		\begin{cases}
			\infty 																								& \bar{\lossfunc} < \frac{1}{D}\log(2) \\
			D\bar{\lossfunc} e^{-D\bar{\lossfunc}} - \left(1-e^{-D\bar{\lossfunc}}\right) \log \left(1- e^{-D\bar{\lossfunc}}\right) \quad	& \bar{\lossfunc} \geq \frac{1}{D}\log(2),
		\end{cases}
		\label{eq:minimum_value_entropy}
	\end{equation}
	guarantees that if a sample $i=1,...,D$ satisfies
	\begin{equation*}
		\entropy(\outsample) < T(\bar{\mathcal{L}}),
	\end{equation*}
	then this sample is classified correctly
	\begin{equation*}
		\hat{c}(\outsample) = y^{(i)}. 
	\end{equation*}
\end{theorem}
\begin{proof}
	The bound 
	\begin{equation*}
		\lossfunc(\hat{\mathbf{y}}) 
		= \frac{1}{D} \sum_{i = 1}^{D} \lossfunc(\outsample)
		\leq \bar{\lossfunc}
	\end{equation*}
	implies, that for each sample $i = 1,..., D$, $\lossfunc(\outsample) \leq D\bar{\lossfunc}$.
	The cross-entropy upper bound $\lossfunc_p(p) = -\log(\left[p\right]_y)\leq D\bar{\lossfunc}$ can be rewritten as a lower bound on the probability of the label class $\left[p\right]_y \geq e^{-D\bar{\lossfunc}}$.
	If $\bar{\lossfunc} < \frac{1}{D}\log(2)$, then $\left[p\right]_y > 0.5$, and thus correct classification is ensured independent of the entropy.
	
	Consequently, we only need to consider the nontrivial case where $\bar{\lossfunc} \geq \frac{1}{D}\log(2)$, so that the loss bound alone cannot guarantee correct classification.
	For this case, we want to find the minimum value of the entropy for which a false classification occurs.
    For this, we search for the probability distribution vector $p \in \left[0,1\right]^C$ with the minimal entropy value and $\left[p\right]_y \geq e^{-D\bar{\lossfunc}}$, for which a false classifications occurs, i.e, there exists a class $j\in \mathbb{N}_{[1,C]}\setminus\left\{y\right\}$ with $\left[p\right]_j \geq \left[p\right]_y$.
	This leads to the optimization problem for the entropy as a function of the probability distribution $p \in \left[0,1\right]^C$
	\begin{subequations}
		\begin{align}
			T(\bar{\lossfunc}) =\min_{ p} 			\;	& \entropy_p(p) \\
			\text{s.t.} \quad 	\left[p\right]_y &\geq e^{-D\bar{\lossfunc}} \\
			\left[p\right]_j &\geq \left[p\right]_y \\
			\left[p\right]_c &\geq 0 \qquad \text{$c \in \left\{1,..., C\right\} \setminus \{j, y\}$} \\
			\sum_{i = 1}^C \left[p\right]_i &= 1.
		\end{align}
		\label{eq:minimization_entropy_loss_bound}
	\end{subequations}
	The feasible set of this optimization problem is
	\begin{equation*}
		\mathcal{P} = \left\{ p \in [0,1]^C \middle| 
		\left[p\right]_y \geq e^{-D\bar{\lossfunc}},\; 
		\left[p\right]_j \geq \left[p\right]_y,\;
		\sum_{c=1}^{C} \left[p\right]_c = 1,  \right\}.
	\end{equation*}
	Define the feasible distribution $p \in \mathcal{P}$ on the boundary of $\mathcal{P}$ element-wise as
	\[
	[p]_i =
	\begin{cases}
		e^{-D\bar{\lossfunc}}, & i = y,\\[2mm]
		1 - e^{-D\bar{\lossfunc}}, & i = j \neq y,\\[1mm]
		0, & i \notin \{y,j\}.
	\end{cases}
	\]
	Then $p$ majorizes any other probability distribution $q \in \mathcal{P}$; that is, if $[q]_c^\downarrow$ and $[p]_c^\downarrow$ denote the components of $q$ and $p$ sorted in decreasing order, respectively, then
	\[
	\sum_{c=1}^j [p]_c^\downarrow \ge \sum_{c=1}^j [q]_c^\downarrow \quad \text{for all } j = 1,\dots,C-1,
	\]
	with equality for $j = C$ due to normalization.
	And since the entropy $\entropy_p$ is strictly Schur-concave \cite{marshall2010inequalities} it follows that $\entropy_p(p) \leq \entropy_p(q)$ for all $ q \in \mathcal{P}$, so $p$ is a minimizer of \eqref{eq:minimization_entropy_loss_bound}.
    Moreover, due to the strict Schur-concavity it holds that $p$ is the unique minimizer and $entropy_p(p) < entropy_p(q)$ for $q \in \mathcal{P} \setminus \left\{p\right\}$.
	The corresponding minimal entropy value is \eqref{eq:minimum_value_entropy}.
	Hence, if a probability distribution $p \in [0,1]^C$ satisfies $\left[p\right]_y \geq e^{-D\bar{\lossfunc}}$ and $\entropy_p(p) < T(\bar{\lossfunc})$, then $[p]_c < [p]_y$ for all $c \in \left\{1,..., C\right\}\setminus\left\{y\right\}$.

	This gives that if $\lossfunc(\hat{\mathbf{y}}) \leq \bar{\mathcal{L}}$ and $\entropy(\outsample) < T(\bar{\mathcal{L}})$, then $\outsample$ is classified correctly, i.e. $\hat{c}(\outsample) = y^{(i)}$. 
\end{proof}

\begin{theorem}[Max softmax probability threshold for correct classification under loss bound]
	\label{thm:max_softmax_probability_threshold_correct_classification}
	Suppose that the loss function satisfies 
	\begin{equation*}
		\lossfunc(\hat{\mathbf{y}}) \leq \bar{\lossfunc},
	\end{equation*}
	for $\bar{\mathcal{L}}>0$, then the early-exiting threshold
	\begin{equation}
		\underbar{p}(\bar{\mathcal{L}}) = 
		\begin{cases}
			1 																	& \bar{\lossfunc} < \frac{1}{D}\log(2), \\
			e^{-D\bar{\lossfunc}}  									\quad	& \bar{\lossfunc} \geq \frac{1}{D}\log(2),
		\end{cases}
		\label{eq:minimum_value_max_softmax}
	\end{equation}
	guarantees that if a sample $i=1,...,D$ satisfies
	\begin{equation*}
				\max_c \left[\hat{p}(\outsample)\right]_c > 1-\underbar{p}(\bar{\mathcal{L}}),
	\end{equation*}
	then that sample is classified correctly
	\begin{equation*}
		\hat{c}(\outsample) = y^{(i)}. 
	\end{equation*}
\end{theorem}
\begin{proof}
	The loss bound
	\begin{equation*}
		\lossfunc(\hat{\mathbf{y}}) 
		= \frac{1}{D} \sum_{i = 1}^{D} \lossfunc(\outsample)
		\leq \bar{\lossfunc}
	\end{equation*}
	implies, that for each sample, $i = 1,..., D$, $\lossfunc(\outsample) \leq D\bar{\lossfunc}$.
	The cross-entropy upper bound $\lossfunc_p(p) = -\log(\left[p\right]_y)\leq D\bar{\lossfunc}$ can be rewritten as a lower bound on the probability of the label class $\left[p\right]_y \geq e^{-D\bar{\lossfunc}}$.
	If $\bar{\lossfunc} < \frac{1}{D}\log(2)$, then $\left[p\right]_y > 0.5$, and thus correct classification is ensured independent of the maximum softmax probability.
	
	Consequently, we only need to consider the nontrivial case where $\bar{\lossfunc} \geq \frac{1}{D}\log(2)$, so that the loss bound alone cannot guarantee correct classification.
	With $\left[p\right]_y \geq e^{-D\bar{\lossfunc}}$ and since the probabilities sum up to one it follows that for all off label classes $c\neq \labelsample$ $\left[p\right]_c \leq 1- e^{-D\bar{\lossfunc}}$.
	Therefore, whenever $\max_c \left[\hat{p}(\outsample)\right]_c > 1 -\underbar{p}(\bar{\mathcal{L}}) $, then $\hat{c}(\outsample) = \argmax_c \left[\hat{p}(\outsample)\right]_c=  y^{(i)}$.
\end{proof}

\begin{theorem}[Bounded and correct early-exiting with entropy by turnpikes]
	\label{thm:turnpikes_in_early_exiting_entropy}
	Let Assumption~\ref{ass:exp_reachability} hold.
	And consider the ResNet with depth $N$ and early-exits attached to the set of residual blocks $\outset$ trained by \eqref{eq:fig_ee_training} with stage cost weight $0 < \gamma \leq 1$ and $\lossfunc$ as the cross-entropy loss, and let its parameters $\mathbf{w}$ be $\delta$-suboptimal for some fixed $0 \leq \delta < \infty$, i.e., $J_N(\mathbf{w}) \leq \inf_{\mathbf{w}'} J_N(\mathbf{w}') + \delta$.
	Then consider the early-exiting scheme with entropy thresholds $T$, where the propagation of a data sample $\sample$ is prematurely terminated at RB $k$ with the output $\outsample_k$ if $\entropy(\outsample_k) \leq T$.
	Then, there exists an early-exiting threshold $T>0$, such that if a sample $i = 1,..., D$ exits before the final output, i.e., $\entropy(\outsample_k) \leq T$ for $k \in \outset$, it is classified correctly
	\begin{equation*}
		\hat{c}(\outsample_k) = y^{(i)}.
	\end{equation*}
	Moreover, there exists $\kappa$ such that each sample $i=1,...,D$ is exited at latest at the $(\kappa+1)$-th early-exit
    \begin{equation*}
		\# \left\{k \in \outsetfinal | \entropy(\outsample_k) > T \right\} \leq \kappa,
	\end{equation*}
    independent of the depth $N$ and the early-exit placement.
    For the early-exit depth in terms of the number of residual blocks this implies that the exit depth is bounded by 
	\begin{equation*}
		k^{(i)} = \min \left( \left\{k \in \outsetfinal | \entropy(\outsample_k) \leq T \right\} \cup \left\{ N \right\}\right) \leq \seq \left(\outsetfinal \right)_{\kappa+1}.
	\end{equation*}
\end{theorem}
\begin{proof}
	We begin by showing that there exists an entropy threshold that guarantees correct classification, which requires deriving an upper bound on the intermediate losses.
	By Assumption~\ref{ass:exp_reachability} and Theorem~\ref{thm:turnpike_resnet}, there exists a constant $\bar{J}$ such that, for every depth $N$ and every placement $\outset$, the training objective evaluated for the $\delta$-suboptimal parameters $\mathbf{w}$ satisfies
	\[
	J_N(\mathbf{w}) \leq \bar{J}.
	\]
	Hence, we have
	\[
	\sum_{k \in \outset} \gamma \lossfunc(\hat{\mathbf{y}}_k) + \lossfunc(\hat{\mathbf{y}}) \leq \bar{J},
	\]
	where $\hat{\mathbf{y}}_k$ denotes the intermediate outputs and $\hat{\mathbf{y}}$ the final output for the training data.  
	Since $\gamma \leq 1$, we have $\lossfunc(\hat{\mathbf{y}}) \geq \gamma \lossfunc(\hat{\mathbf{y}})$, which gives the bound
	\[
	\sum_{k\in \outsetfinal}\gamma \lossfunc(\hat{\mathbf{y}}_k) 
	\leq
	\sum_{k\in \outset} \gamma \lossfunc(\hat{\mathbf{y}}_k) + \lossfunc(\hat{\mathbf{y}})
	\leq 
	\bar{J}.
	\]
	Moreover, due to the non-negativity of the loss function, each intermediate loss is individually bounded:
	\[
	\lossfunc(\hat{\mathbf{y}}_k) \leq \frac{\bar{J}}{\gamma}, \quad k \in \outsetfinal,
	\]
	for any ResNet depth $N$ and any early-exit placement $\outset$.
	
	By Theorem~\ref{thm:entropy_threshold_correct_classification}, if the loss of an intermediate output satisfies $\lossfunc(\hat{\mathbf{y}}_k) \leq \frac{\bar{J}}{\gamma}$, then the entropy threshold $T\left(\frac{\bar{J}}{\gamma}\right)$ guarantees correct classification. 
	That is, if $\entropy(\outsample_k) < T\left(\frac{\bar{J}}{\gamma}\right)$ for a sample $i = 1, \dots, D$, then the predicted class equals the label, $\hat{c}(\outsample_k) = y^{(i)}$.
	Hence, we choose $T = \frac{1}{2}T\left(\frac{\bar{J}}{\gamma}\right)$ and get that if $\entropy(\outsample_k) \leq T$, then the sample is classified correctly.
	
	To relate this sample-wise early-exiting criterion to the average entropy, observe that it is sufficient to ensure
	\[
	\entropy(\hat{\mathbf{y}}) = \frac{1}{D} \sum_{i=1}^{D} \entropy(\outsample) \leq \frac{1}{D}T.
	\]
	Then, since entropy of each individual sample is non-negative $\entropy(\outsample) \geq 0$ for all  $i = 1, \dots, D$, it follows that for every data sample $i = 1, \dots, D$ we have $\entropy(\outsample) \leq T$.
	Moreover, under Assumption~\ref{ass:exp_reachability}, a ResNet trained by \eqref{eq:fig_ee_training} exhibits an entropy turnpike by Theorem~\ref{thm:entropy_turnpike}. 
	That is, there exists constant $\kappa<\infty$ such that the number of intermediate outputs violating this entropy threshold is bounded
	\begin{equation*}
		\#
		\left\{ k \in \outsetfinal \middle| 
		\entropy(\hat{\mathbf{y}}_k)  > \frac{1}{D}T
		\right\} \leq \kappa,
	\end{equation*}
	and hence, the early-exit depth for all samples $i=1,...,D$ is bounded by the $(\kappa+1)$-th early-exit
	\begin{equation*}
		k^{(i)} = \min \left(\left\{k \in \outsetfinal | \entropy(\outsample_k) \leq T \right\}\cup \left\{ N \right\}\right) \leq \seq \left(\outsetfinal \right)_{\kappa+1}.
	\end{equation*}
\end{proof}

\begin{theorem}[Bounded and correct early-exiting with max softmax probabilities by turnpikes]
	\label{thm:turnpikes_in_early_exiting_max_softmax}
	Let Assumption~\ref{ass:exp_reachability} hold.
	And consider the ResNet with depth $N$ trained by \eqref{eq:fig_ee_training} with stage cost weight $0 < \gamma \leq 1$ and $\lossfunc$ as the cross-entropy loss, and let its parameters $\mathbf{w}$ be $\delta$-suboptimal for some fixed $0 \leq \delta < \infty$, i.e., $J_N(\mathbf{w}) \leq \inf_{\mathbf{w}'} J_N(\mathbf{w}') + \delta$.
	Then consider the early-exiting scheme with the probability threshold $\underline{p}$, where the propagation of a data sample $\sample$ is prematurely terminated at RB $k \in \outset$ with the output $\outsample_k$ if $\max_c \left[\hat{p}(\outsample_k)\right]_c \geq 1 -\underline{p}$.
	Then, there exists an early-exiting threshold $\underline{p}>0$ and early-exiting depth $\kappa$, such that for every sample $i = 1,..., D$ the early-exiting depth is bounded independent of the network depth $N$
    \begin{equation}
		\# \left\{k \in \outsetfinal \middle|  \max_c \left[\hat{p}(\outsample_k)\right]_c < 1 -\underline{p} \right\} \leq \kappa.
	\end{equation}
    Moreover, the early-exit depth in terms of the residual block is bounded by
	\begin{equation}
		k^{(i)} = \min \left( \left\{k \in \outsetfinal  \middle|  \max_c \left[\hat{p}(\outsample_k)\right]_c \geq 1-\underline{p} \right\} \cup \left\{N\right\}\right)\leq \seq \left(\outsetfinal \right)_{\kappa+1}.
	\end{equation}
	and if a sample exits before the final output, i.e., $\max_c \left[\hat{p}(\outsample_k)\right]_c \geq 1-\underline{p}$ for $k \in \outset$, it is classified correctly
	\begin{equation*}
		\hat{c}(\outsample_k) = y^{(i)}.
	\end{equation*}
\end{theorem}
\begin{proof}
	We begin by showing that there exists an max softmax probability threshold that guarantees correct classification, which requires deriving an upper bound on the intermediate losses.
	By Assumption~\ref{ass:exp_reachability} and Theorem~\ref{thm:turnpike_resnet}, there exists a constant $\bar{J}$ such that, for every depth $N$, the training objective evaluated for the $\delta$-suboptimal parameters $\mathbf{w}$ satisfies
	\[
	J_N(\mathbf{w}) \leq \bar{J}.
	\]
	Hence, we have
	\[
	\sum_{k\in \outset} \gamma \lossfunc(\hat{\mathbf{y}}_k) + \lossfunc(\hat{\mathbf{y}}) \leq \bar{J},
	\]
	where $\hat{\mathbf{y}}_k$ denotes the intermediate outputs and $\hat{\mathbf{y}} = \hat{\mathbf{y}}_N$ the final output for the training data.  
	Since $\gamma \leq 1$, we have $\lossfunc(\hat{\mathbf{y}}) \geq \gamma \lossfunc(\hat{\mathbf{y}})$, which gives the bound
	\[
	\sum_{k \in \outsetfinal} \gamma \lossfunc(\hat{\mathbf{y}}_k) 
	\leq
	\sum_{k \in \outset} \gamma \lossfunc(\hat{\mathbf{y}}_k) + \lossfunc(\hat{\mathbf{y}})
	\leq 
	\bar{J}.
	\]
	Moreover, due to the non-negativity of the loss function, each intermediate loss is individually bounded:
	\[
	\lossfunc(\hat{\mathbf{y}}_k) \leq \frac{\bar{J}}{\gamma}, \quad k \in \outsetfinal,
	\]
	for any ResNet depth $N$ and any early-exit placement $\outset$.
	
	By Theorem~\ref{thm:max_softmax_probability_threshold_correct_classification}, if the loss of an intermediate output satisfies $\lossfunc(\hat{\mathbf{y}}_k) \leq \frac{\bar{J}}{\gamma}$, then the probability threshold $\underline{p}\left(\frac{\bar{J}}{\gamma}\right)$ guarantees correct classification. 
	That is, if $\max_c \left[\hat{p}(\outsample)\right]_c > 1- \underline{p}\left(\frac{\bar{J}}{\gamma}\right)$ for a sample $i = 1, \dots, D$, then the predicted class equals the label, $\hat{c}(\outsample_k) = \labelsample$.
	Hence, we choose $\underline{p} = \frac{1}{2} \underline{p}\left(\frac{\bar{J}}{\gamma}\right)$ and get that if $\max_c \left[\hat{p}(\outsample)\right]_c \geq 1- \underline{p}$, then the sample is classified correctly.
	
	Secondly, under Assumption~\ref{ass:exp_reachability} a ResNet trained by~\eqref{eq:fig_ee_training} exhibits a max softmax probability turnpike by Theorem~\ref{thm:maximum_softmax_probability_turnpike}.     
    That is, there exists a $\kappa$, such that for every sample $i = 1,...,D$ the number of early-exits with softmax probability less than $1- \underline{p}$ is bounded
	\begin{equation}
		\# \left\{k \in \outsetfinal \middle| \exists i = 1,..., D, \; \max_c \left[\hat{p}(\outsample_k)\right]_c < 1- \underline{p} \right\} \leq \kappa,
	\end{equation}
	independent of the depth $N$ and the early-exit placement $\outset$.
	Moreover, under Assumption~\ref{ass:exp_reachability}, a ResNet trained by \eqref{eq:fig_ee_training} exhibits an max softmax probability turnpike by Theorem~\ref{thm:entropy_turnpike}. 
	That is, since $1- \underline{p} >0$, there exists constant $\kappa<\infty$ such that 
	\begin{equation*}
    \begin{split}
    		k^{(i)} &= 
            \min \left(\left\{k \in \outsetfinal \middle|  \max_c \left[\hat{p}(\outsample_k)\right]_c \geq  1- \underline{p} \right\} \cup \left\{N\right\}\right) \\
            &\leq
    		\seq \left(\outsetfinal \right)_{\kappa+1}
        \end{split}
	\end{equation*}
	for all samples $i=1,...,D$, and hence the early-exiting depth is bounded by $\seq \left(\outsetfinal \right)_{\kappa+1}$.
\end{proof}

\section{Scope and Limitations}
\label{sec:scope_and_limitations}
Our findings and theoretical insights contribute to our understanding of early-exit neural networks. Crucially, by connecting practical training concepts to optimal control theory, we are able to introduce the turnpike phenomenon to the deep learning community and utilize the tools to uncover a systematic similarity behavior of the latent dynamics involved in joint training early-exits of ResNets. 
We want to highlight, that the analysis involved common assumptions about the asymptotic optimality of the data problem leading us to the starting point of our reasoning in Assumption~\ref{ass:exp_reachability}. As noted in the main text, we were able to prove it for a realistic scenario of cross-entropy loss and fully connected ResNets with ReLU activations. 
It remains to future work, to extend this loss decay result to other classes of networks, commonly employed for early-exiting, i.e., Transformer backbones and even networks without the residual connection.
Even though our experimental suite strongly supports our theoretical findings, we want to highlight the inevitable gap between asymptotic statements for networks trained to global optimality and the finite precision of the stochastic gradient training. 
We believe that our work contributes greatly to the rigor involved in bridging the gap between optimal control theory and practical deep learning experiments. 


\newpage
\section*{NeurIPS Paper Checklist}

\begin{enumerate}

\item {\bf Claims}
    \item[] Question: Do the main claims made in the abstract and introduction accurately reflect the paper's contributions and scope?
    \item[] Answer: \answerYes{} 
    \item[] Justification: 
	The main claims of the abstract are supported: the theoretical contributions are described in Section~\ref{sec:mainresult} and Section~\ref{sec:exitingturnpike}, with proofs provided in Appendix~\ref{sec:turnpike_analysis}. The experimental findings supporting these claims are presented in Section~\ref{sec:experiments} and further supplemented in Appendix~\ref{sec:experimental_supplements}.

\item {\bf Limitations}
    \item[] Question: Does the paper discuss the limitations of the work performed by the authors?
    \item[] Answer: \answerYes{} 
    \item[] Justification: In Appendix~\ref{sec:scope_and_limitations}, we discuss the limitations of this work.

\item {\bf Theory assumptions and proofs}
    \item[] Question: For each theoretical result, does the paper provide the full set of assumptions and a complete (and correct) proof?
    \item[] Answer: \answerYes{} 
    \item[] Justification: The paper provides the full proofs and the assumptions in Appendix~\ref{sec:turnpike_analysis}.

    \item {\bf Experimental result reproducibility}
    \item[] Question: Does the paper fully disclose all the information needed to reproduce the main experimental results of the paper to the extent that it affects the main claims and/or conclusions of the paper (regardless of whether the code and data are provided or not)?
    \item[] Answer: \answerYes{} 
    \item[] Justification: 
    The paper discloses all information necessary to reproduce the main experimental results. We use standard ResNet architectures as backbone networks and evaluate on standard benchmark datasets, with full details — including hyperparameters, training configuration, and dataset splits — provided in Appendix~\ref{sec:experimental_setup}. The early-exit architecture is described in Section~\ref{sec:architecture}. 


\item {\bf Open access to data and code}
    \item[] Question: Does the paper provide open access to the data and code, with sufficient instructions to faithfully reproduce the main experimental results, as described in supplemental material?
    \item[] Answer: \answerYes{} 
    \item[] Justification: The code and instructions to reproduce our results is available at \url{https://github.com/OptCon/early-exit-turnpike}. All datasets used in this work are publicly available and are established benchmark datasets in the community.

\item {\bf Experimental setting/details}
    \item[] Question: Does the paper specify all the training and test details (e.g., data splits, hyperparameters, how they were chosen, type of optimizer) necessary to understand the results?
    \item[] Answer: \answerYes{} 
    \item[] Justification: 
    All training and test details are specified. We follow standard data splits for all benchmarks. Hyperparameters and optimizer settings follow established conventions for ResNet architectures, with full details in Appendix~\ref{sec:experimental_setup}. 
    The early-exit architecture and its additional hyperparameters are described in Section~\ref{sec:architecture} and Appendix~\ref{sec:experimental_setup}, and ablated in Section~\ref{sec:evaluation_stage_cost_weight} and Appendix~\ref{sec:gamma_ablation}.

\item {\bf Experiment statistical significance}
    \item[] Question: Does the paper report error bars suitably and correctly defined or other appropriate information about the statistical significance of the experiments?
    \item[] Answer: \answerNo{} 
    \item[] Justification: 
    Error bars and formal statistical significance tests are not reported, as the primary contribution of this paper is theoretical and our extensive ResNet trainings impose across all configurations and datasets an infeasible computational overhead. 
    Nevertheless, the robustness of our findings is supported by consistent results across multiple datasets (Section~\ref{sec:further_experiments}) and the ablation of the hyperparameter stage cost weight in Appendix~\ref{sec:gamma_ablation}.

\item {\bf Experiments compute resources}
    \item[] Question: For each experiment, does the paper provide sufficient information on the computer resources (type of compute workers, memory, time of execution) needed to reproduce the experiments?
    \item[] Answer: \answerYes{} 
    \item[] Justification: GPU resources used in our experiments are reported in Appendix~\ref{sec:experimental_setup}. Execution times are not reported, as our primary contribution is a theoretical analysis of hidden representations, which is independent of the underlying hardware.
    
\item {\bf Code of ethics}
    \item[] Question: Does the research conducted in the paper conform, in every respect, with the NeurIPS Code of Ethics \url{https://neurips.cc/public/EthicsGuidelines}?
    \item[] Answer: \answerYes{} 
    \item[] Justification: 
    This research conforms with the NeurIPS Code of Ethics. Our primary contribution is theoretical, and all experiments are conducted on publicly available benchmark datasets, raising no ethical concerns regarding data privacy, fairness, or misuse.

\item {\bf Broader impacts}
    \item[] Question: Does the paper discuss both potential positive societal impacts and negative societal impacts of the work performed?
    \item[] Answer: \answerNA{} 
    \item[] Justification: 
    As our primary contribution is a theoretical analysis of hidden representations in early-exit networks, we do not foresee any direct negative societal impacts. A broader discussion of the societal implications of early-exit networks in general is beyond the scope of this work.
    
\item {\bf Safeguards}
    \item[] Question: Does the paper describe safeguards that have been put in place for responsible release of data or models that have a high risk for misuse (e.g., pre-trained language models, image generators, or scraped datasets)?
    \item[] Answer: \answerNA{} 
    \item[] Justification: 
    This paper presents a theoretical analysis and experiments on standard benchmark datasets. No new data, pre-trained models, or other high-risk assets are released, so safeguards are not applicable.

\item {\bf Licenses for existing assets}
    \item[] Question: Are the creators or original owners of assets (e.g., code, data, models), used in the paper, properly credited and are the license and terms of use explicitly mentioned and properly respected?
    \item[] Answer: \answerYes{} 
    \item[] Justification: All assets used in this work are properly credited. The benchmark datasets are referenced in Section~\ref{sec:experiments} and Appendix~\ref{sec:experimental_setup}, and the PyTorch version used for implementation is specified in Appendix~\ref{sec:experimental_setup}. All assets are used in accordance with their respective licenses.

\item {\bf New assets}
    \item[] Question: Are new assets introduced in the paper well documented and is the documentation provided alongside the assets?
    \item[] Answer: \answerYes{} 
    \item[] Justification: 
    The only new asset introduced in this paper is our code, which will be publicly released with the camera-ready version and will include documentation sufficient to reproduce our experimental results.

\item {\bf Crowdsourcing and research with human subjects}
    \item[] Question: For crowdsourcing experiments and research with human subjects, does the paper include the full text of instructions given to participants and screenshots, if applicable, as well as details about compensation (if any)? 
    \item[] Answer: \answerNA{} 
    \item[] Justification: 
    This paper does not involve crowdsourcing or research with human subjects.

\item {\bf Institutional review board (IRB) approvals or equivalent for research with human subjects}
    \item[] Question: Does the paper describe potential risks incurred by study participants, whether such risks were disclosed to the subjects, and whether Institutional Review Board (IRB) approvals (or an equivalent approval/review based on the requirements of your country or institution) were obtained?
    \item[] Answer: \answerNA{} 
    \item[] Justification: 
    This paper does not involve crowdsourcing or research with human subjects, and therefore requires no IRB approval.

\item {\bf Declaration of LLM usage}
    \item[] Question: Does the paper describe the usage of LLMs if it is an important, original, or non-standard component of the core methods in this research? Note that if the LLM is used only for writing, editing, or formatting purposes and does \emph{not} impact the core methodology, scientific rigor, or originality of the research, declaration is not required.
    \item[] Answer: \answerNA{} 
    \item[] Justification:
    No LLMs were used as a component of the core methods in this research. 
    Any use of LLMs was limited to writing and editing assistance, which does not impact the methodology or scientific contributions of the paper.

\end{enumerate}

\end{document}